%% file: main.tex
\documentclass[10pt,twocolumn,letterpaper]{article}

\usepackage[pagenumbers]{cvpr} 

\input{preamble}

\definecolor{cvprblue}{rgb}{0.21,0.49,0.74}
\usepackage[pagebackref,breaklinks,colorlinks,allcolors=cvprblue]{hyperref}

\def\paperID{*****} 
\def\confName{***}
\def\confYear{***}

\title{ViewBridge: Revisiting Cross-View Localization from Image Matching}

\author{
Panwang Xia\textsuperscript{1}* \quad
Qiong Wu\textsuperscript{1}* \quad
Lei Yu\textsuperscript{2} \quad
Yi Liu\textsuperscript{1} \quad
Mingtao Xiong\textsuperscript{1} \quad
Xudong Lu\textsuperscript{3} \quad
Yi Liu\textsuperscript{1} \\
Haoyu Guo\textsuperscript{1} \quad
Yongxiang Yao\textsuperscript{1} \quad
Junjian Zhang\textsuperscript{2} \quad
Xiangyuan Cai\textsuperscript{2} \\
Hongwei Hu\textsuperscript{2} \quad
Zhi Zheng\textsuperscript{3} \quad
Yongjun Zhang\textsuperscript{1} \quad
Yi Wan\textsuperscript{1}$^{\dagger}$ \\
\textsuperscript{1}Wuhan University \quad 
\textsuperscript{2}Ant Group \quad
\textsuperscript{3}The Chinese University of Hong Kong \\
{* Equally contribution. $^{\dagger}$ Corresponding author.} \\
{\tt\small \{xiapanwang, mabel\_wq, yi.wan\}@whu.edu.cn}
}

\usepackage{multirow}
\usepackage{tikz}

\begin{document}

\maketitle
\input{sec/0_abstract}    
\input{sec/1_intro}
\input{sec/2_relatedwork}

\input{sec/3_method}
\input{sec/4_experiments}

\input{sec/5_conclusion}

{
    \small
    \bibliographystyle{ieeenat_fullname}
    \bibliography{main}
}


\input{sec/X_suppl}


\end{document}

%% file: preamble.tex
\usepackage{arydshln}
\usepackage{colortbl}
\usepackage{diagbox}
\usepackage{gensymb}
\usepackage{graphicx}
\usepackage{mathtools}
\usepackage{microtype}
\usepackage{multirow}
\usepackage{nicefrac}
\usepackage{scalerel}
\usepackage{siunitx}
\usepackage{soul}
\usepackage{textcomp}
\usepackage[normalem]{ulem}

\usepackage{array}
\newcolumntype{P}[1]{>{\centering\arraybackslash}p{#1}}

\colorlet{first}{Green!30}       
\colorlet{second}{LimeGreen!30}   
\colorlet{third}{Yellow!30}      

\definecolor{darkgreen}{rgb}{0.0,0.6,0.4}

%% file: sec/0_abstract.tex
\begin{abstract}
Cross-view localization aims to estimate the 3-DoF pose of a ground-view image by aligning it with aerial or satellite imagery. Existing methods typically address this task through direct regression or feature alignment in a shared bird’s-eye view (BEV) space. Although effective for coarse alignment, these methods fail to establish fine-grained and geometrically reliable correspondences under large viewpoint variations, thereby limiting both the accuracy and interpretability of localization results. Consequently, we revisit cross-view localization from the perspective of image matching and propose a unified framework that enhances both matching and localization. Specifically, we introduce a Surface Model that constrains BEV feature projection to physically valid regions for geometric consistency, and a SimRefiner that adaptively refines similarity distributions to enhance match reliability. To further support research in this area, we present CVFM, the first benchmark with 32,509 cross-view image pairs annotated with pixel-level correspondences. Extensive experiments demonstrate that our approach achieves geometry-consistent and fine-grained correspondences across extreme viewpoints and further improves the accuracy and stability of cross-view localization. 

\end{abstract}

%% file: sec/1_intro.tex
\section{Introduction}
\label{sec:intro}

Cross-view localization aims to estimate the 3 degrees of freedom (3-DoF) pose of a ground-view image by aligning it with aerial or satellite imagery, where the output typically includes the location and yaw orientation. This task is essential for visual positioning in GNSS-denied environments, such as urban canyons and post-disaster areas. Applications span autonomous navigation~\cite{maddern20171}, urban planning~\cite{ye2024sg}, and emergency response~\cite{li2025cross}.

\begin{figure}[ht]
    \centering
    \includegraphics[width=1\linewidth]{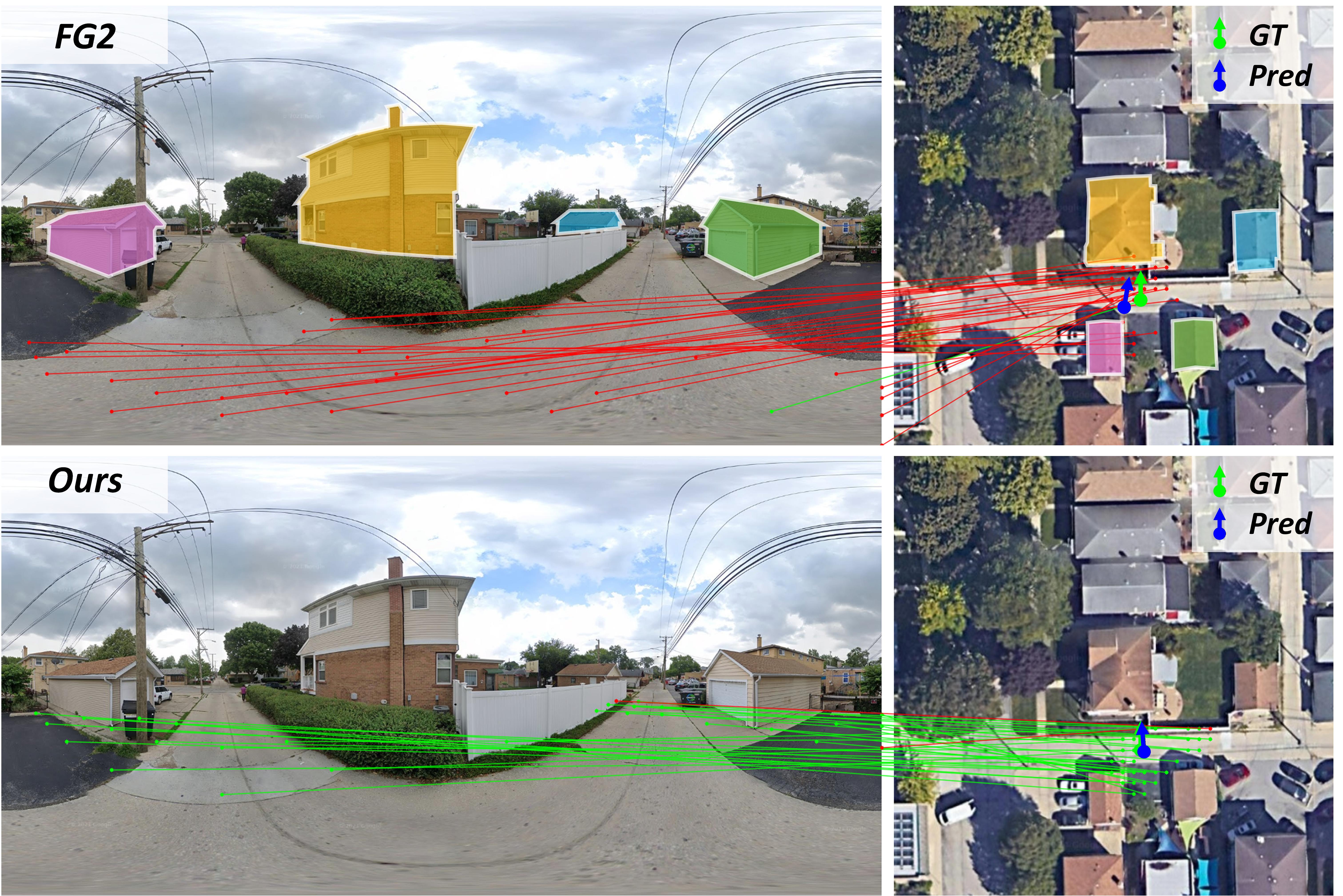}
\caption{Cross-view localization and matching results.
Top: FG2~\cite{xia2025fg}. Bottom: Ours. Green arrows indicate ground-truth poses, and blue arrows indicate predicted poses. Green lines denote correct matches, while red lines denote incorrect ones. For clarity, corresponding objects across views are highlighted with consistent colors. }
    \label{fig:intro}
    \vspace{-1.0em}
\end{figure}

To tackle this problem, existing methods follow two major directions. 
The first directly regresses the camera pose from cross-view pairs to encode structural priors and learn spatial correspondences~\cite{lentsch2023slicematch,xia2022visual,xia2023convolutional,zhang2025cross}.  
The second maps both views into a shared space, such as bird’s-eye view (BEV), where geometric alignment is more tractable~\cite{shi2023boosting,song2023learning,wang2023fine,xia2025fg}.  
Although these approaches differ in formulation, they share a common bottleneck: the need of reliable spatial correspondences across different viewpoints. 
In regression-based methods, correspondences quality directly determines localization accuracy; in BEV-based pipelines, it governs the reliability of BEV representations, as incorrect correspondences lead to degraded feature generation.
Thus, the key to improving localization lies in establishing geometrically consistent correspondences across perspectives, i.e., cross-view image matching. 

Image matching defines how one view can be interpreted from another, and is a cornerstone for tasks such as scene reconstruction, view synthesis, and cross-modal representation learning. 
In the cross-view setting, effective image matching not only improves localization accuracy, but also contributes to broader cross-view understanding by enabling semantic and geometric integration across perspectives. 
However, existing matching methods~\cite{sarlin2020superglue,sun2021loftr,edstedt2024roma,wang2024dust3r,leroy2024grounding,barroso2024matching}, though successful in standard image-pair settings, often struggle to handle the extreme appearance and viewpoint variations between ground and satellite images.

In this work, we revisit cross-view localization from the perspective of image matching and propose a unified framework that enhances both tasks (see Fig.~\ref{fig:intro}).  
Building upon BEV-based localization pipelines, we introduce two key modules:  
(i) a Surface Model Mechanism that explicitly estimates the vertically visible surface from the ground view, ensuring BEV features are sampled from geometrically valid and visually consistent regions; and  
(ii) a SimRefiner module that refines the similarity distribution by jointly modeling local spatial coherence and global correspondence context, producing more reliable matches and stable pose estimation.  
Together, these components enable geometry-consistent feature alignment and robust 3-DoF localization across drastically different perspectives.

Despite the importance of cross-view image matching, it remains underexplored in existing benchmarks. To address this gap, we introduce the first benchmark tailored for ground-to-satellite matching: the Cross-View Fine-grained image Match (CVFM) benchmark. 
CVFM is constructed by sampling ground-view images from the DReSS dataset~\cite{xia2025cross}, and generating dense pixel-level correspondences by projecting ground-view pixels onto satellite-view images using ground-truth depth maps from the DReSS-D~\cite{zhang2025cross} dataset. 
To ensure label quality, we manually select a high-fidelity subset to serve as a reliable benchmark. 
CVFM contains 32,509 high-quality cross-view image pairs for evaluating image matching algorithms under extreme viewpoint disparities, and facilitates future research in geometry-aware cross-view understanding.
To summarize, our main contributions are as follows:

\begin{itemize}
 \item We revisit cross-view localization from an image-matching perspective and present ViewBridge, a unified framework that jointly improves localization and matching under large viewpoint disparities, achieving state-of-the-art performance on public benchmarks.
\item  We introduce a Surface Model Mechanism that constrains BEV feature construction to physically valid surfaces, enabling semantically and geometrically consistent BEV representations. 
\item We develop SimRefiner, a learnable module that refines the similarity field by jointly modeling local and global correspondence context, improving match reliability and pose stability. 
\item We build CVFM, the first large-scale benchmark with pixel-level ground-satellite correspondences for fine-grained cross-view image matching, providing a standardized evaluation for future research.
\end{itemize}

%% file: sec/2_relatedwork.tex
\section{Related Work}
\label{sec:related_work}

\textbf{Cross-View Localization} has traditionally been approached as a large-scale image retrieval problem ~\cite{shi2019spatial,deuser2023sample4geo,wu2024camp,xia2024enhancing,xia2025cross,wu2024cross,lu2025gleam}, where the goal is to identify the most visually similar satellite-view image for a given ground-view image. More recently, research has shifted toward fine-grained estimation of the camera's spatial pose, enabling more precise localization beyond coarse retrieval~\cite{fervers2023uncertainty,lentsch2023slicematch,shi2022beyond,shi2022accurate,shi2023boosting,shi2024weakly,wang2023fine,xia2022visual,xia2023convolutional,xia2024adapting,xia2025fg,song2023learning,zhang2025cross}.

A major challenge in cross-view localization lies in the drastic viewpoint differences between ground and satellite imagery. One prominent line of work addresses this issue by directly regressing the 3-DoF pose (location and orientation) of the ground-view image using deep neural networks trained on cross-view image pairs~\cite{lentsch2023slicematch,xia2022visual,xia2023convolutional,zhang2025cross,xia2025fine}. These methods implicitly learn spatial correspondences and can exploit the full visual content, but they require the network to effectively model complex and diverse urban scenes. An alternative and increasingly popular approach leverages a shared representation in the form of Bird's Eye View (BEV), which captures a co-visible geometry between the ground and satellite views~\cite{fervers2023uncertainty,shi2023boosting,song2023learning,wang2023fine}. By transforming ground-view into the BEV space, the model can alleviate the extreme viewpoint gap. However, BEV representations also introduce challenges: ground images often suffer from occlusions caused by buildings and vegetation, while the BEV-transformed images may contain distortions or lose important visual cues. Recent advances~\cite{xia2025fg} have attempted to mitigate these issues by learning adaptive or view-aware BEV embeddings, aiming to preserve semantic consistency while maintaining the advantages of viewpoint alignment.

\textbf{Image Matching under Extreme Viewpoint Variations} is a highly related task, and can be viewed as a subproblem of general image matching. Traditional matching algorithms~\cite{sarlin2020superglue,sun2021loftr,edstedt2024roma} perform well when the viewpoint variation is small, such as matching different views of the same building from the ground. However, their performance degrades significantly when the images are captured from drastically different perspectives, such as ground versus satellite views. Recent advances~\cite{wang2024dust3r,leroy2024grounding,barroso2024matching,wang2025vggt} have introduced hybrid pipelines that jointly estimate point clouds alongside feature matching. These methods leverage the predicted 3D geometry to provide additional spatial cues, significantly improving matching robustness under large viewpoint changes. Furthermore, efforts~\cite{vuong2025aerialmegadepth} to expand existing datasets to include ground-aerial image pairs have shown to substantially enhance model performance on cross-view matching benchmarks.

Our method enables coarse ground-satellite image matching using only 3-DoF supervision during training, which is readily available in practice. The resulting coarse correspondences provide strong priors for downstream fine-grained matching and can significantly improve performance under extreme viewpoint changes.

%% file: sec/3_method.tex
\section{Methodology}

\begin{figure*}[t]
    \centering
    \includegraphics[width=1.0\linewidth]{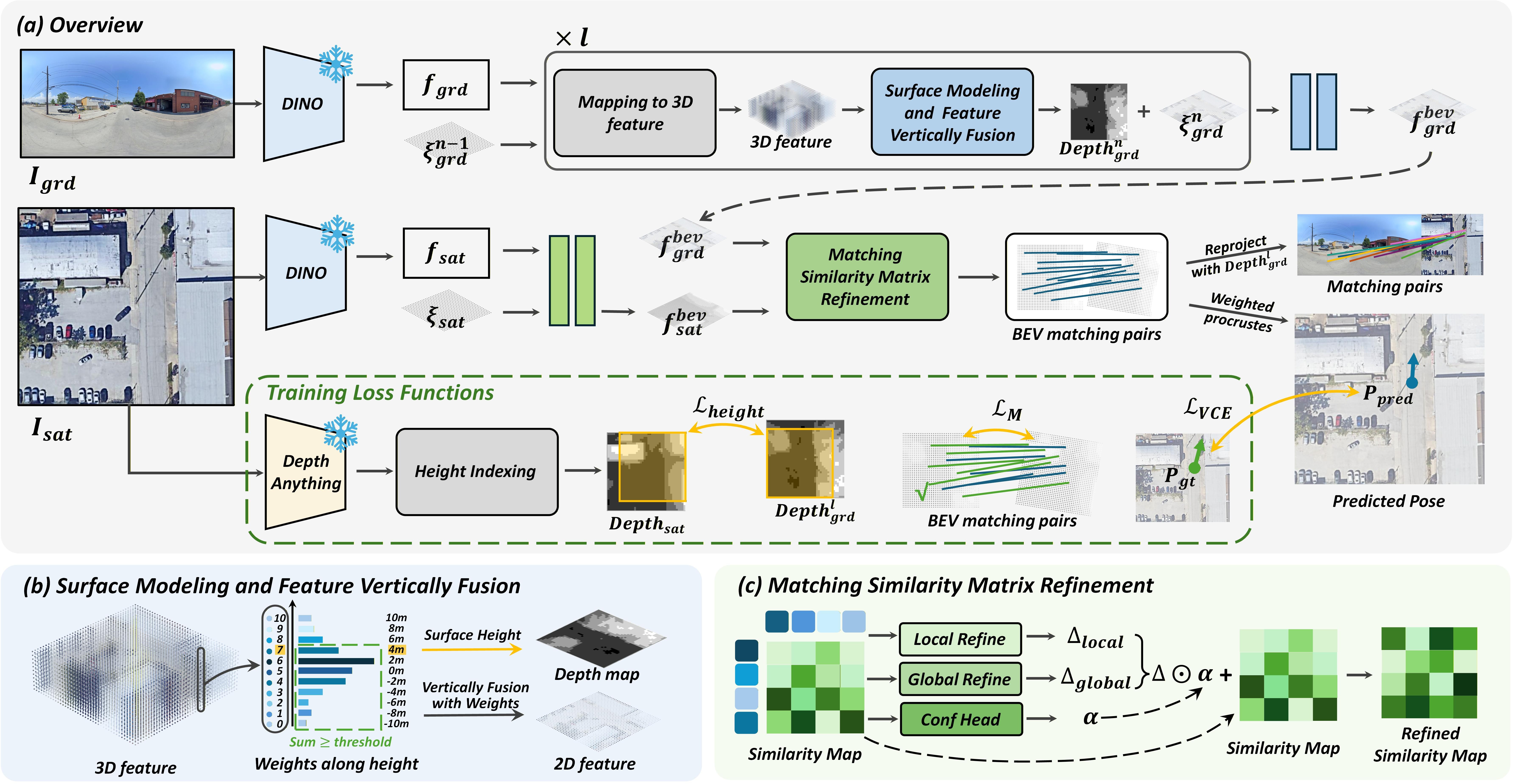}
   \caption{
Illustration of our framework.  
The model constructs BEV feature representations \( f_{\text{grd}}^{\text{bev}} \) and \( f_{\text{sat}}^{\text{bev}} \) by projecting visual information from ground-view and satellite-view images into learnable BEV point sets \(\xi_{\text{grd}}\) and \(\xi_{\text{sat}}\). These BEV features are then matched in BEV space to estimate accurate 3-DoF poses, and the predicted depth reprojects matched points back into the image plane for pixel-level correspondence. 
}
    \label{fig:main_framework}
    \vspace{-1.0em}
\end{figure*}

Given a ground-view image \( I_{\text{grd}} \) and an satellite-view image \( I_{\text{sat}} \) that spatially covers the same region, the goal is to estimate the 3-DoF camera pose \( p = [\text{translation}, \text{orientation}] \). 
Here, the translation corresponds to the camera location in the pixel coordinates of \( I_{\text{sat}} \), and the orientation denotes the yaw angle of the ground-view camera.

\subsection{Overview and Motivation}

\textbf{Overview.}  
As illustrated in Fig.~\ref{fig:main_framework}, our framework consists of three stages: BEV feature construction, BEV feature matching, and a final stage for pose estimation and pixel-level matching.
Given a ground-view panorama \(I_{\text{grd}}\) and an satellite-view image \(I_{\text{sat}}\), we first extract image-space features \(f_{\text{grd}}\) and \(f_{\text{sat}}\) using a frozen DINOv2~\cite{oquab2023dinov2} backbone, followed by view-specific encoders that transform them into a unified BEV space. 
For the ground view, we first build a voxel grid around the camera and map ground-view feature into this 3D volumetric space by projecting each voxel onto the image plane, where deformable attention samples the relevant visual cues. 
The \textit{Surface Model} (Sec.~\ref{sec:surface_model}) then predicts visible surface heights and aggregates valid voxels into a compact 2D BEV map \(f_{\text{grd}}^{\text{bev}}\). For the satellite view,  \(f_{\text{sat}}\) is directly projected onto a 2D BEV grid to obtain \(f_{\text{sat}}^{\text{bev}}\). Next, a dense similarity matrix \(S_{\text{orig}}\) is computed between \(f_{\text{grd}}^{\text{bev}}\) and \(f_{\text{sat}}^{\text{bev}}\) for cross-view BEV matching, which is refined by the \textit{SimRefiner} module (Sec.~\ref{sec:simrefiner}) to model local spatial coherence and global correspondence context. Finally, high-confidence correspondences are fed into a weighted Procrustes solver to estimate the 3-DoF pose \(P_{\text{pred}} = [R, T]\).  
For cross-view image matching, the predicted ground-view depth map \( \text{Depth}^{l}_{\text{grd}} \) is used to reproject BEV correspondences into the image plane, establishing pixel-level correspondences between ground and satellite views. 
The entire framework is trained end-to-end \textbf{using only 3-DoF supervision} with three loss functions (Sec.~\ref{sec:lossfunctions}):  
the virtual correspondence error loss \( \mathcal{L}_{\text{VCE}} \) for pose alignment,  
the matching loss \( \mathcal{L}_{\text{M}} \) for correspondence supervision,  
and the height consistency loss \( \mathcal{L}_{\text{height}} \) for geometric regularization. 

\textbf{Motivation.}  
Recent works estimate the 3-DoF pose by establishing cross-view correspondences in BEV space, constructing learnable BEV features through deformable attention to bridge the large viewpoint gap. 
However, correspondence reliability remains constrained by two major challenges: geometric ambiguity in ground-view BEV construction and limited modeling capacity for the similarity distribution.

(i) \textit{Ground-view BEV construction.}  
Projecting 3D sampling points of different heights into the ground-view plane requires identifying the true visible surface along each vertical ray. Prior approaches simply select the layer with the maximum attention weight, which lacks physical grounding and often leads to incorrect height selection (e.g., points floating in mid-air that do not lie on any physical surface, whose features are essentially inherited from incorrect spatial locations), thereby mixing features from semantically inconsistent regions. We propose a Surface Model Mechanism that adaptively estimates surface height through bottom-up accumulation of height-wise attention and thresholded volumetric integration, inspired by volume rendering~\cite{mildenhall2021nerf,yue2025nerfortho,fan20253d,liu2025stereoinr}. 
This mechanism grounds BEV construction on physical surfaces, ensuring that features are sampled from geometrically valid regions.

(ii) \textit{Similarity distribution modeling.}  
The distribution of the similarity matrix plays a decisive role in determining the stability of the final pose estimation. 
Existing frameworks usually treat it as a fixed outcome, for example by directly selecting top-scoring pairs or applying simple dustbin reweighting, which overlooks the structural information embedded in the distribution itself. 
We introduce SimRefiner, a learnable refinement module that enhances the network’s ability to model match reliability by jointly capturing local spatial regularity and global relational structure. 
Through this learned reasoning over the similarity distribution, the model achieves more geometry-consistent correspondences and consequently more robust localization.

\subsection{Surface Model Mechanism}
\label{sec:surface_model}

To enable accurate ground-to-BEV projection, we introduce a Surface Model Mechanism that predicts the visible surface height for each BEV cell, ensuring geometric accuracy and semantic consistency.

\textbf{Ground-view BEV encoding.}  
Following previous works~\cite{li2024bevformer,xia2025fg}, we construct a uniform 3D voxel grid centered at the ground-view camera. 
Each voxel corresponds to a sampling point \((x, y, z)\), where the height dimension is discretized into \(M\) layers (for example, \(M=11\) covering \([-10\,\text{m}, 10\,\text{m}]\)). Each voxel is geometrically projected onto the ground-view feature map \( f_{\text{grd}} \), where deformable attention~\cite{zhu2020deformable} samples multi-scale visual features to form a volumetric tensor \( \xi_{\text{grd}}^{3D} \in \mathbb{R}^{M \times N \times N \times C} \).

Instead of simply selecting the height layer with the maximum activation, we infer the visible surface through a continuous accumulation process inspired by volume rendering. 
A per-voxel confidence score is learned along the height axis, and its cumulative distribution from bottom to top indicates the likelihood of encountering a visible surface. 
The first layer where the cumulative value surpasses a learnable threshold is treated as the surface layer. 
Features along the height dimension are then fused in a confidence-weighted manner to obtain the final ground-view BEV feature \( f_{\text{grd}}^{\text{bev}} \in \mathbb{R}^{N \times N \times c} \), followed by a projection head for dimensional alignment. 
This physically grounded mechanism ensures that each BEV cell aggregates information from the most plausible surface region in the ground image.

\textbf{Satellite-view BEV encoding and height indexing.}  
For the satellite view, which is inherently top-down, we directly sample features onto a 2D BEV grid \( \xi_{\text{sat}} \) using the known ground sampling distance (GSD), followed by a lightweight projection network to obtain \( f_{\text{sat}}^{\text{bev}} \in \mathbb{R}^{N \times N \times c} \).

To provide weak supervision for surface estimation, we use a pretrained monocular depth estimator (DepthAnything v2-small~\cite{yang2024depth}) to predict a relative depth map \( \text{depth}_{\text{sat}} \) from the satellite image. 
As this prediction lacks an absolute metric scale, we apply a simple yet effective prior: the ground is assumed to be approximately planar, and the ground-view camera is positioned at a typical height of \(2\)-\(3\,\text{m}\). 
Under this assumption, the minimum depth value in \( \text{depth}_{\text{sat}} \) is anchored to \(-3\,\text{m}\), treated as the local ground level, and the remaining values are linearly scaled to approximate real-world heights. 
Although this mapping is approximate due to uncertainties in monocular estimation and camera placement, it provides a practical pseudo-supervision signal for guiding the ground-view surface prediction \( \text{depth}_{\text{grd}} \). 
This supervision encourages the Surface Model to learn height representations that are both semantically meaningful and spatially grounded.

Overall, the proposed Surface Model Mechanism introduces learnable and physically interpretable height modeling into BEV construction, ensuring that ground-view features are projected from geometrically credible regions and leading to more stable and consistent cross-view alignment.

\subsection{Matching Similarity Matrix Refinement}
\label{sec:simrefiner}

To enhance match reliability modeling, we propose SimRefiner, a module that refines the similarity matrix by capturing both local spatial coherence and global correspondence structure, leading to more stable similarity distributions and improved pose estimation.

\textbf{Initial similarity matrix.}  
Given BEV features \( f_{\text{grd}}^{\text{bev}} \) and \( f_{\text{sat}}^{\text{bev}} \in \mathbb{R}^{N \times N \times c} \), we flatten them into patch sequences and compute an initial similarity matrix using scaled cosine similarity:

\begin{equation}
S_{\text{orig}} = \frac{f_{\text{grd}}^{\text{bev}} \cdot {f_{\text{sat}}^{\text{bev}}}^\top}{\tau} \in \mathbb{R}^{N^2 \times N^2},
\label{eq:initial_similarity}
\end{equation}

where \( \tau \) is a temperature parameter and \(N = 41\). Each row of \( S_{\text{orig}} \) represents the similarity scores between one ground-view patch and all satellite-view patches.

\textbf{Dual-branch residual correction.}  
We design a dual-branch refinement mechanism to model both fine-grained local dependencies and long-range global relationships within the similarity space.

(i) Local refinement.  
We reshape \( S_{\text{orig}} \) into a 3D similarity cube \( S_{\text{cube}} \in \mathbb{R}^{N \times N \times N^2} \), where each ground-view position stores its complete similarity vector with the satellite view. 
A compact 3-layer 3D convolutional network extracts spatially-aware residuals:
\begin{equation}
\Delta_{\text{local}} = \text{Conv3D}(S_{\text{cube}}) \in \mathbb{R}^{N^2 \times N^2}.
\end{equation}
This branch enforces local structural regularity, ensuring that nearby BEV patches exhibit consistent similarity patterns.

(ii) Global refinement.  
In parallel, a row-wise MLP is applied to \( S_{\text{orig}} \) to capture higher-order global correspondence patterns:
\begin{equation}
\Delta_{\text{global}} = \text{MLP}(S_{\text{orig}}) \in \mathbb{R}^{N^2 \times N^2}.
\end{equation}
This branch provides global relational reasoning, allowing the model to contextualize each match within the broader similarity landscape.

\textbf{Residual fusion with soft confidence gating.}  
The two correction terms are fused through a soft confidence gating mechanism:
\begin{equation}
S_{\text{refined}} = S_{\text{orig}} + \boldsymbol{\alpha} \odot (\Delta_{\text{local}} + \Delta_{\text{global}}),
\end{equation}
where \(\boldsymbol{\alpha} \in [0,1]^{N^2 \times N^2}\) is a confidence map predicted by a lightweight MLP.  
Each row of \(\boldsymbol{\alpha}\) shares a single confidence value corresponding to the reliability of the query patch,  
which is broadcast across all its candidate matches.  
This soft gating adaptively adjusts the refinement strength according to the confidence of each query patch, suppressing unstable corrections while retaining the reliable ones.

\textbf{Adaptive dustbin and normalization.}  
To accommodate unmatched regions, we append a learnable dustbin row and column:
\begin{equation}
S_{\text{dustbin}} =
\begin{bmatrix}
S_{\text{refined}} & \mathbf{b}_{\text{col}} \\
\mathbf{b}_{\text{row}}^\top & b_{\theta}
\end{bmatrix} \in \mathbb{R}^{(N^2+1) \times (N^2+1)}.
\end{equation}
Doubly-stochastic normalization is then applied via row-wise and column-wise softmax:
\begin{equation}
S = \left[ \text{Softmax}_{\text{row}}(S_{\text{dustbin}}) \cdot \text{Softmax}_{\text{col}}(S_{\text{dustbin}}) \right]_{1:N^2, 1:N^2},
\label{eq:final_matching}
\end{equation}
resulting in the final refined similarity matrix \( S \), which provides both smooth probabilistic matching and reliable input for downstream pose estimation.

Overall, SimRefiner enhances the model’s ability to reason over the structure of the similarity distribution, converting raw similarity scores into spatially coherent and geometrically reliable correspondences.

\subsection{Loss Functions}
\label{sec:lossfunctions}
The framework is trained end-to-end with a composite loss.  
Following FG2~\cite{xia2025fg}, we adopt the Virtual Correspondence Error (VCE) loss \(\mathcal{L}_{\text{VCE}}\) to supervise pose estimation and an InfoNCE-based matching loss \(\mathcal{L}_{\text{M}}\) to enhance feature alignment.  
Detailed formulations of both terms follow~\cite{xia2025fg} and are provided in the \textbf{supplementary material}.

To regularize the predicted surface geometry, we introduce a cross-view height consistency constraint.  
Let \(f_{\text{grd}}^{\text{bev}}\) and \(f_{\text{sat}}^{\text{bev}}\) denote BEV features derived from ground-view and satellite-view images, respectively.  
Each BEV cell \(n_{\text{grd}}\) in \(f_{\text{grd}}^{\text{bev}}\) is projected into \(f_{\text{sat}}^{\text{bev}}\) using the ground-truth transformation \(P_{\text{gt}}\), obtaining its corresponding coordinate \(n_{\text{sat}}'\).  
From the predicted surface maps \(\text{depth}_{\text{grd}}(\cdot)\) and \(\text{depth}_{\text{sat}}(\cdot)\),  
the loss minimizes the depth discrepancy between these cross-view correspondences:
\begin{equation}
\mathcal{L}_{\text{height}} =
\frac{1}{N_s} \sum_{i=1}^{N_s}
\frac{1}{K}
\Big|
\text{depth}_{\text{grd}}(n_{\text{grd}}) -
\text{depth}_{\text{sat}}(n_{\text{sat}}')
\Big|,
\label{eq:loss_height_main}
\end{equation}
where \(N_s\) is the number of sampled correspondences and \(K = 100\) normalizes the scale for stable optimization.  

The overall training objective is:
\begin{equation}
\mathcal{L} =
\mathcal{L}_{\text{VCE}} +
\mathcal{L}_{\text{M}} +
\mathcal{L}_{\text{height}}.
\label{eq:loss_all_main}
\end{equation}

%% file: sec/4_experiments.tex
\section{Experiments}

\subsection{Datasets and Evaluation Protocols}

We conduct extensive experiments on two mainstream public cross-view localization datasets and our introduced fine-grained cross-view image matching benchmark.

\textbf{VIGOR}~\cite{zhu2021vigor} provides ground-view panoramas and satellite images from four major U.S. cities. 
It defines two evaluation protocols: (i) \textit{same-area}, where both training and testing samples are drawn from all cities; and (ii) \textit{cross-area}, where training is performed on two cities and testing on the remaining two. 
Since no official validation split is available, we randomly assign 20\% of the training samples for validation and use the remaining 80\% for training.

\textbf{DReSS-D}~\cite{zhang2025cross} is a reprocessed and geometrically aligned version of DReSS~\cite{xia2025cross}, containing paired ground-view panoramas and satellite images from six globally distributed cities. 
It follows the same evaluation protocols as VIGOR, and we adopt the same 80\%-20\% training-validation division within the official training set.

\begin{figure}[t]
    \centering
    \includegraphics[width=1\linewidth]{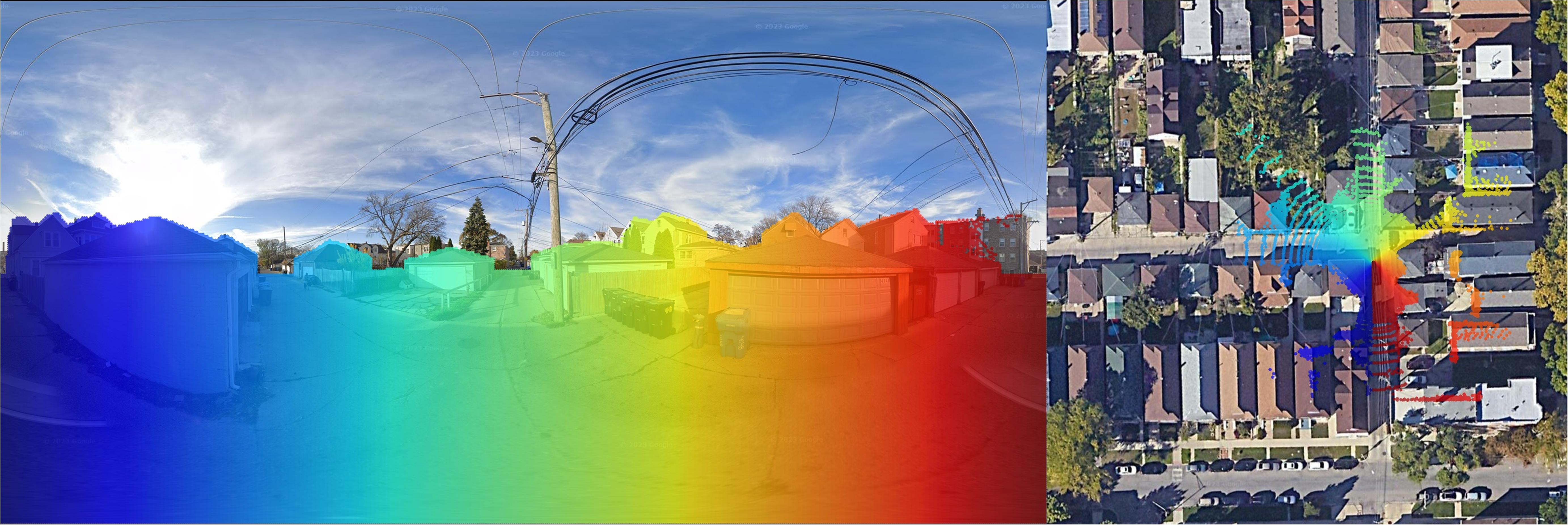}
    \caption{
Projection of ground-view pixels onto the satellite image in the CVFM benchmark. 
Colors represent pixel-wise correspondences across views.
}
    \label{fig:exp_CVFM_dataset}
    \vspace{-1.4em}
\end{figure}

\begin{table*}[t]
    \centering
    \caption{
    VIGOR test results. FG2$^\dagger$ denotes our re-implementation, which achieves better performance than the originally reported results. \textbf{RANSAC results for FG2 and Ours are provided in the supplementary material}. \textbf{Best} and \underline{second-best} results are highlighted.
}
    \begin{tabular}{p{1.4cm}p{2.25cm}p{1.2cm}p{1.2cm}p{1.2cm}p{1.2cm}|p{1.2cm}p{1.2cm}p{1.2cm}p{1.2cm}}
    \toprule
    \multirow{3}{*}{{Orien.}} & \multirow{3}{*}{Methods} & 
    \multicolumn{4}{c}{Same-area} & \multicolumn{4}{c}{Cross-area} \\
    \cline{3-10} 
    & & \multicolumn{2}{c}{Localization (m) $\downarrow$} & \multicolumn{2}{c}{Orientation ($^\circ$) $\downarrow$} & \multicolumn{2}{c}{Localization (m) $\downarrow$} & \multicolumn{2}{c}{Orientation ($^\circ$) $\downarrow$} \\
    \cline{3-10} 
    & & Mean & Median & Mean & Median & Mean & Median & Mean & Median \\
    \hline
    \multirow{6}{*}{Known}& 
    SliceMatch~\cite{lentsch2023slicematch}& 5.18 & 2.58 & - & - & 5.53 & 2.55 & - & - \\
    & CCVPE~\cite{xia2023convolutional}& 3.60 & 1.36 & - & - & 4.97 & 1.68 & - & - \\
    & DenseFlow~\cite{song2023learning}& 3.03 & \textbf{0.97}& - & - & 5.01 & 2.42 & - & - \\
    & HC-Net~\cite{wang2023fine}& 2.65 & 1.17 & - & - & 3.35 & 1.59& - & - \\
 & FG2$^\dagger$~\cite{xia2025fg}
& \underline{2.10}& \underline{1.03}& -& -& \underline{2.73}& \underline{1.49}& -&-\\
 & Ours& \textbf{1.89}& \textbf{0.97}& -& -& \textbf{2.33}& \textbf{1.34}& -&-\\

    \hline
    \multirow{5}{*}{Unknown}& 
    SliceMatch~\cite{lentsch2023slicematch}& 6.49 & 3.13 & 25.46 & 4.71 & 7.22 & 3.31 & 25.97 & 4.51 \\
    & CCVPE~\cite{xia2023convolutional}& \underline{3.74}& \underline{1.42}& 12.83 & 6.62 & 5.41 & \textbf{1.89} & 27.78 & 13.58 \\
    & DenseFlow~\cite{song2023learning}& 4.97 & 1.90 & 11.20 & \textbf{1.59} & 7.67 & 3.67 & 17.63 & \underline{2.94}\\
 & FG2$^\dagger$~\cite{xia2025fg}& 3.90& 2.22& \textbf{7.07}& \underline{1.83}& \underline{4.96}& 2.76& \textbf{11.07}&\textbf{2.40}\\
    & Ours& \textbf{3.11}& \textbf{1.31}& \underline{8.65}& 2.69& \textbf{4.19}& \underline{2.26}& \underline{12.56}& 3.05\\
    \bottomrule
    \end{tabular}
    \label{tab:vigor}
    \vspace{-0.5em}
\end{table*}

\begin{table*}[ht]
    \centering
    \caption{
     DReSS-D test results. FG2$^\dagger$ denotes our re-implementation. \textbf{Best} and \underline{second-best} results are highlighted.
    }
    \begin{tabular}{p{1.4cm}p{2.25cm}p{1.2cm}p{1.2cm}p{1.2cm}p{1.2cm}|p{1.2cm}p{1.2cm}p{1.2cm}p{1.2cm}}
    \toprule
    \multirow{3}{*}{{Orien.}} & \multirow{3}{*}{Methods} & 
    \multicolumn{4}{c}{Same-area} & \multicolumn{4}{c}{Cross-area} \\
    \cline{3-10} 
    & & \multicolumn{2}{c}{Localization (m) $\downarrow$} & \multicolumn{2}{c}{Orientation ($^\circ$) $\downarrow$} & \multicolumn{2}{c}{Localization (m) $\downarrow$} & \multicolumn{2}{c}{Orientation ($^\circ$) $\downarrow$} \\
    \cline{3-10} 
    & & Mean & Median & Mean & Median & Mean & Median & Mean & Median \\
    \hline
    \multirow{4}{*}{Known}& CCVPE~\cite{xia2023convolutional}
& 2.83& 0.98& -& -& 5.91& 1.87& -& -\\
    & FG2$^\dagger$~\cite{xia2025fg}& \underline{1.72}&  \underline{0.76}& -& -& \underline{2.68}& \underline{1.50}& -& -\\
    & Slice-Loc~\cite{zhang2025cross}& 2.10& 0.82& -& -& 3.99& 1.52& -& -\\
 & Ours& \textbf{1.60}& \textbf{0.73}& -& -& \textbf{2.65}& \textbf{1.40}& -&-\\
    \hline
    \multirow{3}{*}{Unknown}& 
    CCVPE~\cite{xia2023convolutional}
& \underline{3.01}& \textbf{1.02}& 14.44& 7.94& 6.05& \textbf{2.23}& 37.39& 10.27\\
    & FG2$^\dagger$~\cite{xia2025fg}& 4.09& 1.98& \textbf{8.16}& \textbf{1.54}& \underline{5.19}& 3.10& \textbf{10.13}& \textbf{2.14}\\
    & Ours& \textbf{2.94}& \underline{1.49}& \underline{8.47}& \underline{1.62}& \textbf{4.22}& \underline{2.37}& \underline{12.58}& \underline{2.60}\\
    \bottomrule
    \end{tabular}
    \label{tab:dress-d}
    \vspace{-1.0em}
\end{table*}

\textbf{CVFM Benchmark} is our newly proposed benchmark for fine-grained cross-view image matching between ground and satellite perspectives, offering the first large-scale evaluation resource with dense pixel-level correspondences. 
All images are sourced from DReSS~\cite{xia2025cross}, while ground-truth correspondences are generated by projecting 3D points from the ground view into the satellite domain using DReSS-D~\cite{zhang2025cross} depth supervision (see Fig.~\ref{fig:exp_CVFM_dataset}). 
To ensure that projections fall within valid satellite coverage, we apply a \(30\,\mathrm{m}\) spatial threshold to filter out-of-bounds pixels, and define the remaining region as the \textbf{valid area}. 
Finally, three graduate students with expertise in computer vision rechecked all candidate pairs to determine their suitability, resulting in a high-quality benchmark comprising 32,509 image pairs. 

\textbf{Evaluation Metrics:} For cross-view localization, we report both mean and median errors in translation and orientation under two configurations: (i) \textit{known orientation}, where ground-view panoramas are aligned northward, and (ii) \textit{unknown orientation}, where the heading is arbitrary. For fine-grained image matching, we evaluate the percentage of correct correspondences under different pixel error thresholds on the \textit{satellite image}. Each method is expected to output a set of matched coordinate pairs, denoted as \([(x_\text{grd}, y_\text{grd}) \leftrightarrow (x_\text{sat}, y_\text{sat})]^{N_{mch}}\). We compute the percentage of correct matches among \(N_{mch}\) matches under thresholds of 5, 10, and 15 pixels based on the ground truth projection.

\begin{figure*}[ht]
    \centering
    \includegraphics[width=1.0\linewidth]{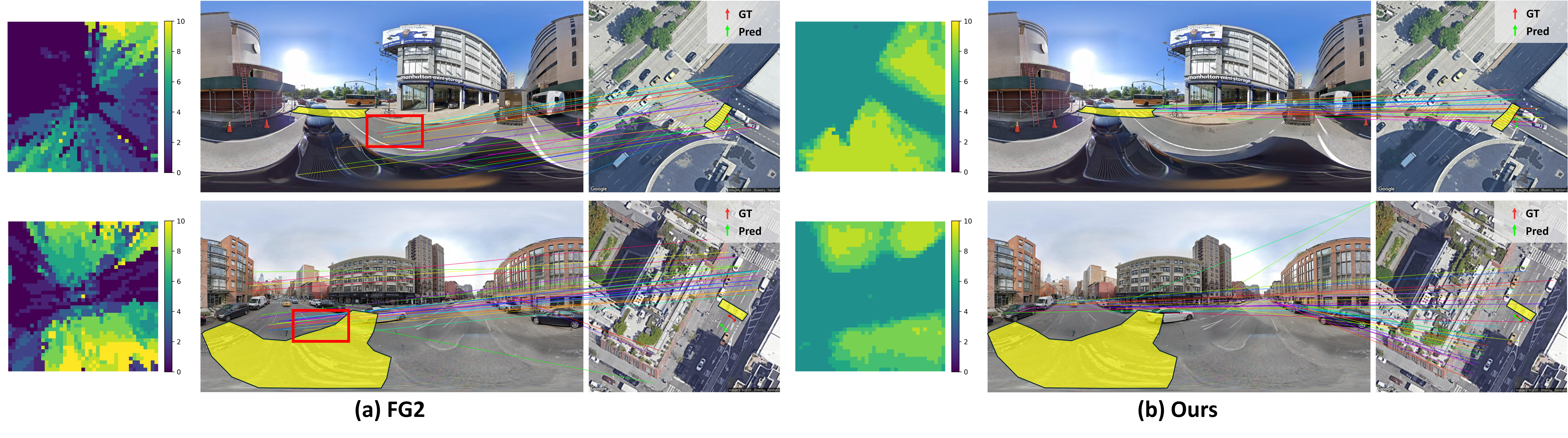}
    \vspace{-2.0em}
    \caption{Visualization of cross-view localization on VIGOR. Corresponding objects are highlighted with consistent colors for visual clarity. Yellow lines indicate predicted matches. Since VIGOR does not provide pixel-level ground-truth correspondences, correctness cannot be assessed directly; instead, obvious mismatches are marked with red circles for qualitative comparison.}
    \label{fig:exp_vis}
    \vspace{-1.0em}
\end{figure*}

\subsection{Implementation Details}
On both the VIGOR and DReSS-D datasets, we define \(41 \times 41\) grid points for \(\xi_{\text{grd}}\) and \(\xi_{\text{sat}}\), uniformly distributed over a \(71\,\text{m} \times 71\,\text{m}\) area centered at the camera. Along the vertical axis, we use \(M = 11\) height layers ranging from \(-10\,\text{m}\) to \(10\,\text{m}\). Following~\cite{li2024bevformer, xia2025fg}, the ground-view BEV encoder is composed of \(l = 6\) stacked layers.
We set the temperature parameter in Eq.~\ref{eq:initial_similarity} to \(\tau = 0.1\). 
For the Virtual Correspondence Error loss, we randomly sample \(N_v = 100\) 2D points from a BEV region of size \(L_v = 5\,\text{m}\). 
For the matching loss and height loss (Eq.~\ref{eq:loss_height_main}), the number of sampled patch pairs is set to \(N_s = 1024\).
All models are trained using the AdamW optimizer with a learning rate of \(1 \times 10^{-4}\), a batch size of 48, and run for 100 epochs on 8 NVIDIA Tesla V100 GPUs.

\subsection{Results on cross-view localization}

\textbf{Quantitative results.}  
We compare our method with recent state-of-the-art cross-view localization approaches on the VIGOR and DReSS-D benchmarks, as summarized in Tab.~\ref{tab:vigor} and Tab.~\ref{tab:dress-d}.  
In the \textit{Known Orientation} configuration, our method consistently achieves the lowest mean and median localization errors on both datasets.  
These results demonstrate that the proposed modules substantially enhance cross-view feature alignment, enabling more accurate 3-DoF pose estimation.  
Moreover, the small performance gap between same-area and cross-area settings highlights the robustness and generalization ability of our framework for real-world deployment.
In the more challenging \textit{Unknown Orientation} setting, our method again attains the best localization accuracy, outperforming the strongest baseline (FG2) by up to 0.8\,m on VIGOR and 1.0\,m on DReSS-D in mean translation error.  
For orientation estimation, our model maintains competitive accuracy, trailing FG2 by less than \(3^\circ\) on average while achieving the lowest combined localization-orientation error.  
It is worth noting that our re-implementation of FG2 (denoted as FG2$^\dagger$) achieves significantly better results than those originally reported; all comparisons in Tab.~\ref{tab:vigor} and Tab.~\ref{tab:dress-d} are therefore made against this stronger baseline to ensure fairness. 
Additional RANSAC-refined results for FG2 and our method are provided in the \textbf{supplementary material}.

\textbf{Qualitative results.}  
Fig.~\ref{fig:exp_vis} provides a visual comparison between FG2 and our method on the VIGOR dataset.  
For each sample, the leftmost column depicts the estimated height map from the ground-view image, where indices 0-10 correspond to real-world heights from \(-10\,\mathrm{m}\) to \(10\,\mathrm{m}\).  
As shown in Fig.~\ref{fig:exp_vis}(a), inaccurate height selection in FG2 causes mismatches highlighted in red circles, where BEV grids mistakenly sample features from non-visible or semantically incorrect regions.  
Our method, by explicitly estimating the visible surface height, ensures that each BEV grid gathers features from physically valid regions, leading to geometrically consistent BEV representations and more stable pose predictions.  
The color-marked corresponding regions in Fig.~\ref{fig:exp_vis} make it clear that our method maintains more consistent alignment between the ground and satellite views. More visualizations are provided in the \textbf{supplementary material}.

\subsection{Results on cross-view image matching}

\textbf{Quantitative results.}  
We evaluate our method on the proposed CVFM benchmark against existing approaches capable of cross-view matching. 
LoFTR~\cite{sun2021loftr}, SuperGlue~\cite{sarlin2020superglue}, and RoMA~\cite{edstedt2024roma} are representative state-of-the-art image matching methods. 
Aerial-MD~\cite{vuong2025aerialmegadepth} is a fine-tuned variant of Mast3r~\cite{leroy2024grounding}, trained on the aerial-MegaDepth dataset for cross-view adaptation. 
For evaluation, we follow the official setup of Aerial-MD by projecting each panoramic image into three views with a \(120^\circ\) field of view, performing pairwise matching for each projection, and selecting the top matches across all trials.
FG2~\cite{xia2025fg} serves as a recent BEV-based localization framework that implicitly learns cross-view correspondences during training.
All compared methods use their released models, while our method is trained on the VIGOR dataset. \textbf{None of the methods are trained on CVFM}, and all are required to predict \(N_{\text{mch}} = 30\) correspondences during evaluation.
The final column in Tab.~\ref{tab:cvfm} reports the proportion of matches located within the valid region defined by ground-truth projections. As shown in Tab.~\ref{tab:cvfm}, our method achieves substantial improvements across all error thresholds, demonstrating its ability to establish geometrically consistent and semantically meaningful correspondences across diverse viewpoints. Additionally, Aerial-MD ranks second, largely due to its pretraining on the aerial-MegaDepth dataset, which contains ground-aerial image pairs with pixel-level correspondences. This dataset spans a wider cross-view range than conventional image matching datasets, allowing better learning of geometric relationships across perspectives. Additional results on dense matching are provided in the \textbf{supplementary material}.

\begin{table}[t]
    \centering
    \caption{CVFM cross-view image matching results (measured on the satellite image) using the top-30 matches. 1 pixel equals approximately \(0.12\,\mathrm{m}\). \textbf{Best in bold.}}
    \begin{tabular}{p{2.3cm}P{0.95cm}P{0.95cm}P{0.95cm}P{0.95cm}}
    \toprule
    Method & $\le$5px$\uparrow$ & $\le$10px$\uparrow$ & $\le$15px$\uparrow$ & Valid\\
    \hline
    LoFTR~\cite{sun2021loftr}& 0.0\%& 0.1\%& 0.3\%&0.692\\
 SuperGlue~\cite{sarlin2020superglue}& 0.0\%& 0.2\%& 0.3\%&0.228\\
 RoMa~\cite{edstedt2024roma}& 0.1\%& 0.4\%& 0.8\%&0.778\\
 Aerial-MD~\cite{vuong2025aerialmegadepth}& 1.5\%& 5.4\%& 9.5\%&0.422\\
    FG2~\cite{xia2025fg}& 0.6\%& 2.3\%& 5.0\%&0.997\\
    Ours& \textbf{3.1\%}& \textbf{11.7\%}& \textbf{24.0\%}&0.915\\
    \bottomrule
    \end{tabular}
    \label{tab:cvfm}
    \vspace{-1.7em}
\end{table}

\begin{figure*}[ht]
    \centering
    \includegraphics[width=1.0\linewidth]{figures/4_experiments/CVFM_exp.jpg}
    \vspace{-2.0em}
    \caption{Cross-view image matching results on the CVFM benchmark. Green and red lines denote correct and incorrect matches.}
    \label{fig:exp_vis_cvfm}
    \vspace{-1.0em}
\end{figure*}

\textbf{Qualitative results.}  
Fig.~\ref{fig:exp_vis_cvfm} qualitatively illustrates the advantage of our approach on the CVFM benchmark.  
Cross-view image matching remains highly challenging due to extreme viewpoint differences and partial visibility.
While previous methods such as FG2 project features into BEV space and attempt direct correspondence search, the lack of geometric grounding makes their matches visually plausible but physically inconsistent.  
In contrast, our method explicitly constrains correspondence formation through surface modeling, ensuring that matches originate from geometrically valid regions in both views.  
As shown in Fig.~\ref{fig:exp_vis_cvfm}, we visualize the top-30 predicted correspondences, where green and red lines denote correct and incorrect matches respectively, determined by a 15-pixel threshold in the ground-truth projection.  
This geometric constraint substantially improves the physical correctness of correspondences and markedly reduces false positives, demonstrating that our model captures a more faithful and interpretable cross-view relationship.  
Additional visual comparisons of different matching methods are provided in the \textbf{supplementary material}.

\begin{table}[t]
    \centering
    \caption{Ablations on proposed modules in VIGOR Known orientation setting. \textbf{Best in bold.}}
    \begin{tabular}{p{3.0cm}cc}
    \toprule
    Methods& Localization (m) & Matching\\
    \hline
        \textbf{\textit{Surface mechanism}}& Mean / Median& $\le$15px\\
 Base& 2.10 / 1.03& 5.0\%\\
    Surface Model& 2.03 / 1.05& 20.2\%\\
    \hline
 \textbf{\textit{Similarity refinement}}& Mean / Median& $\le$15px\\
 Base& 2.10 / 1.03& 5.0\%\\
 SimRefiner& 2.03 / 1.01& 10.5\%\\
 \hline
    Ours  & \textbf{1.89} / \textbf{0.97}& \textbf{24.0\%}\\
    \bottomrule
    \end{tabular}
    \label{tab:ablation}
    \vspace{-1.0em}
\end{table}

\subsection{Ablation study}
We conduct ablation studies on localization and matching using the VIGOR and CVFM datasets to analyze the contribution of each proposed component, with results summarized in Tab.~\ref{tab:ablation}. 
The baseline follows the standard BEV-based localization pipeline without our proposed modules.

Incorporating the Surface Model improves both localization and matching performance by constraining BEV sampling to geometrically valid regions, enhancing feature quality, and providing physically consistent surfaces for establishing cross-view correspondences. 
Adding the SimRefiner further boosts performance by enforcing spatial coherence in the similarity field and suppressing unreliable matches. 
The two modules operate at different stages of the pipeline (BEV construction and BEV matching), their effects are complementary, and jointly achieve the highest accuracy. 
Ablations on the \textbf{design choices within the Surface Model and SimRefiner} are provided in the \textbf{supplementary material}.

%% file: sec/5_conclusion.tex
\section{Conclusion}

We revisited cross-view localization from an image-matching perspective, emphasizing that accurate cross-view correspondence is essential for localization under large viewpoint disparities. 
To address this challenge, we proposed ViewBridge, a unified framework that integrates two complementary modules: the Surface Model, which grounds BEV construction on estimated surface geometry, and the SimRefiner, which refines similarity distributions to enhance match reliability through local and global consistency. 
Trained only with 3-DoF localization supervision, ViewBridge jointly improves localization accuracy and cross-view matching robustness. 
Beyond the method, we introduced CVFM, the first large-scale benchmark with manually verified pixel-level ground-satellite cross-view correspondences, enabling systematic evaluation of fine-grained matching algorithms. 
Although our framework is not a traditional image-matching method, it can still generate reliable cross-view correspondences that facilitate cross-view alignment and semantic understanding. More importantly, we believe this study will inspire broader advances in cross-view perception and contribute to future research toward unified, geometry-aware scene understanding across diverse viewpoints.

%% file: sec/X_suppl.tex
\clearpage
\setcounter{page}{1}
\setcounter{table}{0} 
\setcounter{figure}{0}
\setcounter{equation}{0}
\maketitlesupplementary
\appendix

In this supplementary material, we present additional technical details, analyses, and experiments that further support the contributions of the main paper:
\begin{enumerate}
    \item Section \ref{sup:sec:CVFM} provides a detailed description of the CVFM benchmark, including data collection, annotation procedures, and pairing strategy.
    \item Section \ref{sup:sec:loss} elaborates on the loss functions used in training and clarifies several implementation details omitted from the main paper.
    \item Section \ref{sup:sec:module} offers comprehensive ablations on our module design. It includes a detailed examination of component choices in Section \ref{sup:subsec:moduledesign} and additional investigations of the Surface Model in Section \ref{sup:subsec:surface}.
    \item Section \ref{sup:sec:ransac} presents extended evaluations with RANSAC-based refinement.
    \item Section \ref{sup:sec:orientation} analyzes orientation errors and discusses their influence on localization performance.
    \item Section \ref{sup:sec:matching} contains additional studies on cross-view image matching, including the influence of the number of selected matches (Section \ref{sup:subsec:matchingnum}) and dense matching results (Section \ref{sup:subsec:densematching}).
    \item Section \ref{sup:sec:crossdataset} evaluates the cross-dataset transferability of our method.
    \item Section \ref{sup:sec:limitations} discusses the limitations of cross-view image matching in challenging scenarios.
    \item Finally, Section \ref{sup:sec:qualitative} provides extra qualitative results for both localization and image matching.
\end{enumerate}

\section{CVFM Benchmark Details}
\label{sup:sec:CVFM}

The CVFM (Cross-View Fine-grained image Matching) benchmark is constructed using imagery from the DReSS~\cite{xia2025cross} and DReSS-D~\cite{zhang2025cross} datasets. 
The DReSS dataset provides ground-view panoramas collected from cities across multiple continents, while DReSS-D additionally includes co-registered satellite images and per-pixel depth maps aligned with these panoramas. 
We leverage the ground-truth depth maps to project pixels from each ground-view panorama into the satellite image space, thereby establishing dense pixel-level correspondences between the two views.

To ensure geometric validity, we constrain the valid region to within 30\,m of the ground camera center, excluding distant, sky, and occluded regions. 
However, temporal discrepancies between the ground panoramas and the 3D model-rendered depth maps may still lead to structural inconsistencies or partial occlusions (e.g., newly constructed buildings or seasonal vegetation). 
To guarantee annotation quality, all projected correspondences were manually inspected by three graduate students with computer vision expertise. 
After this verification, the final CVFM benchmark comprises 32,509 high-quality image pairs with reliable pixel-level correspondences. 

\begin{figure*}[ht]
    \captionsetup[subfigure]{labelformat=empty}
    \tikzset{inner sep=0pt}
    \setkeys{Gin}{height=2.8cm, keepaspectratio}
    \centering
    \subfloat[]{%
    \tikz{\node (a) {\includegraphics{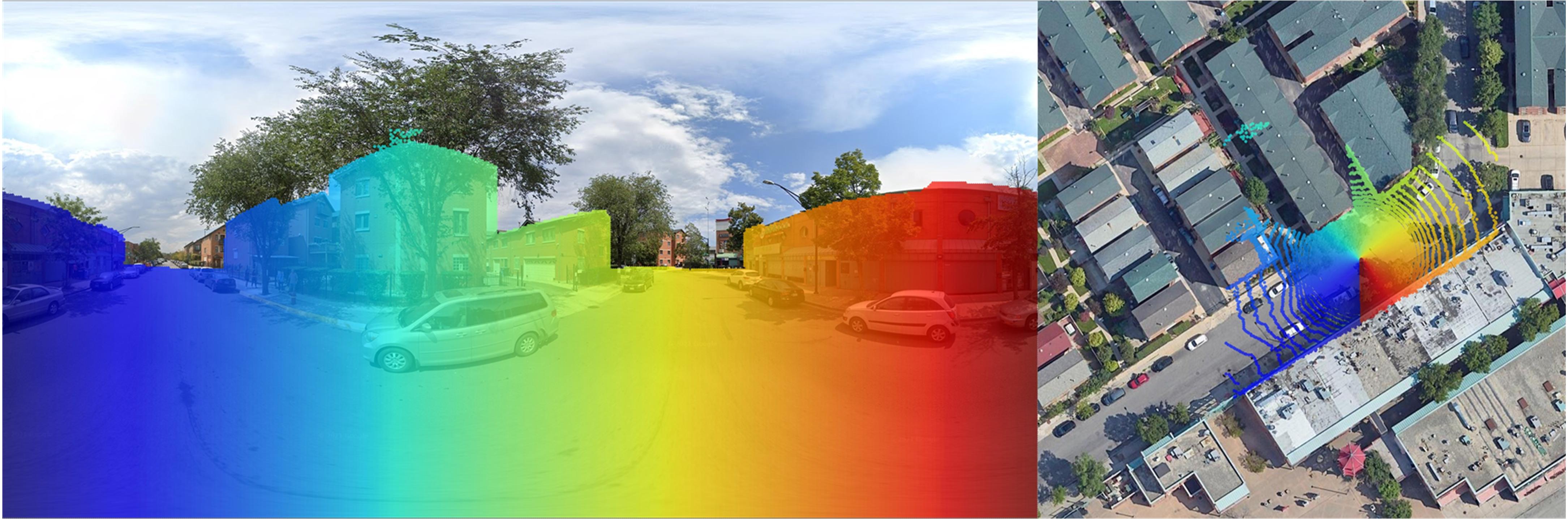}};
        \node[below right=2mm] at (a.north west) {(a)}; 
      }}
    \hfil
    \subfloat[]{%
    \tikz{\node (a) {\includegraphics{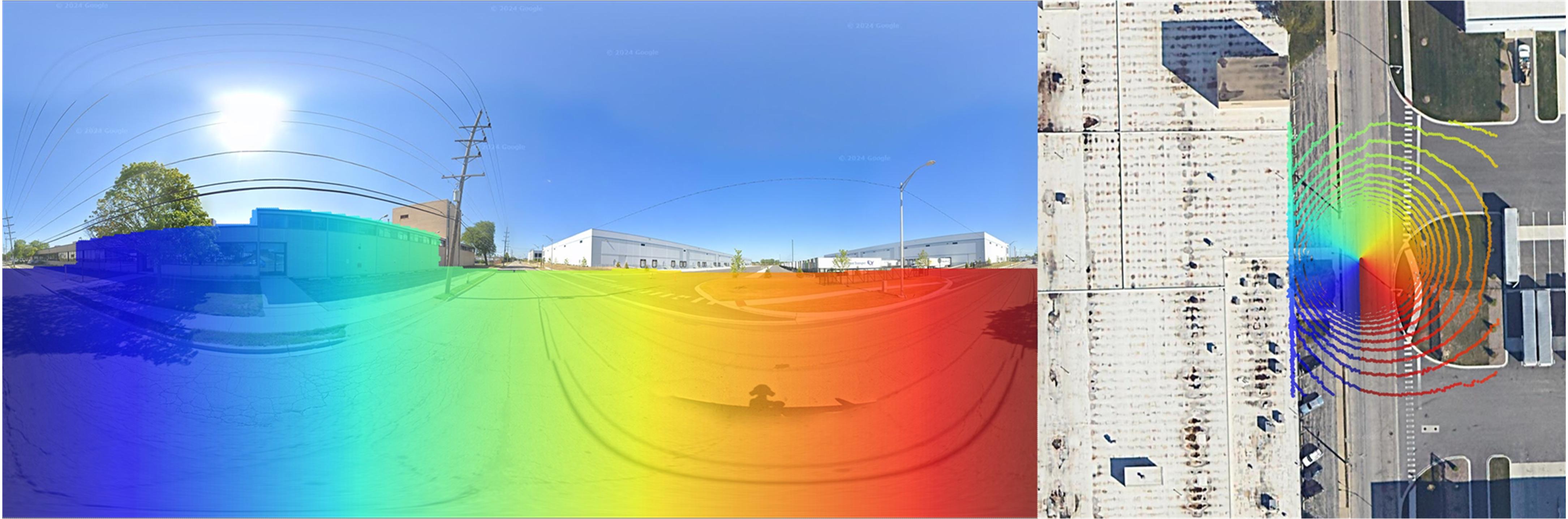}};
        \node[below right=2mm] at (a.north west) {(b)}; 
      }}
    \hfil
    \subfloat[]{%
    \tikz{\node (a) {\includegraphics{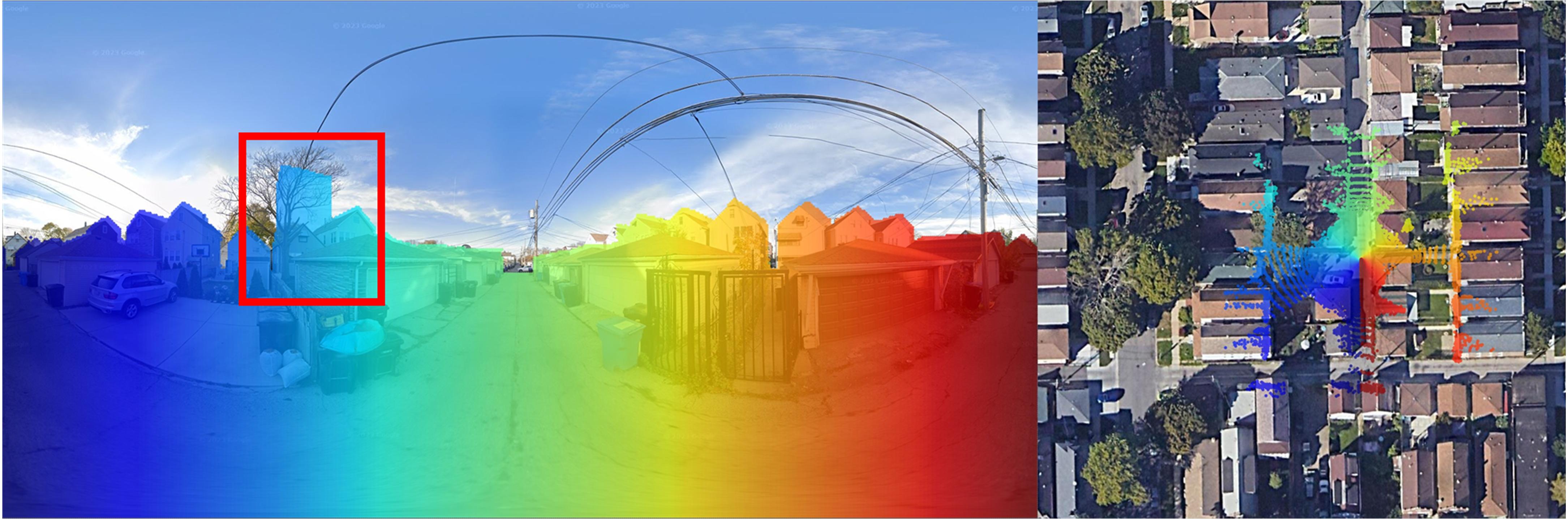}};
        \node[below right=2mm] at (a.north west) {(c)}; 
      }}
    \hfil
    \subfloat[]{%
    \tikz{\node (a) {\includegraphics{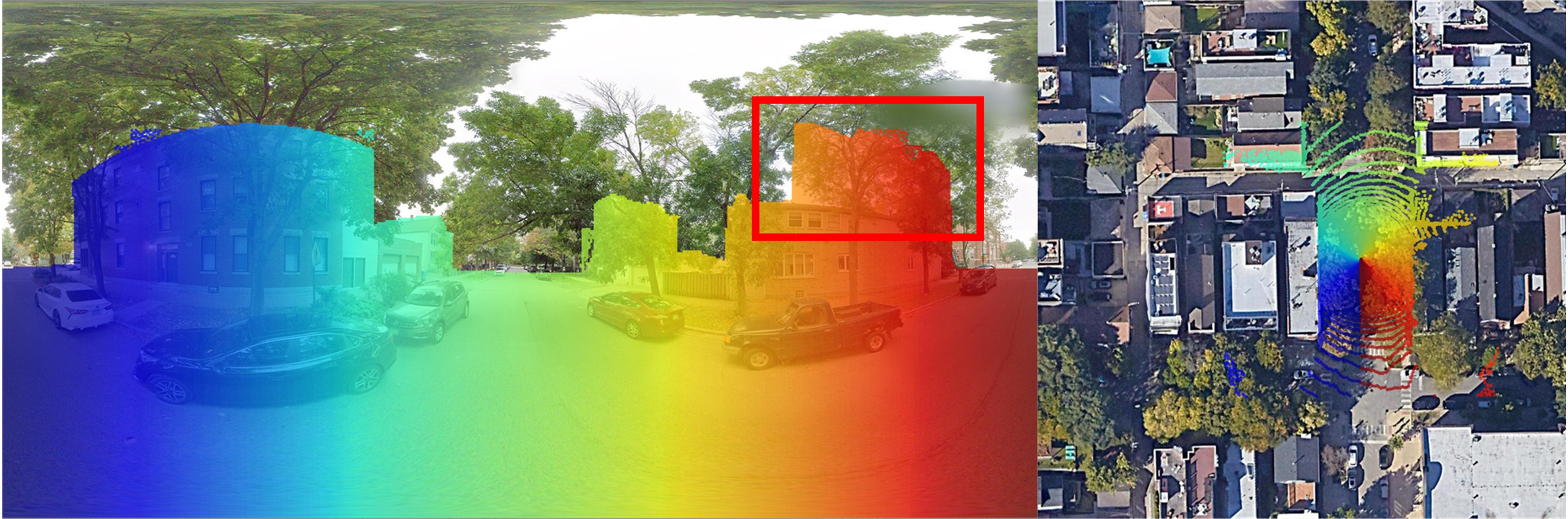}};
        \node[below right=2mm] at (a.north west) {(d)}; 
      }}
    \hfil
   \caption{
    Representative examples from the CVFM benchmark. 
    (a,b) Accepted samples showing precise pixel-level correspondences. 
    (c,d) Rejected samples containing structural misalignment or occlusion, with erroneous regions highlighted in red boxes.
    }
    \label{fig:CVFM_examples}
\end{figure*}

Fig.~\ref{fig:CVFM_examples} shows representative examples of accepted and rejected samples. 
The accepted examples exhibit precise alignment between visible structures in both views, while rejected ones contain geometric inconsistencies or missing structures (highlighted in red). 
These examples illustrate the annotation rigor and diversity of the CVFM benchmark, which establishes a reliable testbed for fine-grained cross-view image matching.

\section{Loss Functions and Training Details}
\label{sup:sec:loss}

The entire framework is trained end-to-end with a composite loss that jointly optimizes geometric alignment, feature correspondence, and surface estimation. 
Specifically, we use three complementary objectives: (i) a Virtual Correspondence Error (VCE) loss \(\mathcal{L}_{\text{VCE}}\) to supervise the predicted 3-DoF pose, (ii) a matching loss \(\mathcal{L}_{\text{M}}\) to enforce patch-level alignment in BEV space, and (iii) a height consistency loss \(\mathcal{L}_{\text{height}}\) to regularize the estimated surfaces. 
The overall loss is defined as:
\begin{equation}
\mathcal{L} = \mathcal{L}_{\text{VCE}} + \mathcal{L}_{\text{M}} + \mathcal{L}_{\text{height}}.
\label{eq:loss_all}
\end{equation}

\textbf{Virtual Correspondence Error (VCE).}  
To supervise the predicted pose \(P_{\text{pred}} = [R, T]\), we adopt the Virtual Correspondence Error~\cite{arnold2022map}, which measures the geometric deviation between projected BEV points under predicted and ground-truth transformations. 
A set of uniformly sampled 2D query points \(\{pts_i\}_{i=1}^{N_v}\) is defined within a square BEV region of size \(L_v \times L_v\). 
Each point is projected into the reference frame using both \(P_{\text{pred}}\) and \(P_{\text{gt}}\), producing transformed coordinates \(pts_i^{\text{pred}}\) and \(pts_i^{\text{gt}}\). 
The loss minimizes the average Euclidean distance between these two sets:
\begin{equation}
\mathcal{L}_{\text{VCE}} = \frac{1}{N_v} \sum_{i=1}^{N_v} \left\| pts_i^{\text{pred}} - pts_i^{\text{gt}} \right\|_2.
\label{eq:loss_vce}
\end{equation}
This term directly constrains the spatial correctness of the predicted transformation in BEV space.

\textbf{Matching loss.}  
To enhance cross-view feature alignment, we follow~\cite{xia2025fg} and supervise the similarity matrix using ground-truth correspondences derived from \(P_{\text{gt}}\). 
For each ground patch \(n_{\text{grd}} = (x, y)\), we project its BEV coordinate into the satellite frame to obtain its counterpart \(n_{\text{sat}}'\). 
The reverse mapping \(n_{\text{sat}} \rightarrow n_{\text{grd}}'\) is constructed symmetrically. 
We then apply two directional InfoNCE losses over the similarity matrix \(S_{\text{orig}} \in \mathbb{R}^{N^2 \times N^2}\):
\begin{align}
\mathcal{L}_{\text{infoNCE}}^{\text{grd} \rightarrow \text{sat}} &= - \frac{1}{N_s} \sum_{i=1}^{N_s} 
\log \left( \frac{e^{{S_{orig}}_{n_{\text{grd}}, n_{\text{sat}}'}}}
{\sum_{j=1}^{N^2} e^{{S_{orig}}_{n_{\text{grd}}, j}}} \right), \\
\mathcal{L}_{\text{infoNCE}}^{\text{sat} \rightarrow \text{grd}} &= - \frac{1}{N_s} \sum_{i=1}^{N_s} 
\log \left( \frac{e^{{S_{orig}}_{n_{\text{grd}}', n_{\text{sat}}}}}
{\sum_{j=1}^{N^2} e^{{S_{orig}}_{j, n_{\text{sat}}}}} \right).
\end{align}
The bidirectional matching loss is their average:
\begin{equation}
\mathcal{L}_{\text{M}} = \frac{1}{2} 
\left( \mathcal{L}_{\text{infoNCE}}^{\text{grd} \rightarrow \text{sat}} + 
\mathcal{L}_{\text{infoNCE}}^{\text{sat} \rightarrow \text{grd}} \right).
\label{eq:loss_m}
\end{equation}
This contrastive formulation encourages each BEV patch to find its most consistent counterpart across views, improving correspondence precision and match separability.

\textbf{Height consistency loss.}  
To regularize the predicted height fields from both modalities, we impose a cross-view consistency constraint. 
Using the same correspondences \(n_{\text{grd}} \leftrightarrow n_{\text{sat}}'\), we extract their estimated depths from the ground and satellite surface maps and minimize their absolute difference:
\begin{equation}
\mathcal{L}_{\text{height}} = 
\frac{1}{N_s} \sum_{i=1}^{N_s} 
\frac{1}{K} \left| 
\text{depth}_{\text{grd}}(n_{\text{grd}}) - 
\text{depth}_{\text{sat}}(n_{\text{sat}}') 
\right|,
\label{eq:loss_height}
\end{equation}
where \(K = 100\) normalizes the scale to stabilize training. 
This loss enforces structural agreement between height predictions, reinforcing geometric consistency across views.

\section{Ablation on Module Design}
\label{sup:sec:module}

\subsection{Detailed module design and selection}
\label{sup:subsec:moduledesign}
To further clarify the motivation and design rationale behind our proposed modules, we conduct extended ablation studies on both the \textit{Surface Model} and \textit{SimRefiner}. 
These experiments aim to verify that the observed improvements are not the result of ad-hoc additions, but rather stem from targeted solutions to specific weaknesses in BEV-based cross-view localization frameworks.  
All experiments are performed on the VIGOR dataset under the known-orientation setting, with identical seeds and training configurations.  
Results are summarized in Tab.~\ref{tab:detailed_ablation}.

\textbf{Motivation and problem analysis.}  
Our preliminary investigation revealed two fundamental limitations in conventional BEV-based localization pipelines:  
(1) the absence of a mechanism to filter geometrically invalid BEV samples, leading to feature contamination from non-visible or elevated regions; and  
(2) the lack of structure-aware reasoning within the similarity matrix, resulting in unstable or spatially inconsistent match distributions.  
The \textit{Surface Model} and \textit{SimRefiner} are explicitly designed to address these two bottlenecks from complementary perspectives: BEV construction and BEV matching.

\textbf{Surface mechanism analysis.}  
As detailed in the main paper, the Surface Model operates on a 3D voxel grid and aggregates information along the vertical dimension. 
Since only a subset of these layers correspond to physically valid surfaces (e.g., ground, rooftops), we experimented with different strategies to model vertical aggregation:
\begin{itemize}[leftmargin=1.2em]
    \item Weighted surface: computes a soft weighted sum across vertical layers.  
    \item Accumulated surface: progressively integrates features from bottom to top until the estimated visible surface is reached.  
\end{itemize}
As shown in \ref{tab:detailed_ablation}, the accumulated variant yields lower localization error and better interpretability, since it directly mimics the physical accumulation of visible surfaces in real-world height fields.  
This variant is therefore adopted as our default setting.

\textbf{Similarity distribution refinement (SimRefiner).}  
We also investigate multiple internal configurations of the SimRefiner module.  
It consists of two complementary refinement branches, local and global, and a confidence gating:
\begin{itemize}[leftmargin=1.2em]
    \item Local refinement: captures short-range spatial coherence through 3D convolution in the similarity cube. 
    \item Global refinement: models long-range structural dependencies using a row-wise MLP.  
    \item Confidence gating: adaptively weights correction strength based on estimated patch-wise confidence.  
\end{itemize}
The ablation results confirm that each component contributes distinct benefits.  
Adding confidence gating stabilizes optimization and further reduces cross-area localization error.

\textbf{Joint module evaluation.}  
Finally, we examine the combined contribution of both modules.  
Integrating the SimRefiner on top of the Surface Model leads to the lowest overall localization error, showing that geometric grounding and probabilistic refinement are mutually reinforcing.  
In particular, the confidence gating within SimRefiner stabilizes training and provides consistent improvements across both same-area and cross-area settings.

\begin{table}[t]
    \centering
    \caption{
    Extended ablations on module design under the VIGOR known-orientation setting. 
    We report mean and median localization errors (in meters). 
    \textbf{Best results are highlighted in bold.}
    }
    \begin{tabular}{lcc}
    \toprule
    Methods & Same-area & Cross-area \\
    \hline
    \textbf{\textit{Surface mechanism}} & & \\
    Base: Max & 2.10 / 1.03 & 2.73 / 1.49 \\
    + Weighted surface & 2.12 / 1.10 & 2.69 / 1.52 \\
    + Accumulated surface & 2.03 / 1.05 & 2.45 / 1.38 \\
    \hline
    \textbf{\textit{Similarity refinement}} & & \\
    Base & 2.10 / 1.03 & 2.73 / 1.49 \\
    + Local only & 2.06 / 1.02 & 2.56 / 1.42 \\
    + Global only & 2.08 / 1.04 & 2.60 / 1.44 \\
    + SimRefiner (w/o conf) & 2.05 / 1.02 & 2.51 / 1.42 \\
    + SimRefiner (w/ conf) & 2.03 / 1.01 & 2.45 / 1.37 \\
    \hline
    \textbf{\textit{Combination}} & & \\
    Base: Accumulated surface & 2.03 / 1.05 & 2.45 / 1.38 \\
    + SimRefiner (w/o conf) & 1.91 / 0.99 & 2.39 / 1.36 \\
    + SimRefiner (w/ conf) & \textbf{1.89} / \textbf{0.97} & \textbf{2.33} / \textbf{1.34} \\
    \bottomrule
    \end{tabular}
    \label{tab:detailed_ablation}
\end{table}

\subsection{Additional ablations on Surface Model}
\label{sup:subsec:surface}
\textbf{Selection on the assumed ground-to-camera height.}
As described in the main paper, we set the minimum depth value in the satellite-derived depth map to \(-3\,\text{m}\) in the ground-view camera coordinate system, following the common observation that the physical distance from the ground plane to the camera center typically falls within 2-3\,m. 
To evaluate the sensitivity of our Surface Model to this assumption, we vary the pseudo ground height from \(-1\) to \(-5\) meters and report the localization accuracy in Tab.~\ref{tab:ablation on surface height}.

Across both same-area and cross-area settings, the performance remains stable over a broad range, and the best results are achieved when the assumed height is close to the real-world camera elevation. 
Larger deviations in either direction cause slight degradation, likely because the BEV feature sampling shifts away from geometrically valid visible surfaces. 
These results demonstrate that our Surface Model is robust to height perturbations while still benefiting from a physically meaningful initialization.

\begin{table}[t]
    \centering
    \caption{Ablations on the assumed height in Surface Model in VIGOR Known orientation setting.
    We report mean and median localization errors (in meters). 
    \textbf{Best in bold.}}
    \begin{tabular}{p{2cm}cc}
    \toprule
    \multirow{2}{*}{{Height.}}& Same-area& Cross-area\\
        & Mean / Median& Mean / Median\\
        \hline
 \(-1\,\text{m}\)& 2.02 / 1.03& 2.37 / 1.36\\
 \(-2\,\text{m}\)& 1.91 / 0.99& 2.34 / \textbf{1.33}\\
 \(-3\,\text{m}\)& \textbf{1.89} / \textbf{0.97} & \textbf{2.33} / 1.34 \\
 \(-4\,\text{m}\)& 1.92 / 1.01& 2.36 / 1.35\\
    \(-5\,\text{m}\)& 1.97 / 1.02& 2.36 / 1.37\\
    \bottomrule
    \end{tabular}
    \label{tab:ablation on surface height}
\end{table}

\textbf{Influence of the Surface Model on Cross-View Matching.}
To better understand how the Surface Model contributes to matching performance, we conduct a detailed ablation using base+SimRefiner as the baseline. 
We decompose the Surface Model into two factors: 
(1) the method used to determine the visible surface, and 
(2) the height consistency loss constraint applied during training.

For (1), we compare three strategies for selecting the surface depth:  
\textit{Base: Max}, which takes the maximum-weighted layer;  
\textit{Weighted}, which uses a weighted average over layers;  
and \textit{Accumulated}, our default design that identifies the surface via cumulative visibility.  

For (2), we additionally evaluate the effect of applying height consistency loss during training under the \textit{Weighted} and \textit{Accumulated} settings.

Results in Tab.~\ref{tab:ablation on surface on cvfm} show that the accumulated strategy is the most effective formulation for approximating the surface geometry.  
Moreover, incorporating a height consistency loss substantially improves fine-grained matching accuracy, demonstrating that reliable surface estimation plays a central role in establishing cross-view correspondences.

\begin{table}[t]
    \centering
    \caption{CVFM cross-view image matching results (measured on the satellite image) using the top-30 matches. 1 pixel equals approximately \(0.12\,\mathrm{m}\). 
    * represents training with height consistency loss. 
    \textbf{Best in bold.}}
    \begin{tabular}{p{2.3cm}p{0.95cm}p{0.95cm}p{0.95cm}p{0.95cm}}
    \toprule
    Method & $\le$5px$\uparrow$ & $\le$10px$\uparrow$ & $\le$15px$\uparrow$ & Valid\\
    \hline
    Base: Max& 1.3\%& 5.1\%& 10.5\%&0.987\\
 Weighted& 1.1\%& 4.2\%& 9.0\%&0.986\\
 Accumulated& 1.6\%& 6.3\%& 12.7\%&0.705\\
 \hline
    Weighted*& 1.9\%& 7.5\%& 15.7\%&0.690\\
    Accumulated*& \textbf{3.1\%}& \textbf{11.7\%}& \textbf{24.0\%}&0.915\\
    \bottomrule
    \end{tabular}
    \label{tab:ablation on surface on cvfm}
\end{table}

\section{RANSAC-based Refinement Study}
\label{sup:sec:ransac}

Following FG2~\cite{xia2025fg}, we further evaluate the impact of applying RANSAC as a post-processing refinement during inference. 
RANSAC serves as a geometric verification step that refines the estimated 3-DoF pose based on the predicted correspondences. 
It is applied only at test time with $R = 100$ iterations, an inlier threshold of 2.5\,m, and 256 randomly sampled correspondence pairs per iteration.

As summarized in Tab.~\ref{tab:vigor_ransac}, incorporating RANSAC consistently improves both translation and orientation accuracy across all protocols on the VIGOR benchmark.  
Under the \textit{known-orientation} setting, our method achieves mean localization errors of \textbf{1.78\,m} (same-area) and \textbf{2.20\,m} (cross-area), outperforming FG2-RANSAC by 0.18\,m and 0.35\,m, respectively.  
In the more challenging \textit{unknown-orientation} setting, RANSAC further enhances both translation and rotation estimation: the same-area mean orientation error decreases from 8.65$^\circ$ to \textbf{6.02$^\circ$}, while the cross-area translation error is reduced from 4.19\,m to \textbf{3.79\,m}.  
These improvements demonstrate that our predicted correspondences are geometrically more consistent and robust for pose fitting.

Importantly, this refinement introduces only a marginal computational overhead.  
Inference is measured on an NVIDIA Tesla V100 GPU setup with batch size = 1, where the average runtime increases from 75.5\,ms to 89.1\,ms (less than 20\% additional latency) while delivering notable accuracy gains.  
Given its low cost and consistent benefit, RANSAC serves as a practical geometric refinement module for our correspondence-based localization framework.

\begin{table*}[t]
    \centering
    \caption{
     Comparison with RANSAC on the VIGOR benchmark. 
    FG2$^\dagger$ denotes our re-implementation, which achieves slightly higher accuracy than the originally reported results. 
    RANSAC is applied only at inference time. 
    \textbf{Best} and \underline{second-best} results are highlighted.
}
    \begin{tabular}{p{1cm}p{2cm}p{1.5cm}p{1.0cm}p{1.0cm}p{1.0cm}p{1.0cm}|p{1.0cm}p{1.0cm}p{1.0cm}p{1.0cm}}
    \toprule
    \multirow{3}{*}{{Orien.}} & \multirow{3}{*}{Methods}  &\multirow{3}{*}{Time}& 
    \multicolumn{4}{c}{Same-area} & \multicolumn{4}{c}{Cross-area} \\
    \cline{4-11} 
    &  && \multicolumn{2}{c}{$\downarrow$ Localization (m)} & \multicolumn{2}{c}{$\downarrow$ Orientation ($^\circ$)} & \multicolumn{2}{c}{$\downarrow$ Localization (m)} & \multicolumn{2}{c}{$\downarrow$ Orientation ($^\circ$)} \\
    \cline{4-11} 
    &  && Mean & Median & Mean & Median & Mean & Median & Mean & Median \\
    \hline
    \multirow{4}{*}{Known}& 
    FG2$^\dagger$~\cite{xia2025fg} &74.2ms& 2.10& 1.03& -& -& 2.73& 1.49& -& -\\
 & Ours &75.5ms& \underline{1.89}& 0.97& -& -& \underline{2.33}& \underline{1.34}& -&-\\
 & FG2$^\dagger$-Ransac&88.8ms& 1.96& \underline{0.96}& -& -& 2.55& 1.40& -&-\\
 & Ours-Ransac &89.1ms& \textbf{1.78}& \textbf{0.91}& -& -& \textbf{2.20}& \textbf{1.28}& -&-\\
    
    \hline
    \multirow{4}{*}{Unknown}& 
    FG2$^\dagger$~\cite{xia2025fg} &74.2ms
& 3.90& 2.22& 7.07& 1.83& 4.96& 2.76& 11.07& 2.40\\
 & Ours &75.5ms
& \underline{3.11}& \textbf{1.31}& 8.65& 2.69& \underline{4.19}& \underline{2.26}& 12.56&3.05\\
    & FG2$^\dagger$-Ransac&88.8ms
& 3.45& 1.89& \underline{6.03}& \underline{1.23}& 4.49& 2.39& \textbf{9.89}& \textbf{1.64}\\
 & Ours-Ransac &89.1ms& \textbf{3.08}& \underline{1.57}& \textbf{6.02}& \textbf{1.20}& \textbf{3.79}& \textbf{2.01}& \underline{10.59}&\underline{2.04}\\
 \bottomrule
    \end{tabular}
    \label{tab:vigor_ransac}
\end{table*}

\section{Discussion on orientation errors}
\label{sup:sec:orientation}

Under the unknown-orientation setting, our method exhibits slightly larger orientation errors than FG2, with mean and median gaps within \(3^\circ\). To understand whether this discrepancy affects practical usage, we further evaluate a simple but realistic two-stage strategy commonly adopted in applications requiring high-precision orientation.

In practice, the orientation predicted in a first-stage “unknown-orientation” localization is often sufficient to bring the yaw error into a moderate range. Once the orientation is roughly corrected, a second-stage model trained under the known-orientation setting can be applied to refine the final 3-DoF pose. This mimics real deployment: the first stage provides a coarse orientation prior, and the second stage performs precise localization conditioned on a narrowed orientation range.

To simulate this scenario, we take our model trained under the known-orientation configuration and evaluate it on queries whose yaw deviation is artificially restricted to \(\pm 20^\circ\) and \(\pm 30^\circ\), respectively. These ranges are chosen to reflect the observed error distribution from the unknown-orientation results on VIGOR and DReSS-D.

As shown in Tab.~\ref{tab:vigor_two_stage}, the known-orientation model maintains high localization accuracy under these limited-offset conditions and outperforms FG2. This demonstrates that although our orientation estimates are slightly less accurate in the fully unknown case, this difference does not hinder achieving high-precision localization in realistic usage. Once a coarse orientation prior is available, even a simple two-stage procedure effectively recovers strong accuracy.

\begin{table*}[t]
    \centering
    \caption{
Evaluation of a simulated two-stage localization setting. 
A model trained under the Known Orientation setting is tested on queries with restricted orientation ranges (\(\pm 20^\circ\) and \(\pm 30^\circ\)) to mimic localization with a coarse orientation prior. 
FG2$^\dagger$ denotes our re-implementation with slightly higher accuracy than the originally reported model. 
\textbf{Best results are highlighted.}
}
    \begin{tabular}{p{2.5cm}p{2cm}p{1.0cm}p{1.0cm}p{1.0cm}p{1.0cm}|p{1.0cm}p{1.0cm}p{1.0cm}p{1.0cm}}
    \toprule
    \multirow{3}{*}{{Orien.}} & \multirow{3}{*}{Methods}  & 
    \multicolumn{4}{c}{Same-area} & \multicolumn{4}{c}{Cross-area} \\
    \cline{3-10}&  & \multicolumn{2}{c}{$\downarrow$ Localization (m)} & \multicolumn{2}{c}{$\downarrow$ Orientation ($^\circ$)} & \multicolumn{2}{c}{$\downarrow$ Localization (m)} & \multicolumn{2}{c}{$\downarrow$ Orientation ($^\circ$)} \\
    \cline{3-10}&  & Mean & Median & Mean & Median & Mean & Median & Mean & Median \\
    \hline
    \multirow{2}{*}{Unknown $\pm 20^\circ$}& 
    FG2$^\dagger$~\cite{xia2025fg} & 2.34& \textbf{1.20}& 3.31& 2.23& 3.03& 1.69& 3.71& 2.48\\
 & Ours& \textbf{2.15}& 1.32& \textbf{3.01}& \textbf{1.85}& \textbf{2.60}& \textbf{1.49}& \textbf{3.41}&\textbf{2.05}\\
    
    \hline
    \multirow{2}{*}{Unknown $\pm 30^\circ$}& 
    FG2$^\dagger$~\cite{xia2025fg} & 2.76& 1.47& 5.03& 3.35& 3.45& 1.96& 5.42& 3.61\\
 & Ours& \textbf{2.62}& \textbf{1.40}& \textbf{4.59}& \textbf{2.69}& \textbf{3.02}& \textbf{1.71}& \textbf{4.79}&\textbf{2.78}\\
 \bottomrule
    \end{tabular}
    \label{tab:vigor_two_stage}
\end{table*}

\section{Cross-View Image Matching Experiments}
\label{sup:sec:matching}

\subsection{Influence of Number of Selected Matches}
\label{sup:subsec:matchingnum}
We further investigate the influence of the number of selected top-$N_{\text{mch}}$ matches on cross-view image matching accuracy. 
During inference, the highest-similarity correspondences predicted by the model are used to form sparse ground-satellite matches. 
We vary $N_{\text{mch}}$ from 10 to 100 and evaluate the resulting projection accuracy on the CVFM benchmark using the same setup as in the main paper.

As shown in Tab.~\ref{tab:top-k_match}, all metrics gradually decrease as the number of matches increases. 
This trend suggests that while including more correspondences provides broader coverage, it also introduces lower-confidence pairs that slightly degrade precision and valid ratio. 
Overall, the results remain highly stable across different $N_{\text{mch}}$ values, suggesting that the similarity distribution is compact and well-calibrated, and that most reliable matches are concentrated within the top-ranked predictions.

In practice, there exists a trade-off between the number of selected matches and the resulting geometric accuracy. 
Users can flexibly adjust $N_{\text{mch}}$ based on their precision requirements or computational constraints. 
We set $N_{\text{mch}}=30$ as the default configuration for all sparse matching and localization experiments, which achieves a favorable balance between accuracy and efficiency.

\begin{table}[t]
    \centering
    \caption{CVFM cross-view image matching results using the different top-\(N_{\text{mch}}\) matches. 1 pixel equals approximately \(0.12\,\mathrm{m}\).}
    \begin{tabular}{p{1.3cm}p{1.2cm}p{1.2cm}p{1.2cm}p{1.2cm}}
    \toprule
    \(N_{\text{mch}}\)& $\le$5px$\uparrow$& $\le$10px$\uparrow$& $\le$15px$\uparrow$&Valid\\
    \hline
    10& 3.0\%& 11.8\%& 24.4\%&91.6\%\\
 20& 3.0\%& 11.8\%& 24.2\%&91.6\%\\
 30& 3.1\%& 11.7\%& 24.0\%&91.5\%\\
 40& 3.0\%& 11.6\%& 23.7\%&91.2\%\\
    50& 3.0\%& 11.5\%& 23.4\%&91.0\%\\
    60& 3.0\%& 11.4\%& 23.1\%&90.7\%\\

 70& 3.0\%& 11.2\%& 22.7\%&90.5\%\\
 80& 2.9\%& 11.1\%& 22.3\%&90.3\%\\
 90& 2.9\%& 10.9\%& 21.9\%&90.0\%\\
 100& 2.8\%& 10.7\%& 21.5\%&89.8\%\\
     \bottomrule
    \end{tabular}
    \label{tab:top-k_match}
\end{table}

\subsection{Dense Cross-View Matching}
\label{sup:subsec:densematching}
We further extend our framework to evaluate its capability for dense cross-view image matching. 
Inspired by recent advances in joint depth estimation-matching~\cite{xia2025fine}, we combine the predicted depth maps of ground-view panoramas with the estimated 3-DoF poses to generate dense correspondences between ground and satellite views. 
Specifically, we use a frozen DepthAnything v2-small model~\cite{yang2024depth} to predict per-pixel metric depth for each ground panorama, which is then converted into a 3D point cloud. 
Points located farther than 35\,m from the camera center are filtered out to suppress noise from distant or sky regions. 
Using the predicted translation and orientation from our localization stage, along with the top-30 high-confidence cross-view matches for scale estimation, we project every valid ground pixel onto the satellite image plane to obtain dense correspondence maps.

\textbf{Quantitative results.}
We evaluate dense matching accuracy on the CVFM benchmark using 400K projected correspondences and compare our method against FG2~\cite{xia2025fg}. 
As shown in Tab.~\ref{tab:cvfm_densematching}, our method significantly outperforms FG2 across all pixel thresholds, demonstrating the advantage of geometry-grounded BEV representations and refined similarity distributions.
When equipped with RANSAC-based geometric refinement at test time, both methods show improved precision, yet our framework maintains a clear margin, achieving the highest accuracy under all evaluation criteria.
These results verify that the proposed modules not only enhance localization but also enable more robust and geometrically consistent dense correspondence estimation under extreme viewpoint disparities. 

\begin{table}[t]
    \centering
    \caption{
Dense cross-view image matching results on the CVFM benchmark using 400K correspondences. 1 pixel equals approximately \(0.12\,\mathrm{m}\). \textbf{Best} and \underline{second-best} results are highlighted.}
    \begin{tabular}{p{2.5cm}p{0.9cm}p{0.9cm}p{0.9cm}p{0.9cm}}
    \toprule
    Method& $\le$5px$\uparrow$& $\le$10px$\uparrow$& $\le$15px$\uparrow$&Valid\\
    \hline
    FG2~\cite{xia2025fg}& 1.5\%& 5.0\%& 9.4\%&0.892\\
    Ours& \underline{6.5\%}& \underline{19.7\%}& \underline{31.9\%}&0.828\\
  
 FG2-Ransac& 1.6\%& 5.5\%& 10.3\%&0.892\\
 Ours-Ransac& \textbf{7.2\%}& \textbf{21.7\%}& \textbf{34.7\%}&0.828\\
   \bottomrule
    \end{tabular}
    \label{tab:cvfm_densematching}
\end{table}

\begin{figure*}[ht]
    \captionsetup[subfigure]{labelformat=empty}
    \tikzset{inner sep=0pt}
    \setkeys{Gin}{height=3.2cm, keepaspectratio}
    \centering
    \subfloat[]{%
    \tikz{\node (a) {\includegraphics{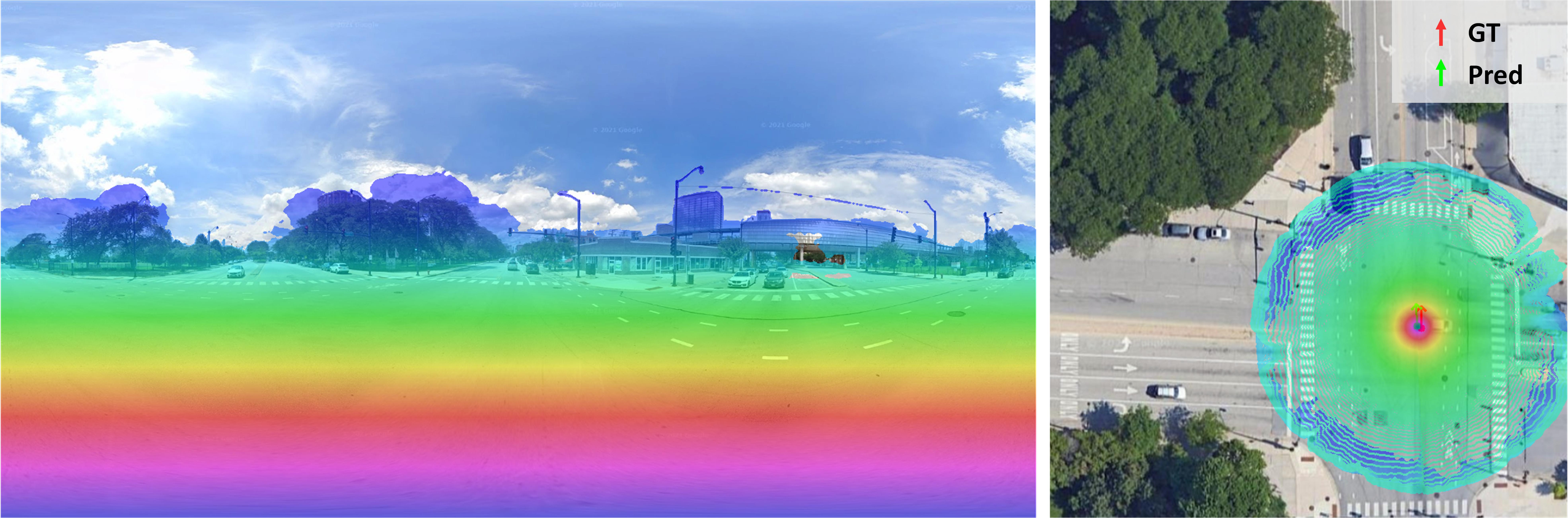}};
        \node[below right=2mm] at (a.north west) {(a)}; 
      }}
    \hfil
    \subfloat[]{%
    \tikz{\node (a) {\includegraphics{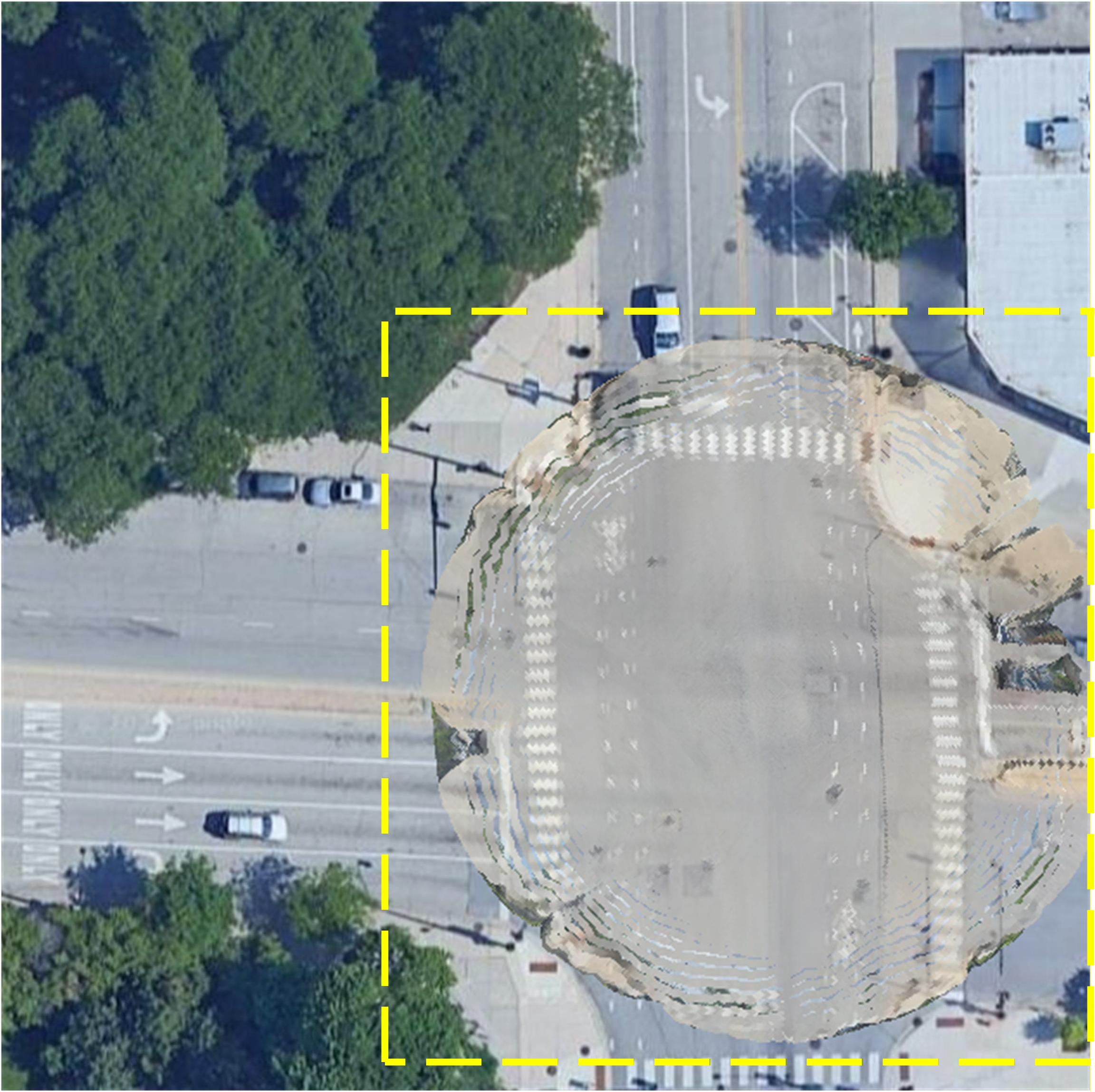}};
        \node[below right=2mm] at (a.north west) {(b)}; 
      }}
    \hfil
    \subfloat[]{%
    \tikz{\node (a) {\includegraphics{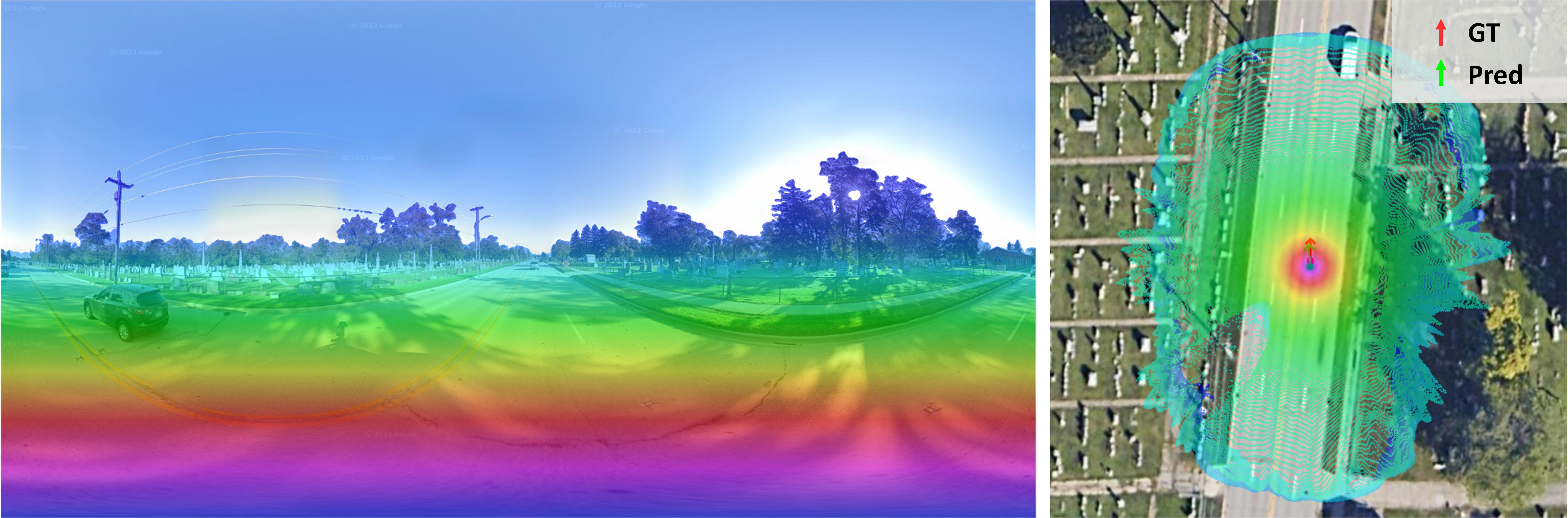}};
        \node[below right=2mm] at (a.north west) {(c)}; 
      }}
    \hfil
    \subfloat[]{%
    \tikz{\node (a) {\includegraphics{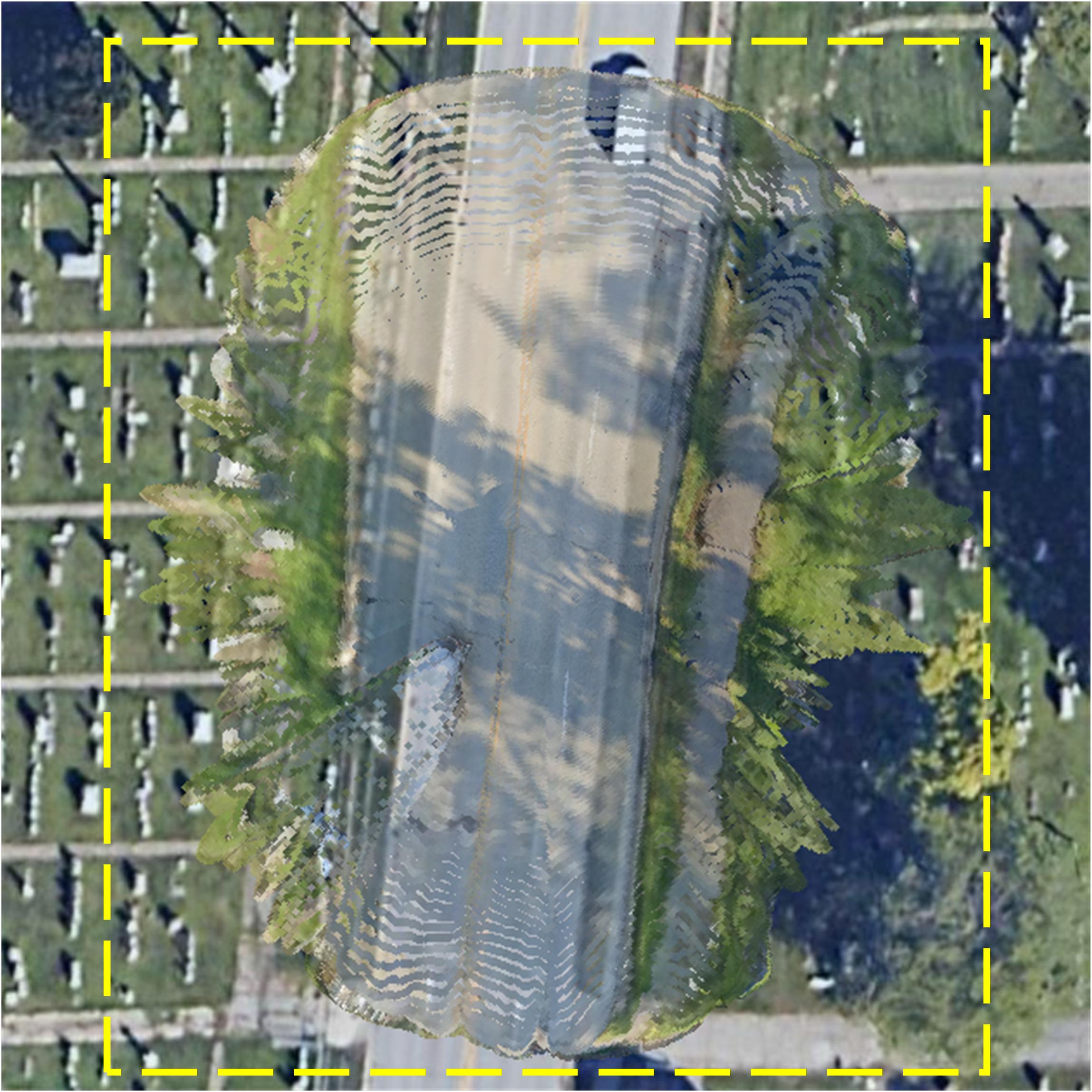}};
        \node[below right=2mm] at (a.north west) {(d)}; 
      }}
    \hfil
    \subfloat[]{%
    \tikz{\node (a) {\includegraphics{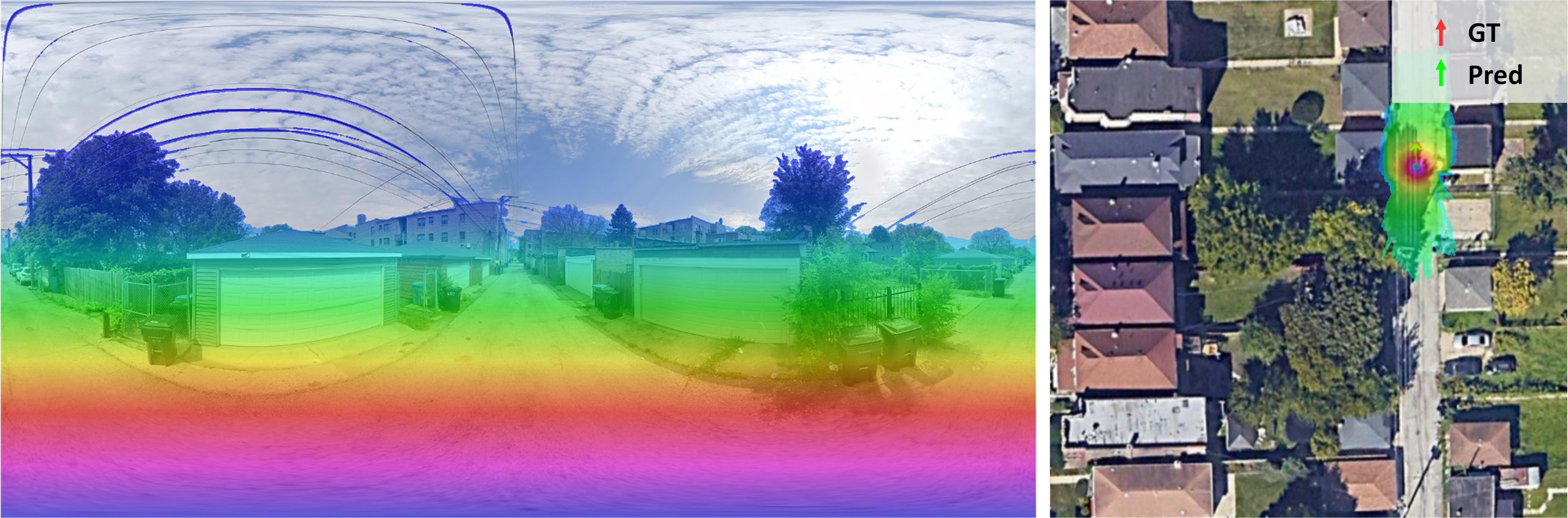}};
        \node[below right=2mm] at (a.north west) {(e)}; 
          }}
    \hfil
    \subfloat[]{%
    \tikz{\node (a) {\includegraphics{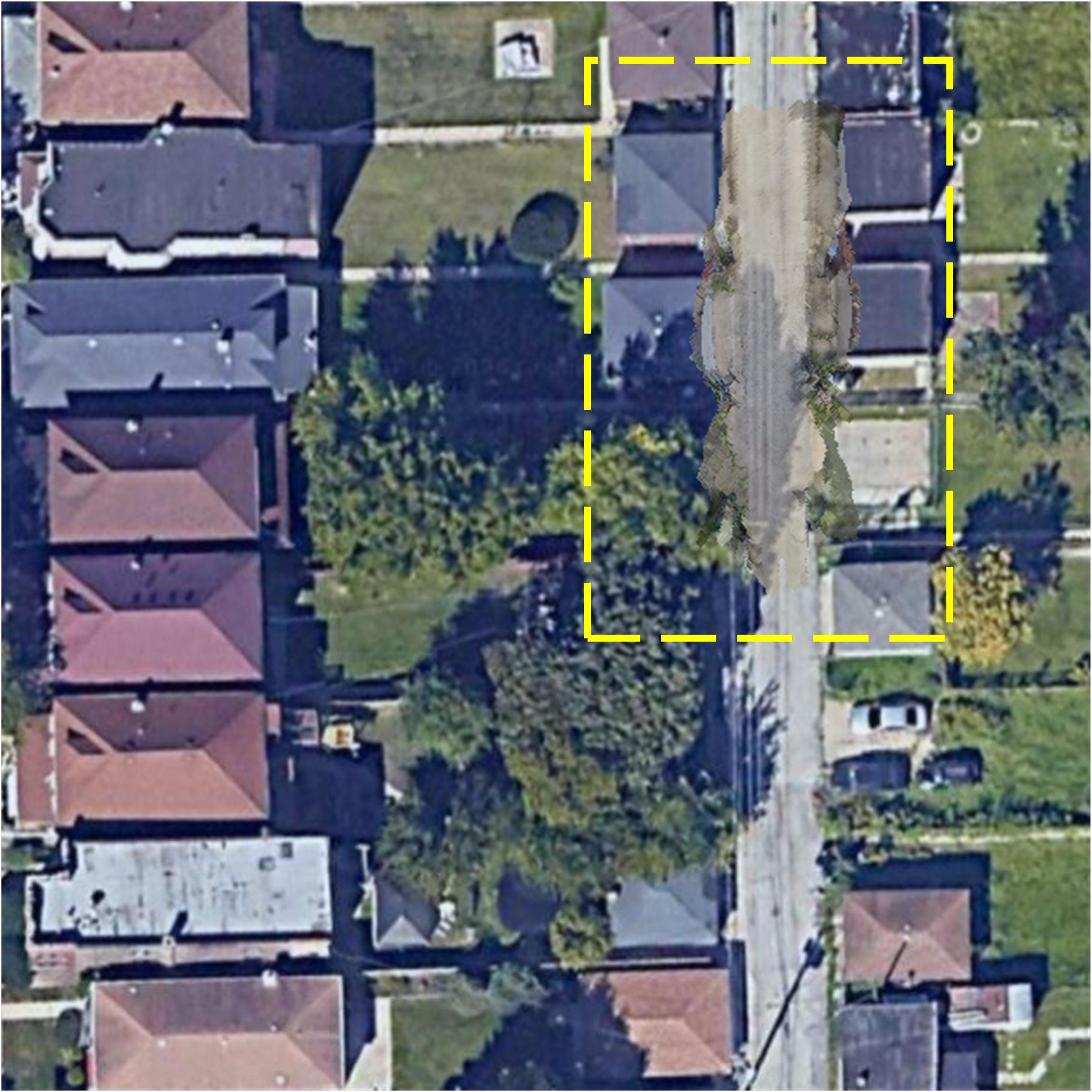}};
        \node[below right=2mm] at (a.north west) {(f)}; 
      }}
    \hfil
    \subfloat[]{%
    \tikz{\node (a) {\includegraphics{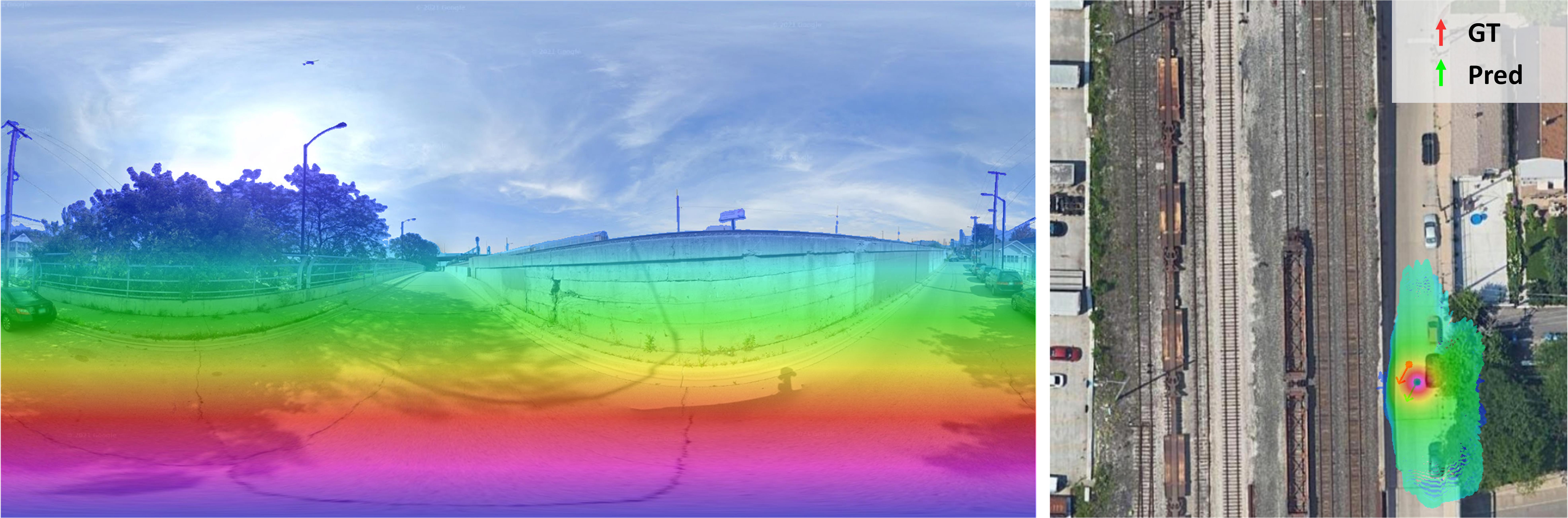}};
        \node[below right=2mm] at (a.north west) {(g)}; 
      }}
    \hfil
    \subfloat[]{%
    \tikz{\node (a) {\includegraphics{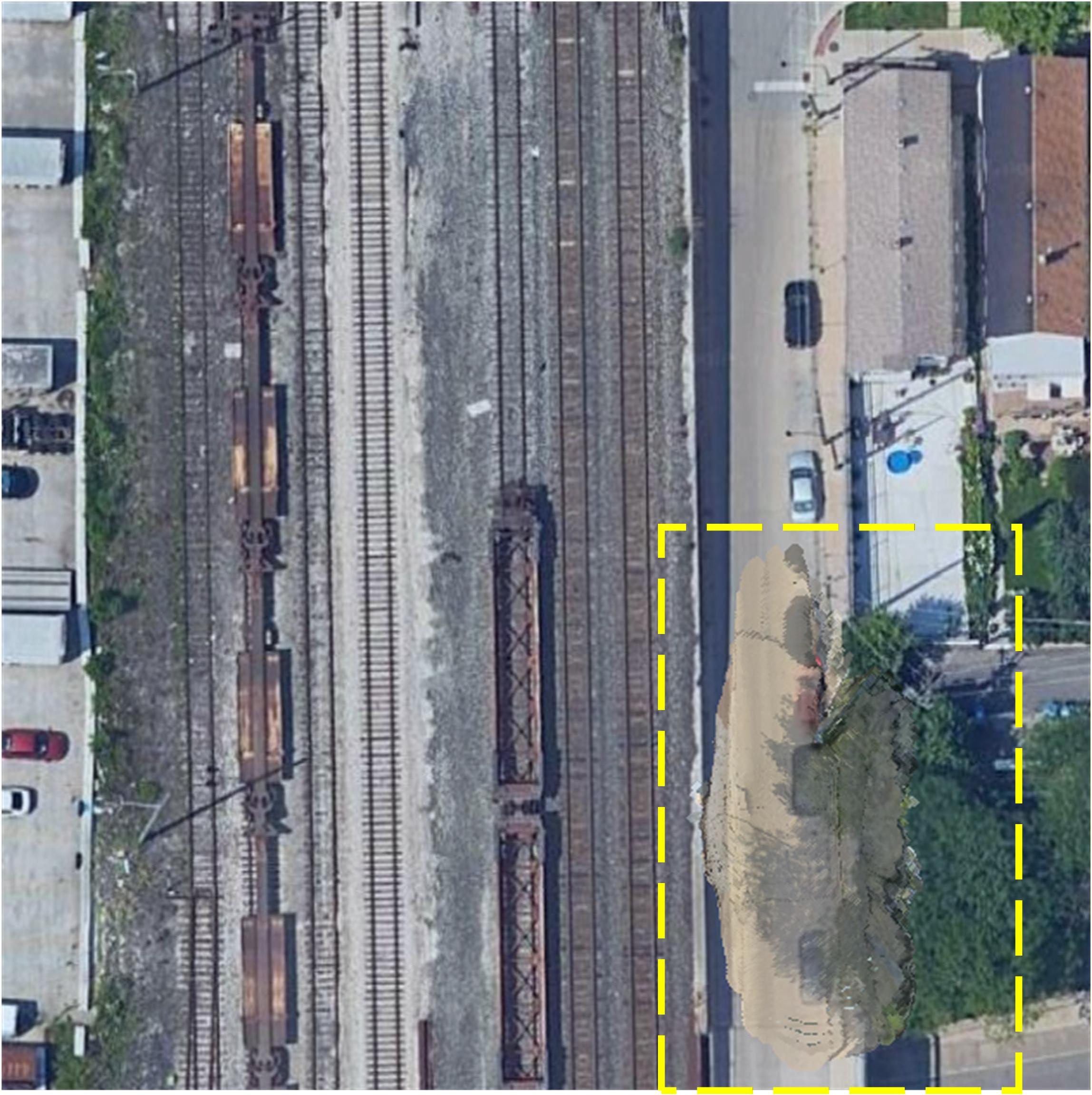}};
        \node[below right=2mm] at (a.north west) {(h)}; 
      }}
    \hfil
    \subfloat[]{%
    \tikz{\node (a) {\includegraphics{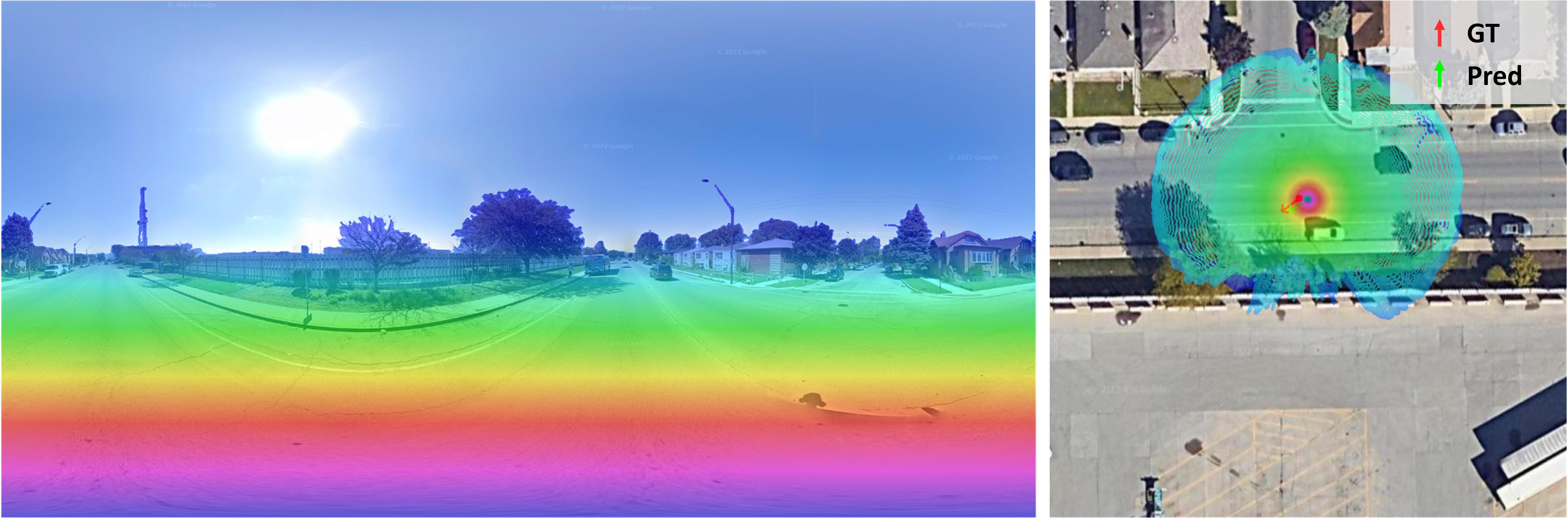}};
        \node[below right=2mm] at (a.north west) {(i)}; 
      }}
    \hfil
    \subfloat[]{%
    \tikz{\node (a) {\includegraphics{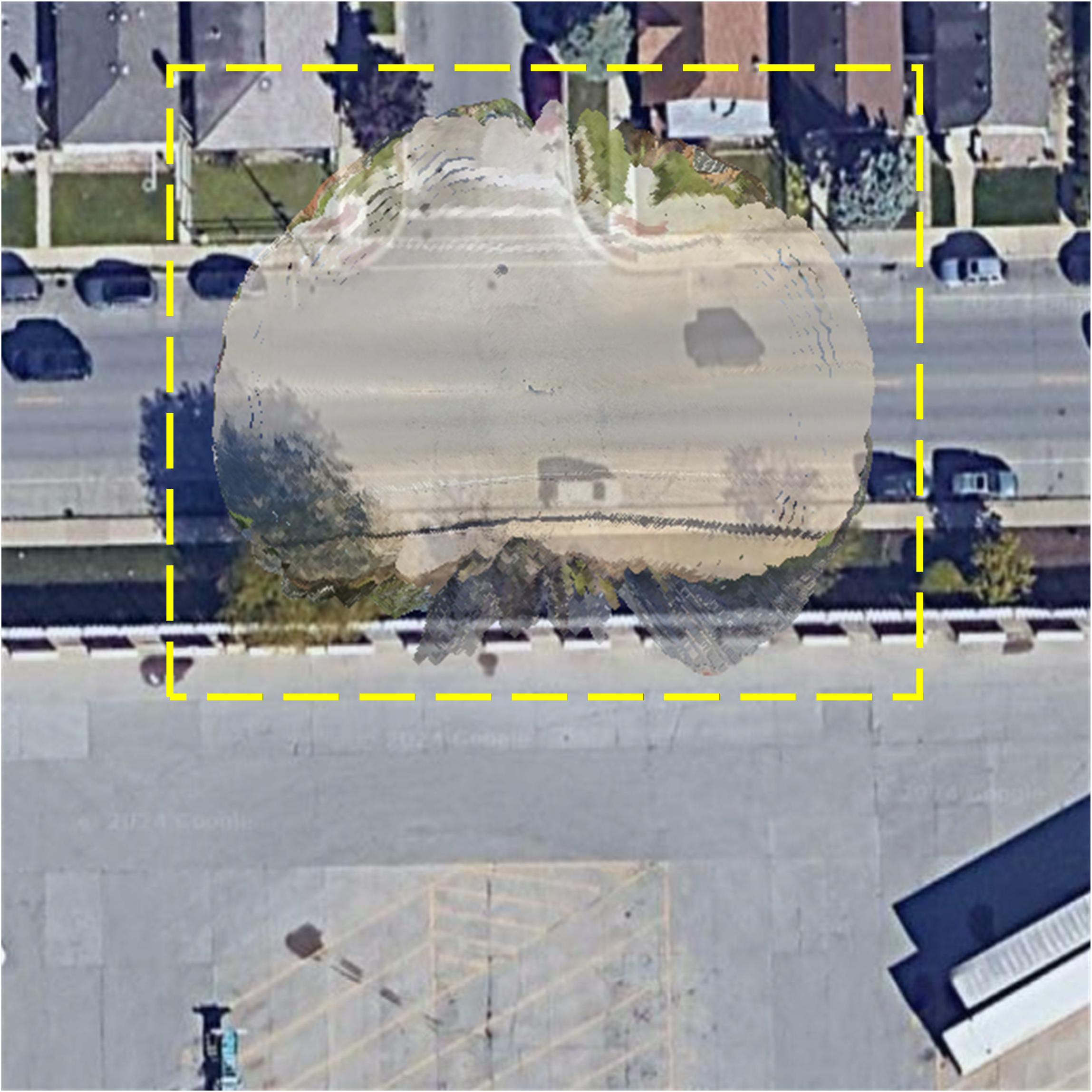}};
        \node[below right=2mm] at (a.north west) {(j)}; 
      }}
    \hfil
    \subfloat[]{%
    \tikz{\node (a) {\includegraphics{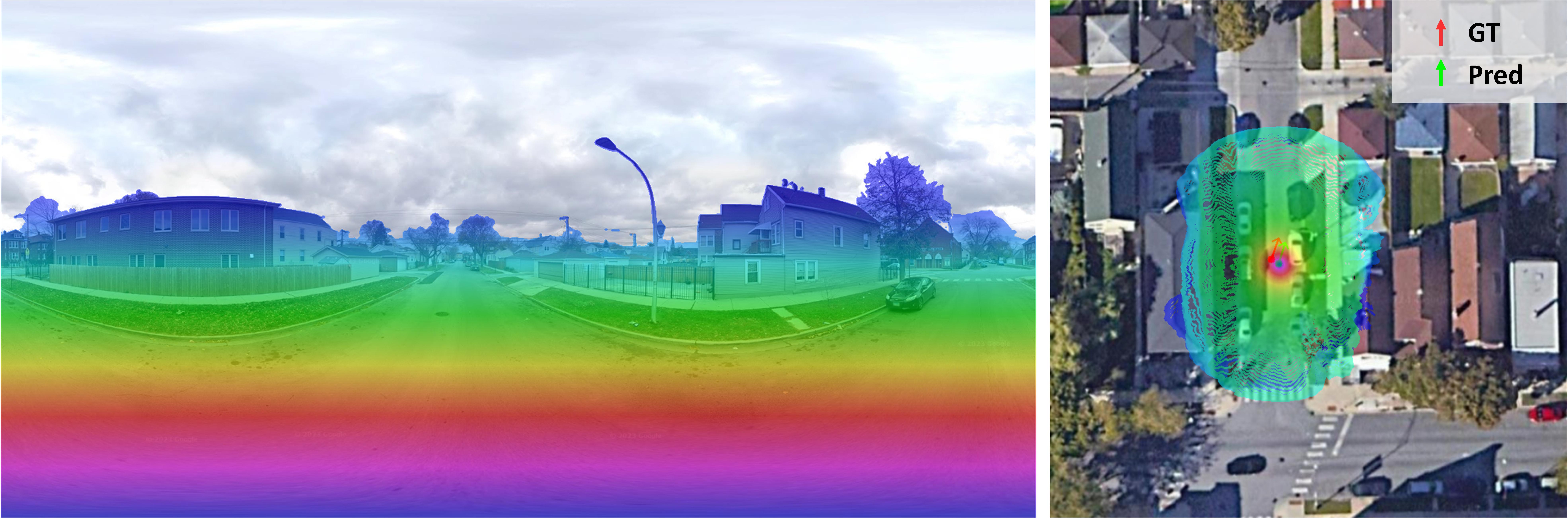}};
        \node[below right=2mm] at (a.north west) {(k)}; 
      }}
    \hfil
    \subfloat[]{%
    \tikz{\node (a) {\includegraphics{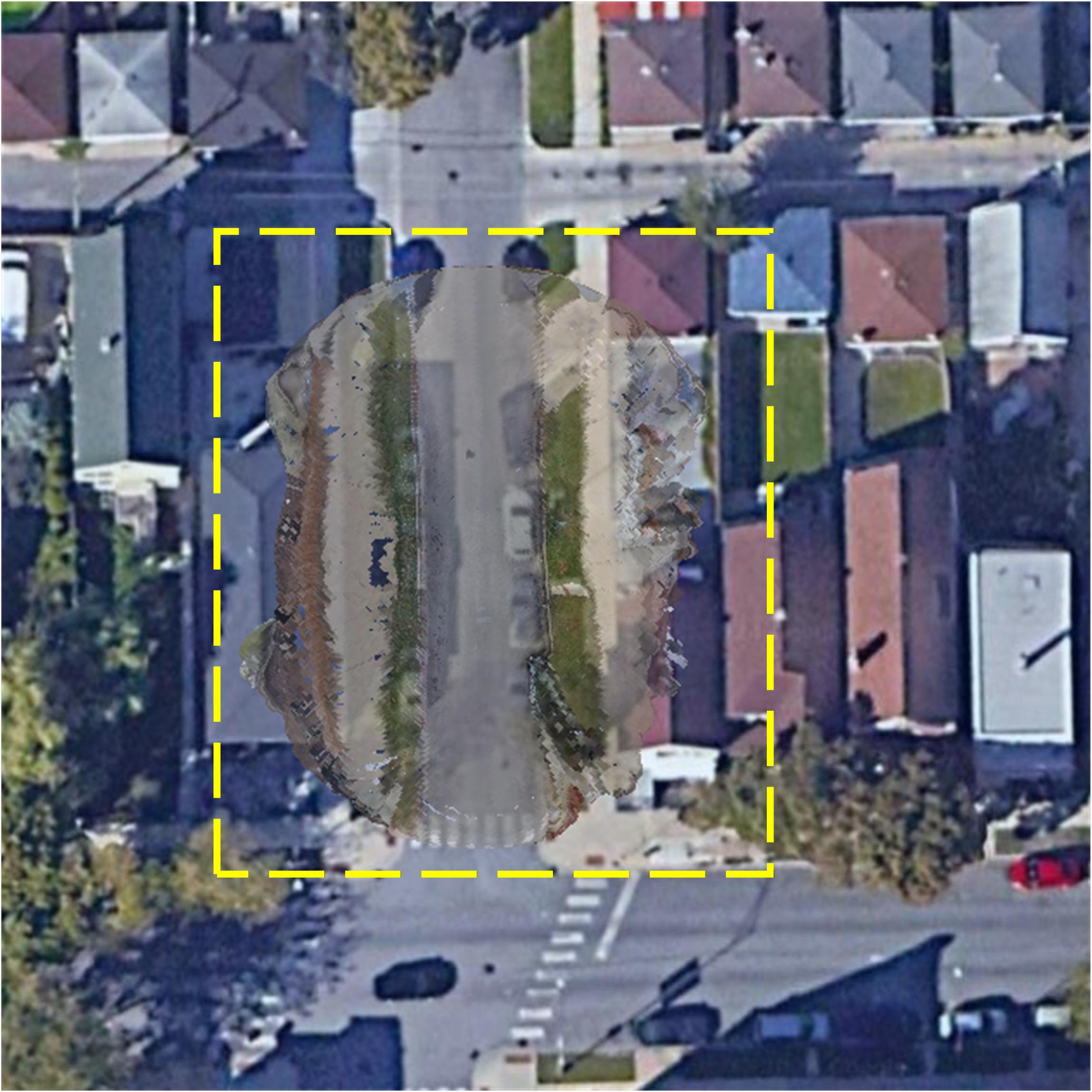}};
        \node[below right=2mm] at (a.north west) {(l)}; 
      }}
    \hfil
   \caption{
Qualitative results of dense cross-view image matching on the CVFM benchmark. 
Left: dense correspondence visualization. Right: projection onto satellite view with 40\% transparency.
}

    \label{fig:dense_vis}
\end{figure*}

\textbf{Qualitative results.}
Fig.~\ref{fig:dense_vis} presents representative visualizations of dense cross-view image matching. 
For each pair, the \textbf{left column} illustrates the dense matching map between the ground-view panorama and the satellite image, where color encodes the pixel-wise correspondence relationship across the two views. 
The \textbf{right column} shows the corresponding satellite projection obtained by reprojecting all non-sky pixels from the ground panorama according to the estimated 3D geometry, rendered with 40\% transparency for visual clarity. 
From these visualizations, the boundary alignment and overlapping regions in the projected images reveal that the dense matching results are geometrically coherent and show strong potential for fine-grained pixel-level cross-view alignment.

\textbf{Limitations.}
Although our framework exhibits promising potential for dense cross-view image matching, its current performance still falls short of practical deployment. 
The accuracy of dense correspondences is largely constrained by the quality of monocular depth estimated from panoramic ground views. 
Existing depth estimators often struggle with panoramic distortion and complex materials such as glass façades or vegetation, leading to local geometric inconsistencies when projecting ground pixels onto the satellite domain. 
These limitations hinder the reliability of fine-grained alignment. 
Future work could explore cross-view depth optimization or geometry-aware regularization to mitigate such distortions and unlock the full potential of dense cross-view image matching.

\section{Cross-Dataset Transferability}
\label{sup:sec:crossdataset}
To further evaluate the generalization ability of our framework, we conduct a cross-dataset transfer experiment from DReSS-D~\cite{zhang2025cross} to VIGOR~\cite{zhu2021vigor}. 
The model is trained on DReSS-D and directly tested on VIGOR without any fine-tuning, forming a zero-shot transfer setting that examines robustness under domain shifts in city layout, imaging conditions, and data distribution.

As reported in Tab.~\ref{tab:vigor_transferability}, our method exhibits strong cross-dataset generalization. 
In the known-orientation setting, it achieves a mean localization error of 2.42\,m, outperforming FG2 by 0.16\,m. 
In the more challenging unknown-orientation setting, our model reduces the mean translation error from 4.88\,m to 3.98\,m, an improvement of nearly 20\%. 
Despite being trained on a different dataset, our model maintains accuracy close to its in-domain performance (1.89\,m vs. 2.42\,m). 
These results suggest that the proposed Surface Model and SimRefiner effectively provide geometry-aware, transferable representations that generalize across diverse urban domains.

\begin{table}[t]
    \centering
    \caption{
Cross-dataset transfer results on the VIGOR benchmark. 
${\dagger\dagger}$ indicates models trained on the DReSS-D dataset and evaluated on VIGOR in a zero-shot manner, without fine-tuning. 
All results are reported under the same evaluation protocol as in Tab.~\ref{tab:vigor_ransac}.
}
    \begin{tabular}{p{1.2cm}p{1.3cm}p{0.8cm}p{0.8cm}p{0.8cm}p{0.8cm}}
    \toprule
    \multirow{3}{*}{{Orien.}} & \multirow{3}{*}{Methods} & 
    \multicolumn{4}{c}{Same-area} \\
    \cline{3-6} 
    & & \multicolumn{2}{c}{$\downarrow$ Loc (m)} & \multicolumn{2}{c}{$\downarrow$ Orien ($^\circ$)} \\
    \cline{3-6} 
    & & Mean & Median & Mean & Median \\
    \hline
    \multirow{4}{*}{Known}& 
    FG2& 2.10& 1.03& -& -\\
 & Ours & 1.89& 0.97& -& -\\
 & FG2$^{\dagger\dagger}$& 2.58& 1.29& -& -\\
 & Ours$^{\dagger\dagger}$& 2.42& 1.26& -& -\\

    \hline
    \multirow{4}{*}{Unknown}& 
    FG2& 3.90& 2.22& 7.07& 1.83\\
 & Ours & 3.11& 1.31& 8.65& 2.69\\
 & FG2$^{\dagger\dagger}$& 4.88& 2.52& 11.76& 2.01\\
 & Ours$^{\dagger\dagger}$& 3.98& 1.96& 13.47& 2.22\\
 \bottomrule
    \end{tabular}
    \label{tab:vigor_transferability}
\end{table}

\section{Limitations of Cross-View Image Matching}
\label{sup:sec:limitations}
Although our framework achieves strong localization accuracy and robust cross-dataset generalization, several limitations remain in its image matching component.  
First, the current design is not a strictly geometry-consistent matching framework.  
Dense correspondences are generated under the simplifying assumption that the ground-view camera center is approximately 2-3\,m above the ground.  
This assumption has little impact on 3-DoF localization because only planar projection is required, but it can introduce minor spatial bias in pixel-level matching, particularly in scenes with significant height variations or complex occlusions.

Second, monocular depth estimation from ground panoramas may cause point cloud deformation in regions with reflective surfaces or dense vegetation.  
These geometric distortions propagate through the projection process and limit the attainable precision of dense correspondences.  
Future research could address these challenges by explicitly modeling cross-view geometric relationships and jointly optimizing depth, correspondence, and pose.  
Such unified optimization may further strengthen the mutual benefits between localization and dense image matching.

\section{Additional Qualitative Results}
\label{sup:sec:qualitative}
We provide additional qualitative results to complement the visualizations in the main paper. 

\begin{figure*}[ht]
    \captionsetup[subfigure]{labelformat=empty}
    \tikzset{inner sep=0pt}
    \setkeys{Gin}{width=0.49\textwidth}
    \centering
    \subfloat[]{%
    \tikz{\node (a) {\includegraphics{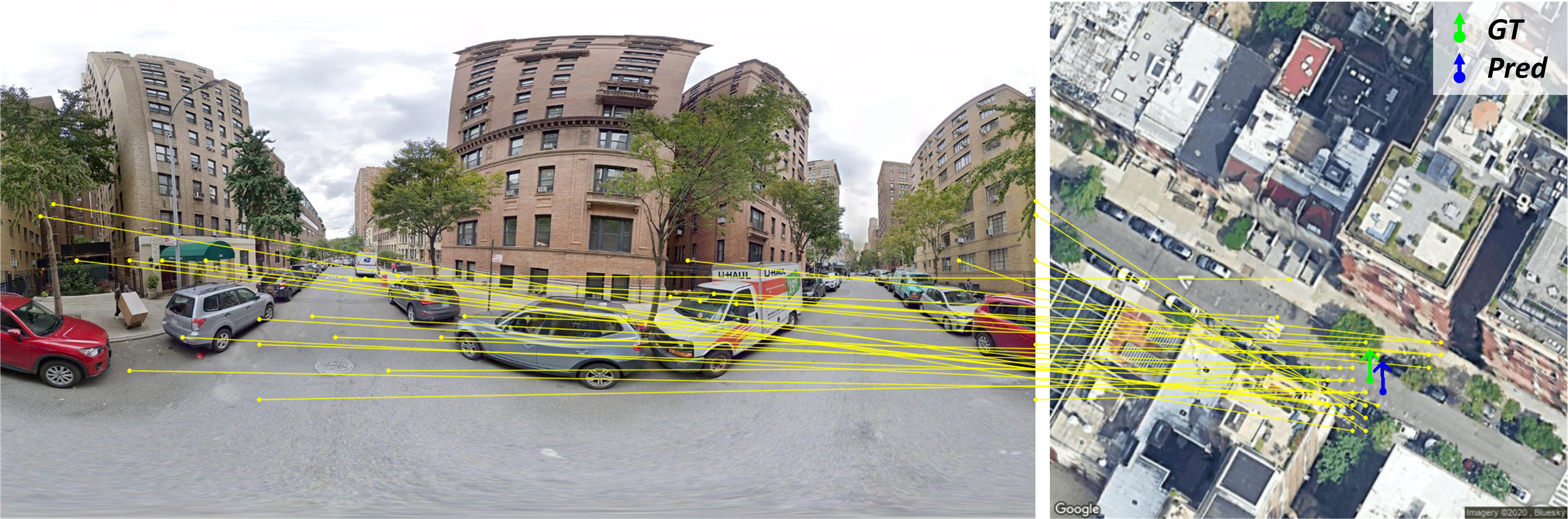}};
        \node[below right=2mm] at (a.north west) {(a)}; 
      }}
    \hfil
    \subfloat[]{%
    \tikz{\node (a) {\includegraphics{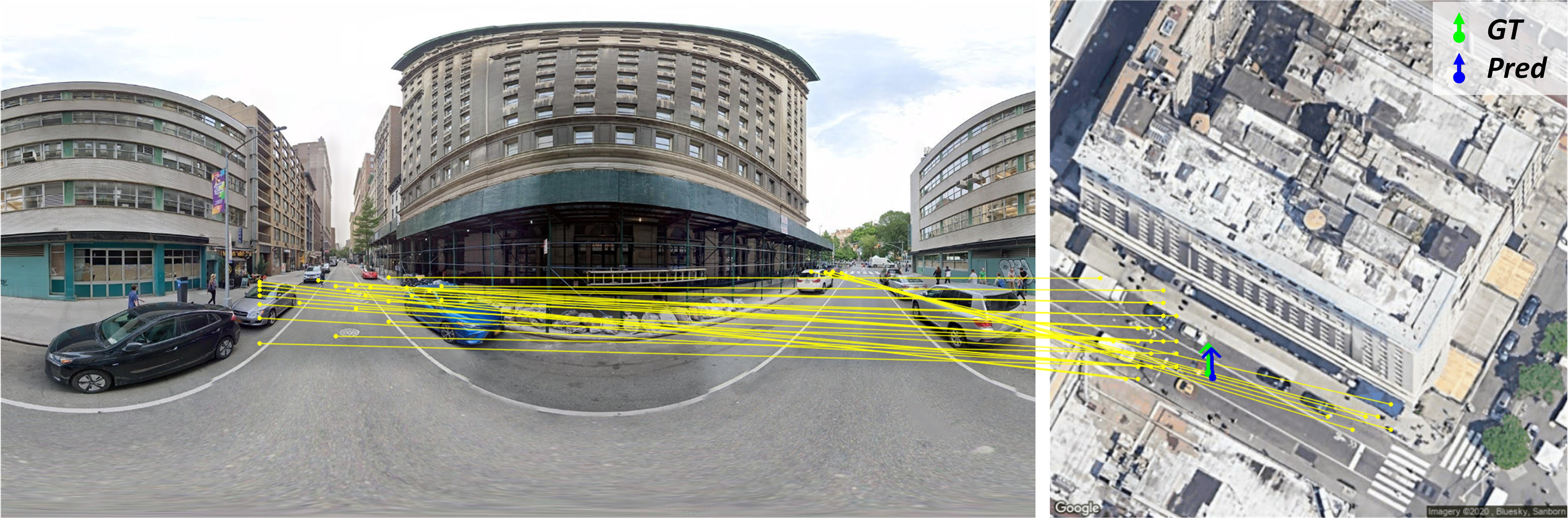}};
        \node[below right=2mm] at (a.north west) {(b)}; 
      }}
    \hfil
    \subfloat[]{%
    \tikz{\node (a) {\includegraphics{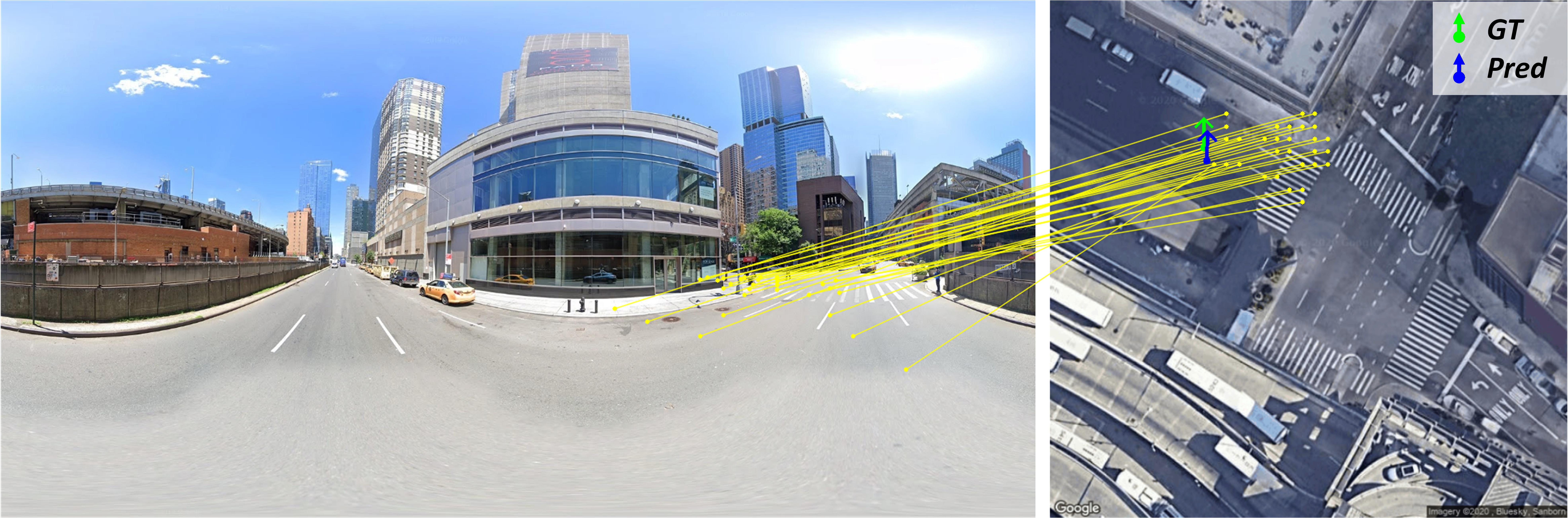}};
        \node[below right=2mm] at (a.north west) {(c)}; 
      }}
    \hfil
    \subfloat[]{%
    \tikz{\node (a) {\includegraphics{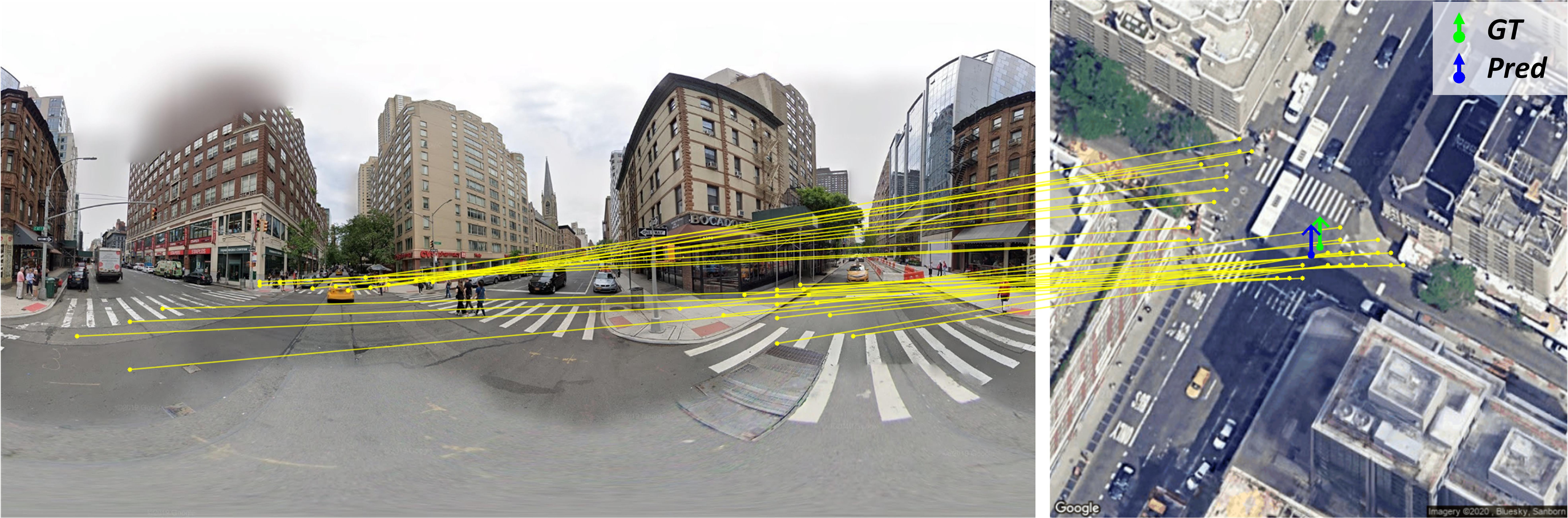}};
        \node[below right=2mm] at (a.north west) {(d)}; 
      }}
    \hfil
    \subfloat[]{%
    \tikz{\node (a) {\includegraphics{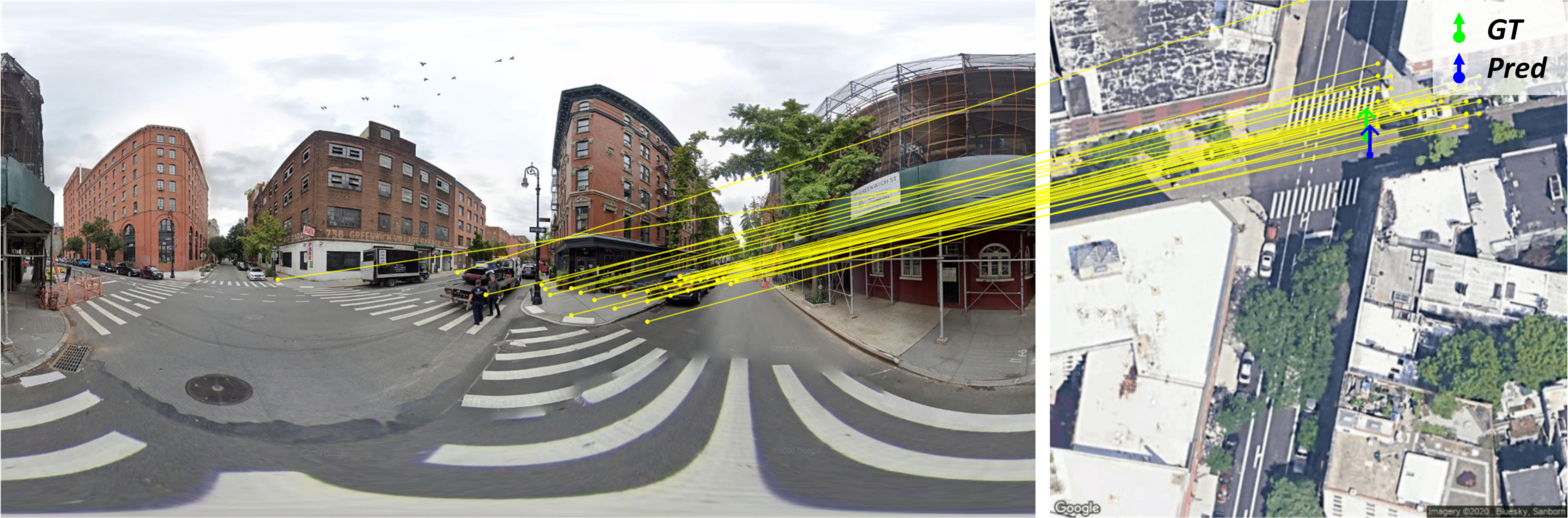}};
        \node[below right=2mm] at (a.north west) {(e)}; 
          }}
    \hfil
    \subfloat[]{%
    \tikz{\node (a) {\includegraphics{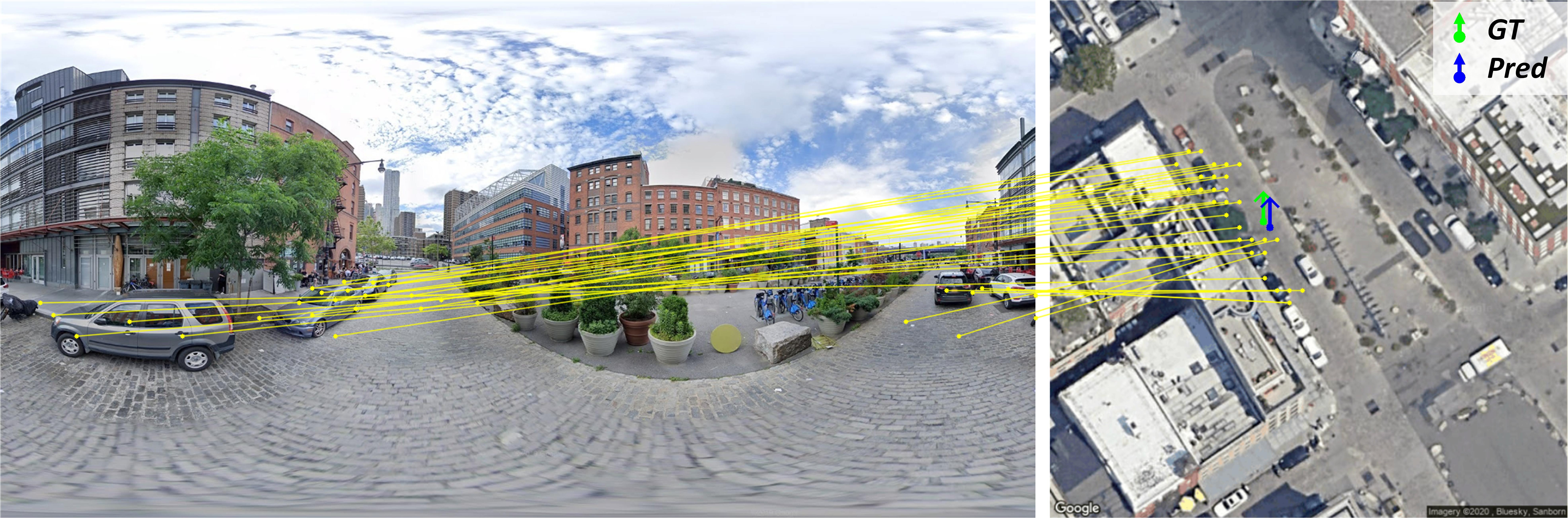}};
        \node[below right=2mm] at (a.north west) {(f)}; 
      }}
    \hfil
    \subfloat[]{%
    \tikz{\node (a) {\includegraphics{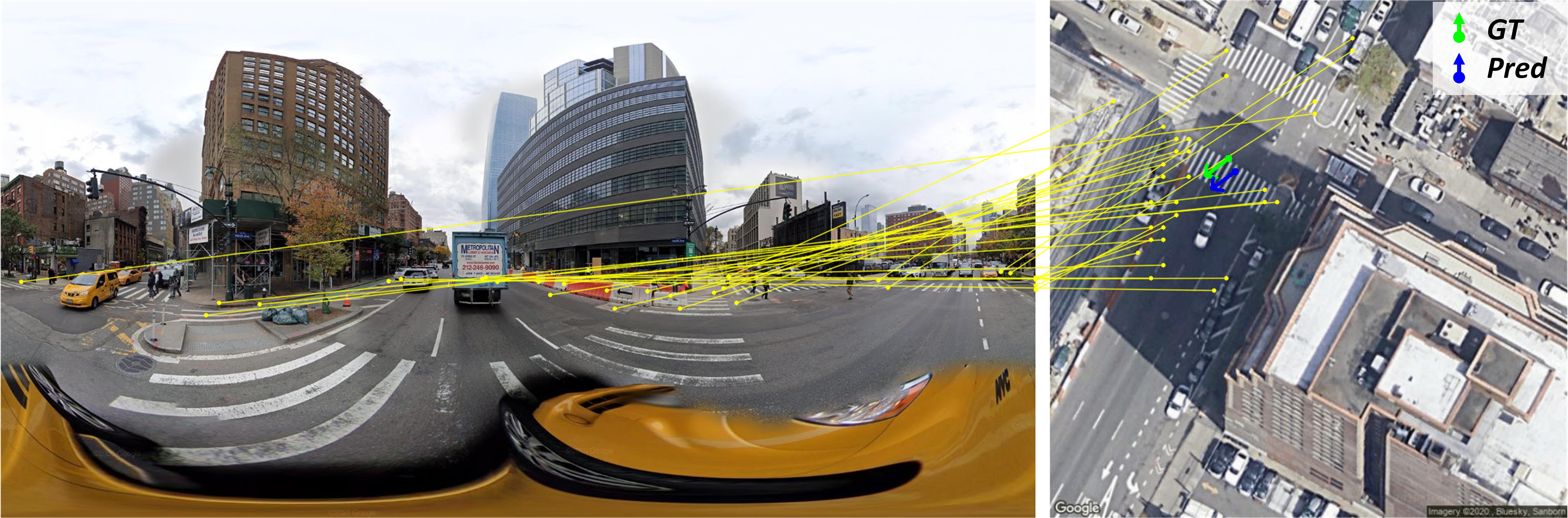}};
        \node[below right=2mm] at (a.north west) {(g)}; 
      }}
    \hfil
    \subfloat[]{%
    \tikz{\node (a) {\includegraphics{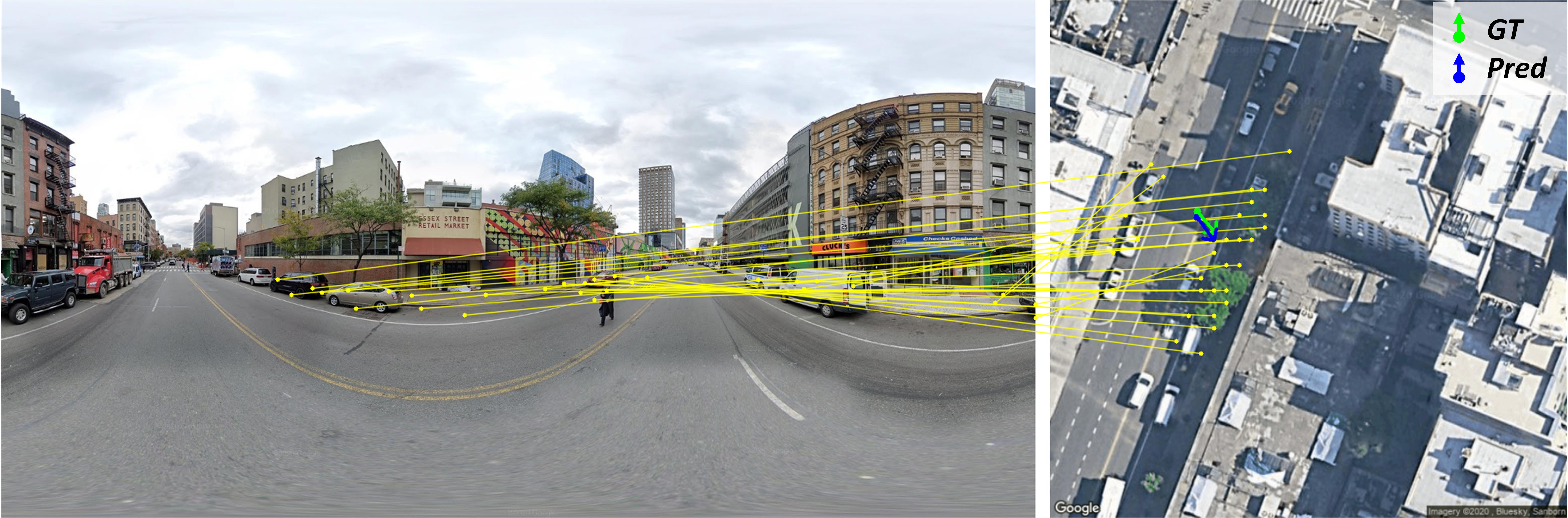}};
        \node[below right=2mm] at (a.north west) {(h)}; 
      }}
    \hfil
    \subfloat[]{%
    \tikz{\node (a) {\includegraphics{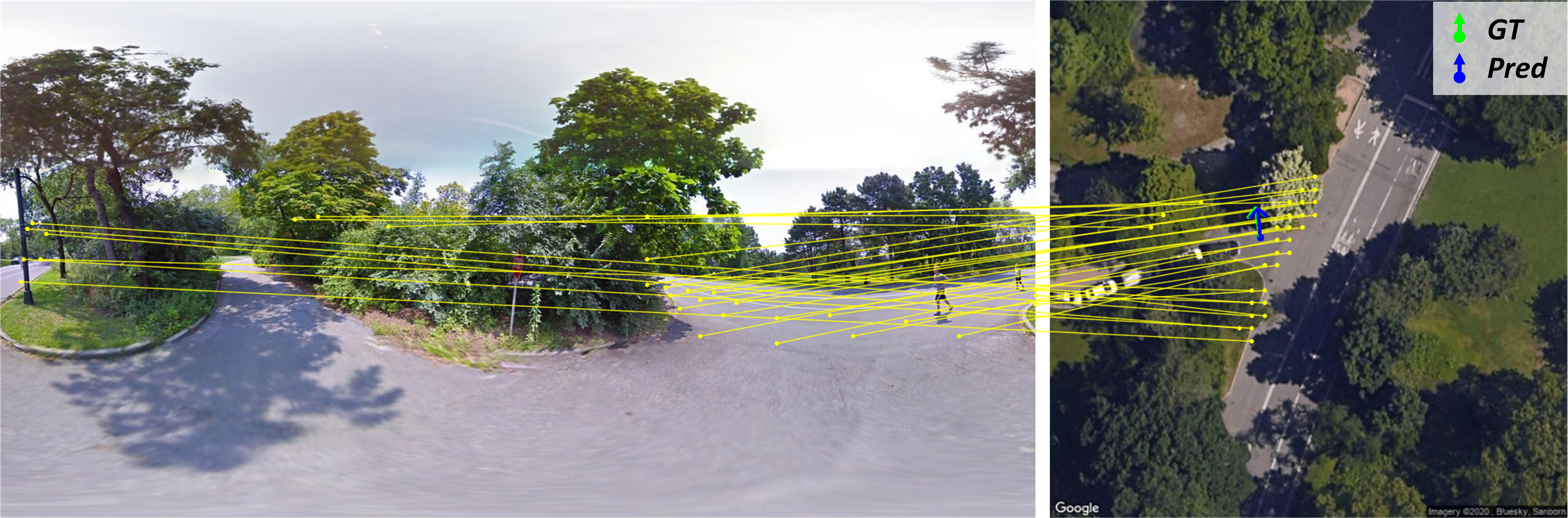}};
        \node[below right=2mm] at (a.north west) {(i)}; 
      }}
    \hfil
    \subfloat[]{%
    \tikz{\node (a) {\includegraphics{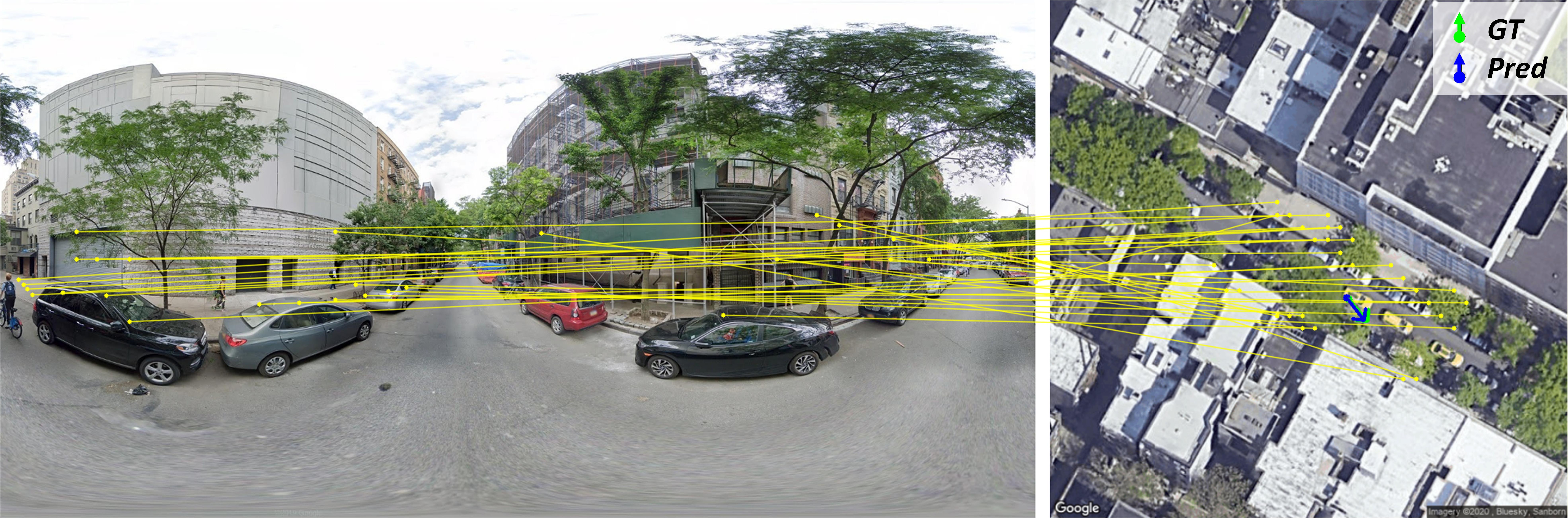}};
        \node[below right=2mm] at (a.north west) {(j)}; 
      }}
    \hfil
    \subfloat[]{%
    \tikz{\node (a) {\includegraphics{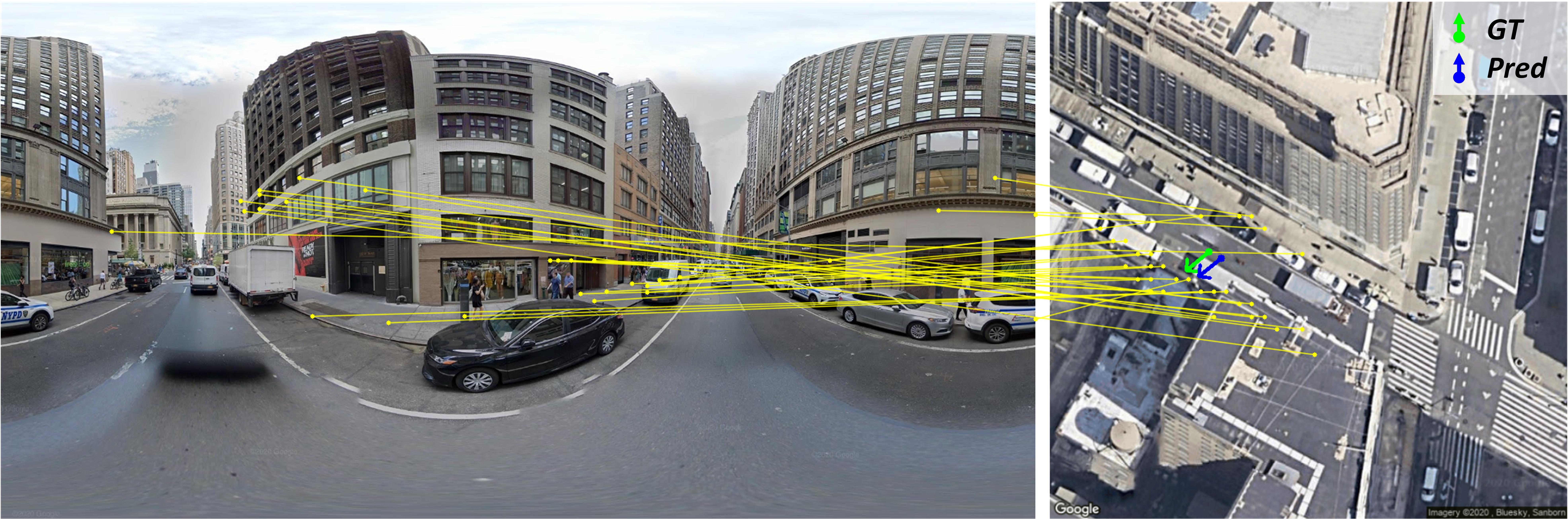}};
        \node[below right=2mm] at (a.north west) {(k)}; 
      }}
    \hfil
    \subfloat[]{%
    \tikz{\node (a) {\includegraphics{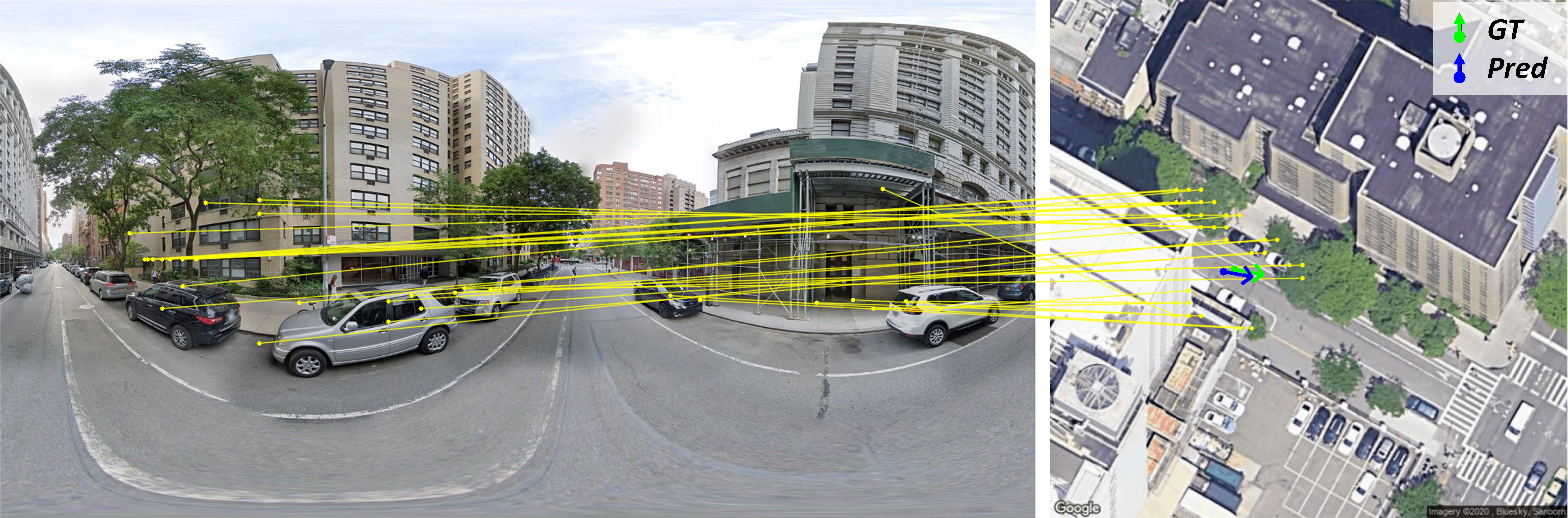}};
        \node[below right=2mm] at (a.north west) {(l)}; 
      }}
    \hfil
    \caption{
Additional qualitative results on cross-view localization using the VIGOR dataset. 
The top three rows correspond to the known-orientation setting, while the bottom three rows show results under the unknown-orientation setting. 
}
    \label{fig:extra_loc}
\end{figure*}

\textbf{Cross-view localization.}  
Fig.~\ref{fig:extra_loc} presents representative examples of cross-view localization on the VIGOR dataset. 
Each example includes the input ground-view panorama, the corresponding satellite image, and the predicted pose overlay. 
The top three rows show results under the \textit{known-orientation} setting, while the bottom three rows illustrate the more challenging \textit{unknown-orientation} setting.  
Our method accurately aligns the ground-view image with the correct satellite region, even under large viewpoint changes. 
It also maintains consistent orientation estimation and precise translational alignment across diverse urban scenes, demonstrating strong robustness to severe appearance variations and geometric distortions.

\begin{figure*}[ht]
    \captionsetup[subfigure]{labelformat=empty}
    \tikzset{inner sep=0pt}
    \setkeys{Gin}{width=0.49\textwidth}
    \centering
    \subfloat[]{%
    \tikz{\node (a) {\includegraphics{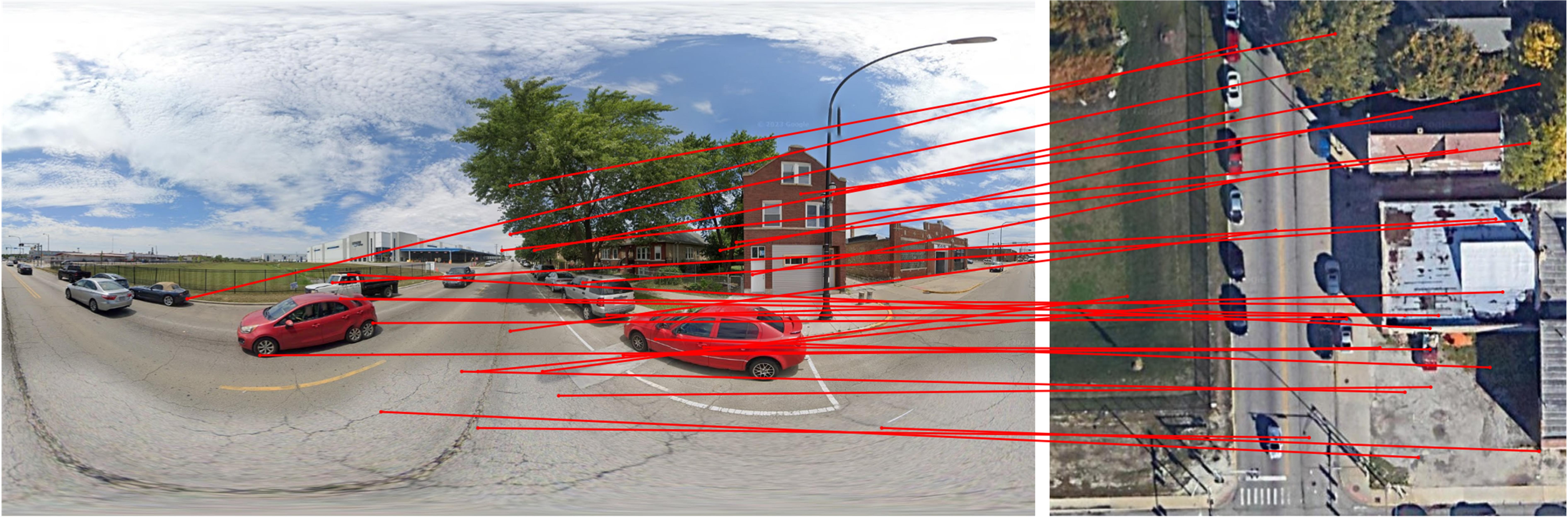}};
        \node[below right=2mm] at (a.north west) {LoFTR}; 
      }}
    \hfil
    \subfloat[]{%
    \tikz{\node (a) {\includegraphics{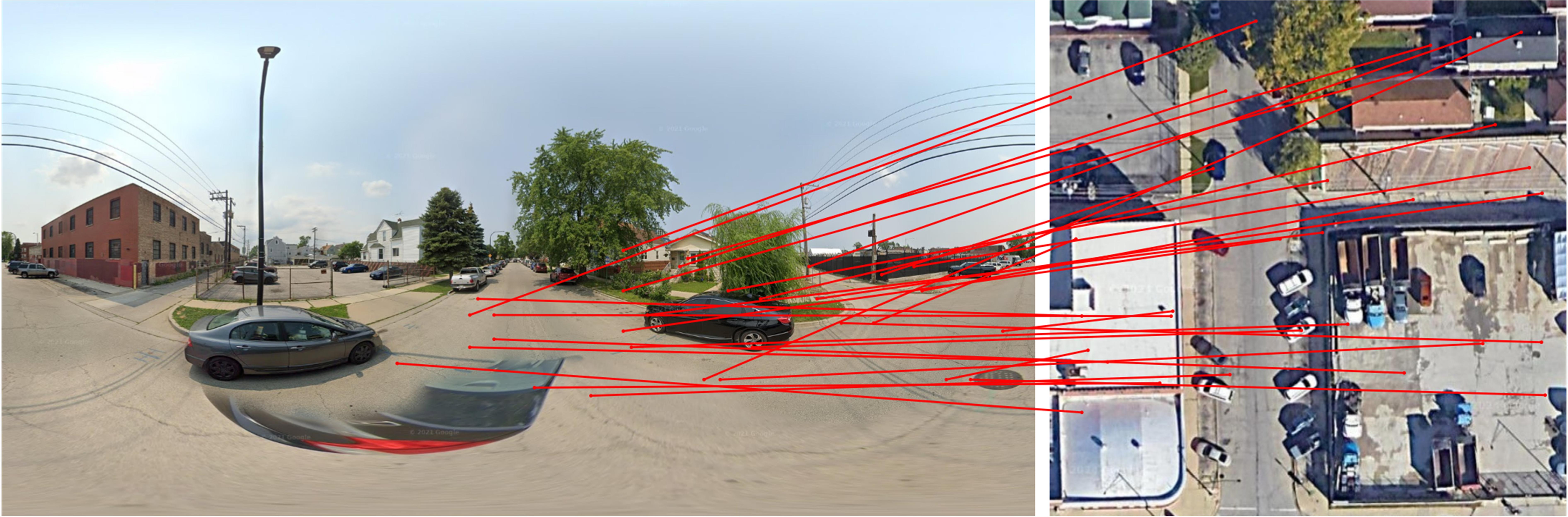}};
        \node[below right=2mm] at (a.north west) {LoFTR}; 
      }}
    \hfil
    \subfloat[]{%
    \tikz{\node (a) {\includegraphics{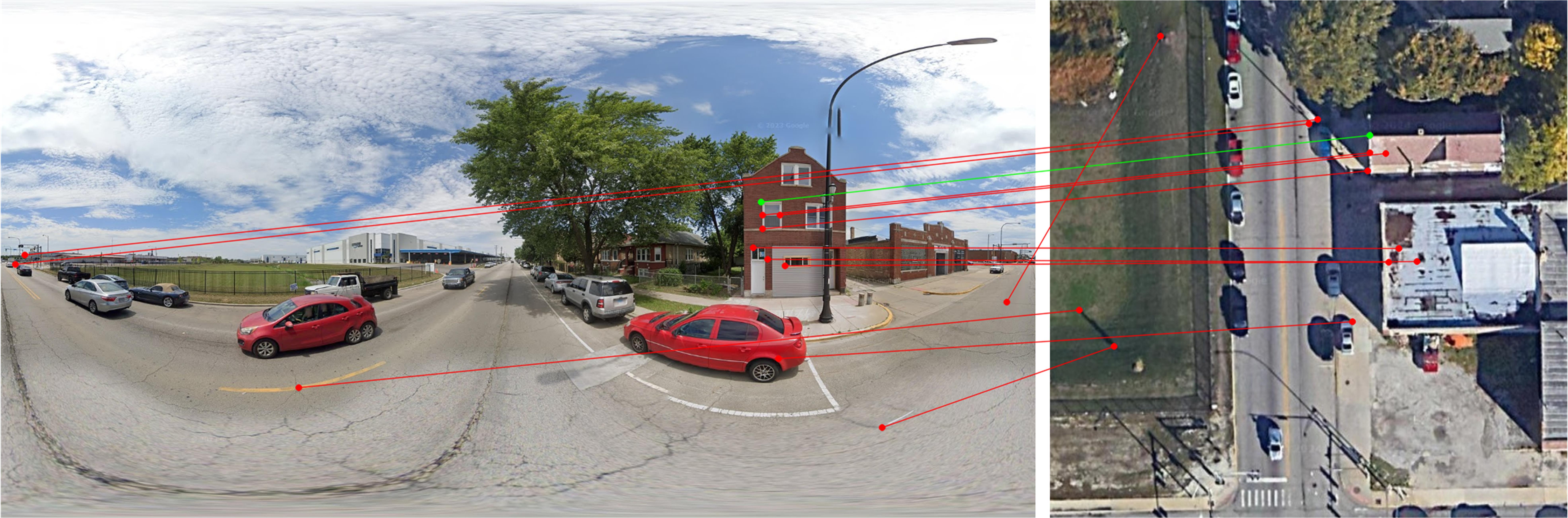}};
        \node[below right=2mm] at (a.north west) {SuperGlue}; 
      }}
    \hfil
    \subfloat[]{%
    \tikz{\node (a) {\includegraphics{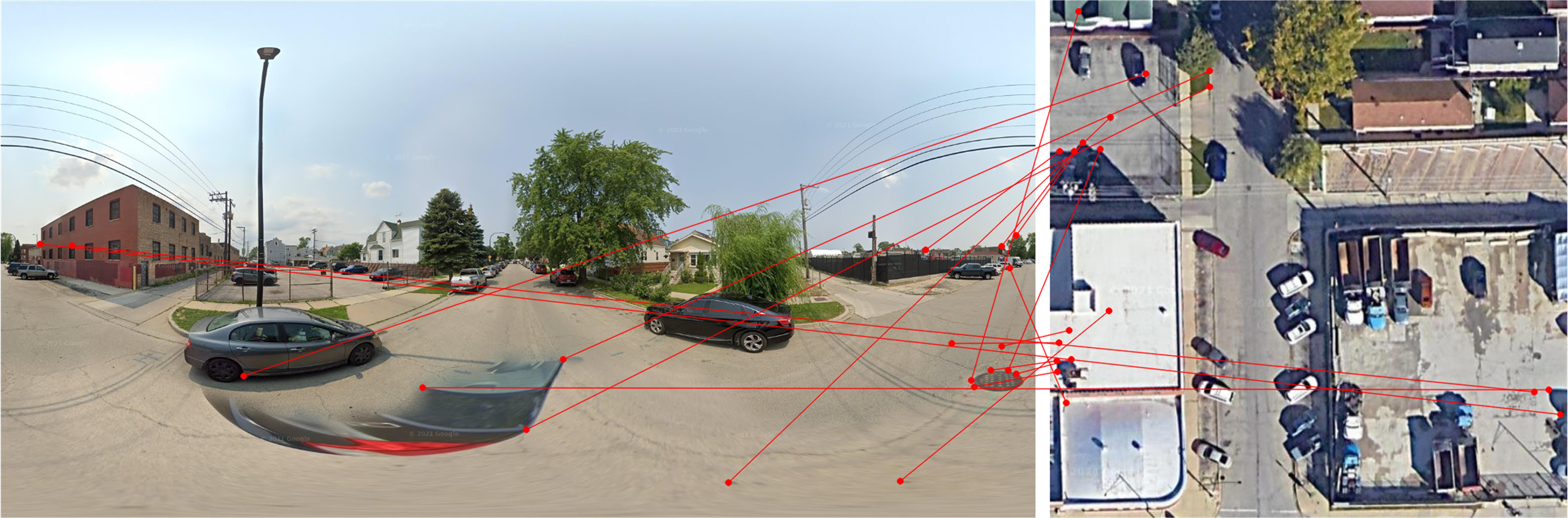}};
        \node[below right=2mm] at (a.north west) {SuperGlue}; 
      }}
    \hfil
    \subfloat[]{%
    \tikz{\node (a) {\includegraphics{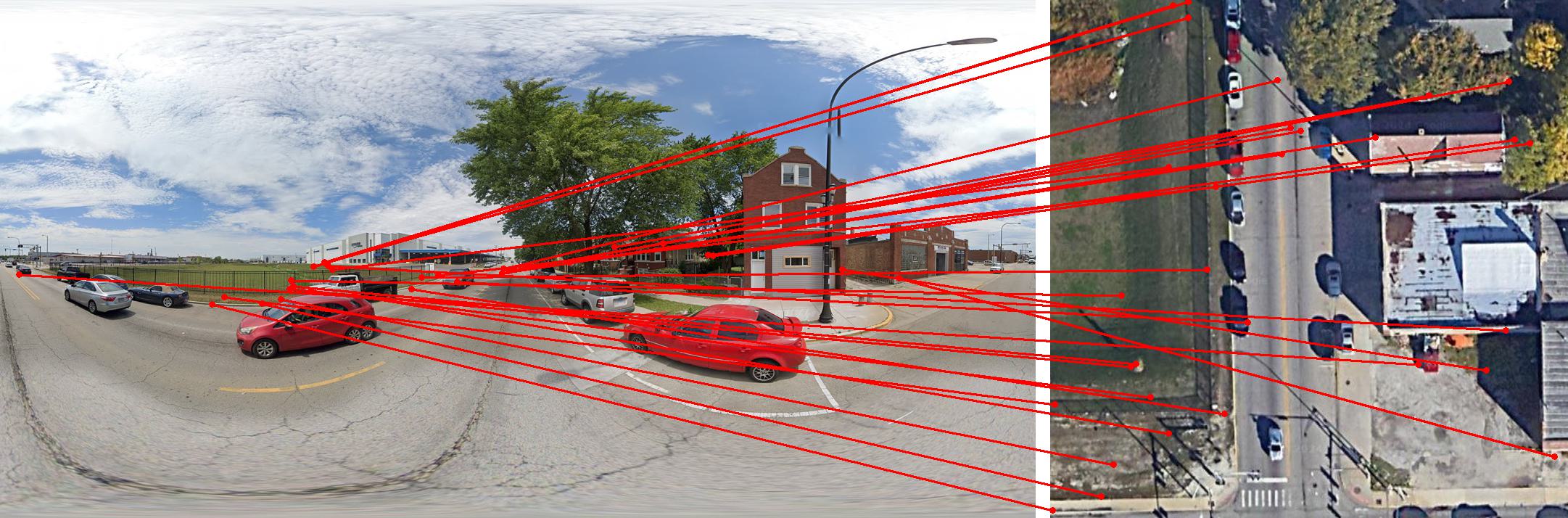}};
        \node[below right=2mm] at (a.north west) {RoMa}; 
          }}
    \hfil
    \subfloat[]{%
    \tikz{\node (a) {\includegraphics{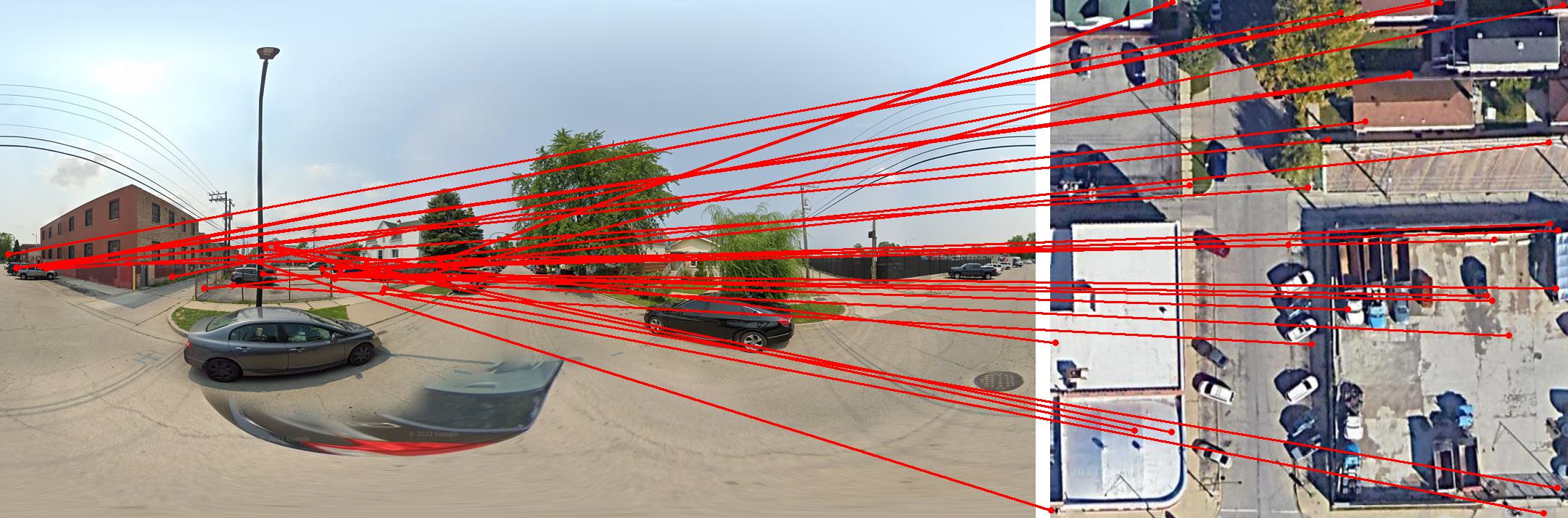}};
        \node[below right=2mm] at (a.north west) {RoMa}; 
      }}
    \hfil
    \subfloat[]{%
    \tikz{\node (a) {\includegraphics{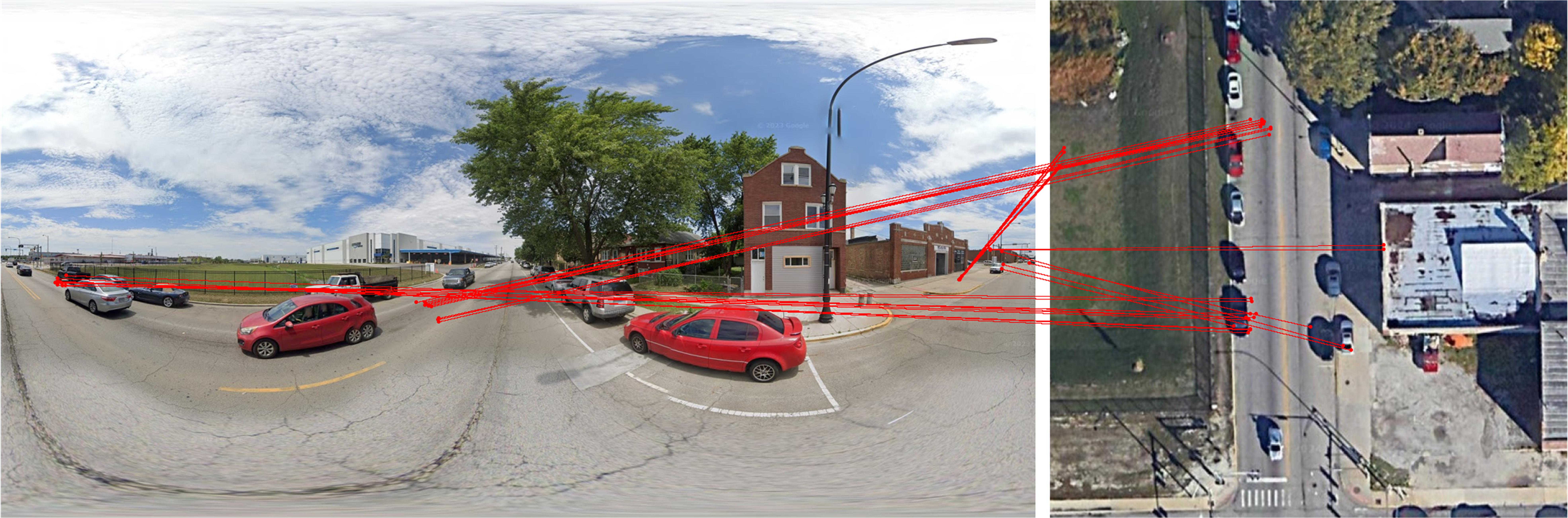}};
        \node[below right=2mm] at (a.north west) {Aerial-Megadepth}; 
      }}
    \hfil
    \subfloat[]{%
    \tikz{\node (a) {\includegraphics{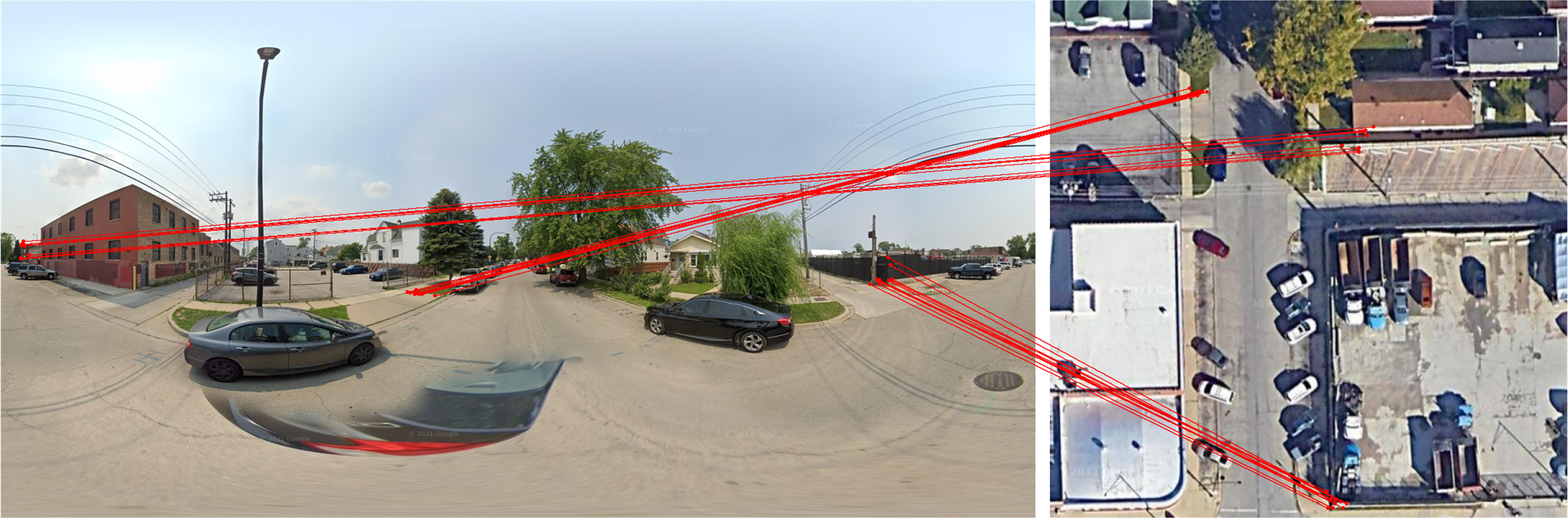}};
        \node[below right=2mm] at (a.north west) {Aerial-Megadepth}; 
      }}
    \hfil
    \subfloat[]{%
    \tikz{\node (a) {\includegraphics{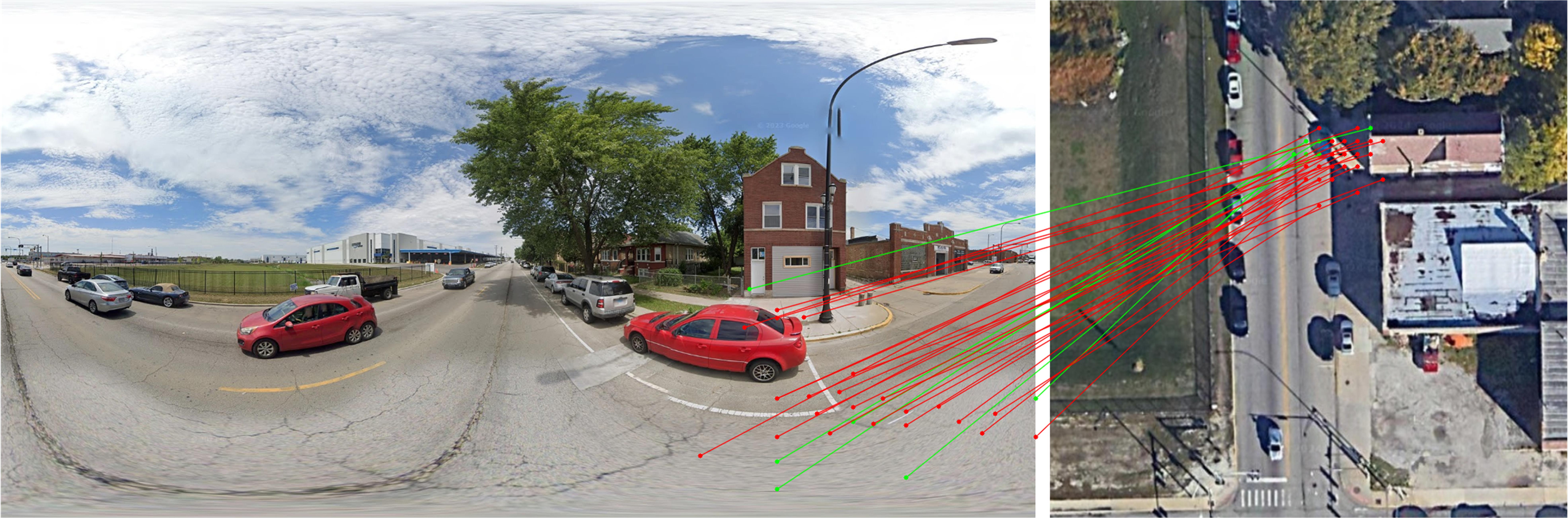}};
        \node[below right=2mm] at (a.north west) {FG2}; 
      }}
    \hfil
    \subfloat[]{%
    \tikz{\node (a) {\includegraphics{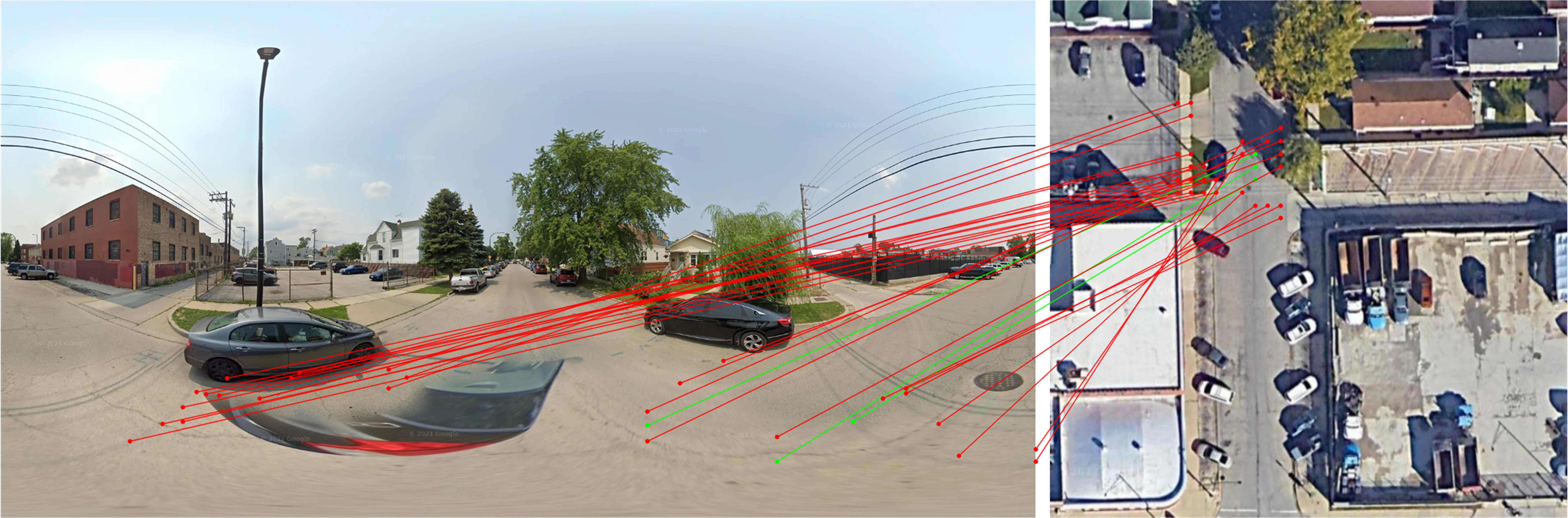}};
        \node[below right=2mm] at (a.north west) {FG2}; 
      }}
    \hfil
    \subfloat[]{%
    \tikz{\node (a) {\includegraphics{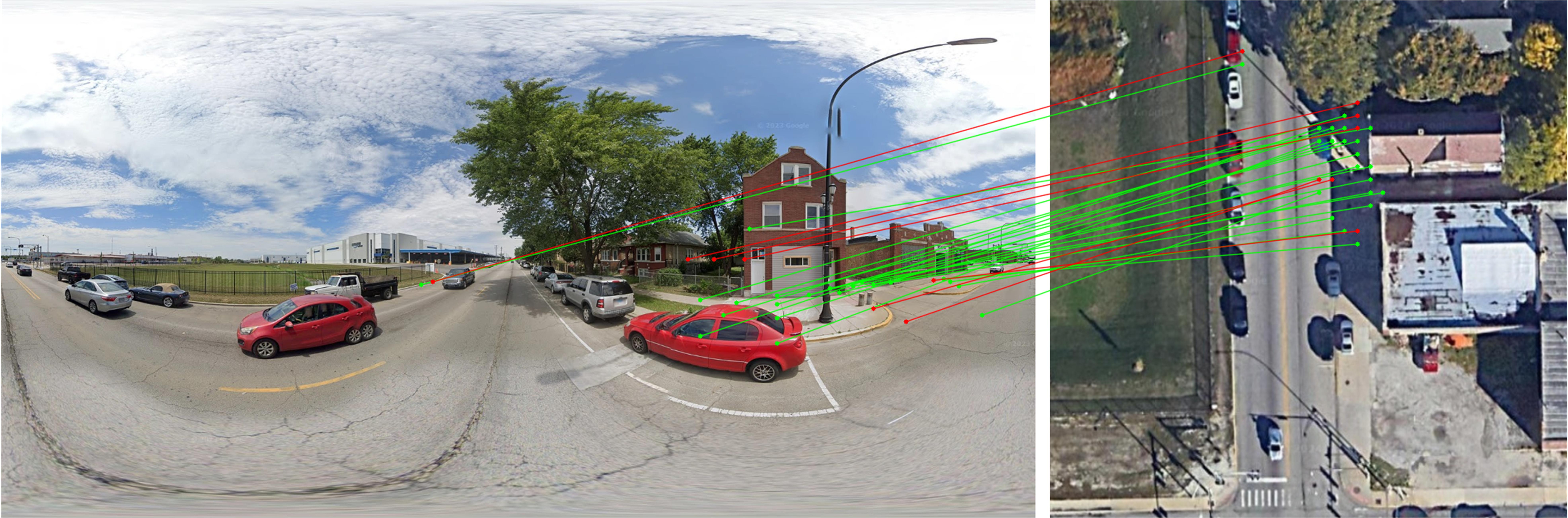}};
        \node[below right=2mm] at (a.north west) {Ours}; 
      }}
    \hfil
    \subfloat[]{%
    \tikz{\node (a) {\includegraphics{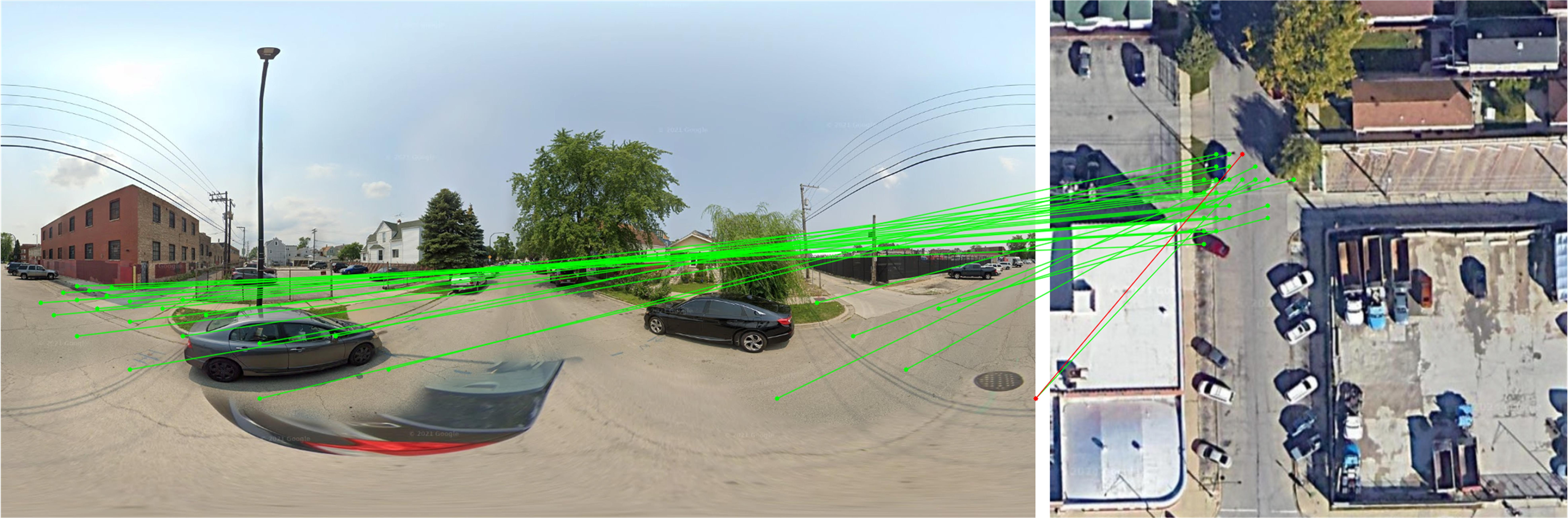}};
        \node[below right=2mm] at (a.north west) {Ours}; 
      }}
    \hfil
   \caption{
Comparison of different cross-view image matching methods on the CVFM benchmark. 
Shown are LoFTR, SuperGlue, RoMa, Aerial-Megadepth, FG2, and our method. 
For Aerial-Megadepth, ground panoramas are reprojected into three perspective images with a $120^{\circ}$ field of view, matched separately, and the top 30 correspondences (out of 90) are reprojected back for visualization. 
Green and red lines denote correct and incorrect correspondences under a 15-pixel threshold. 
}

    \label{fig:extra_match_compare}
\end{figure*}

\textbf{Comparison of image matching methods.}  
Fig.~\ref{fig:extra_match_compare} compares different cross-view image matching methods evaluated on the CVFM benchmark, including LoFTR~\cite{sun2021loftr}, SuperGlue~\cite{sarlin2020superglue}, RoMa~\cite{edstedt2024roma}, Aerial-Megadepth~\cite{vuong2025aerialmegadepth}, FG2~\cite{xia2025fg}, and our approach. 
Green and red lines indicate correct and incorrect correspondences, respectively, determined using a 15-pixel threshold.  
For the Aerial-Megadepth method, we project the ground-view panorama into three perspective images with a 120° field of view, perform matching separately, and then select the top 30 correspondences (out of 90 total) to reproject back into the panorama domain for visualization.  
Compared with existing methods, our approach achieves superior performance in large ground-satellite viewpoint differences, producing more reliable and spatially consistent correspondences across wide baselines.

\begin{figure*}[ht]
    \captionsetup[subfigure]{labelformat=empty}
    \tikzset{inner sep=0pt}
    \setkeys{Gin}{width=0.49\textwidth}
    \centering
    \subfloat[]{%
    \tikz{\node (a) {\includegraphics{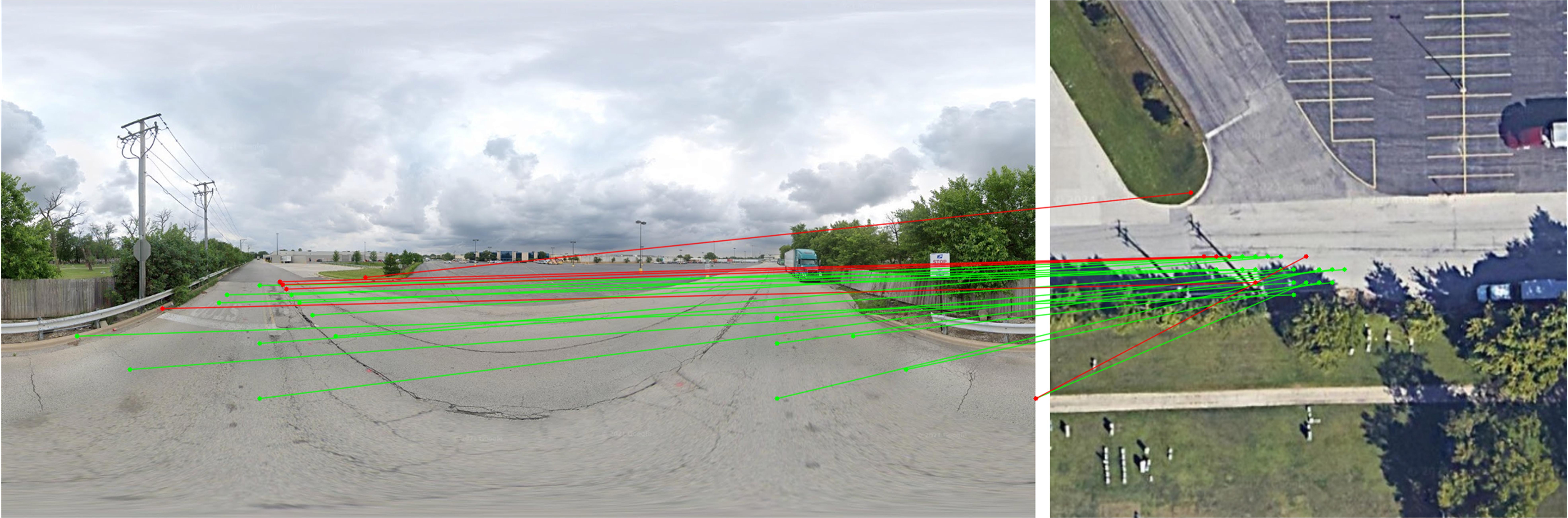}};
        \node[below right=2mm] at (a.north west) {(a)}; 
      }}
    \hfil
    \subfloat[]{%
    \tikz{\node (a) {\includegraphics{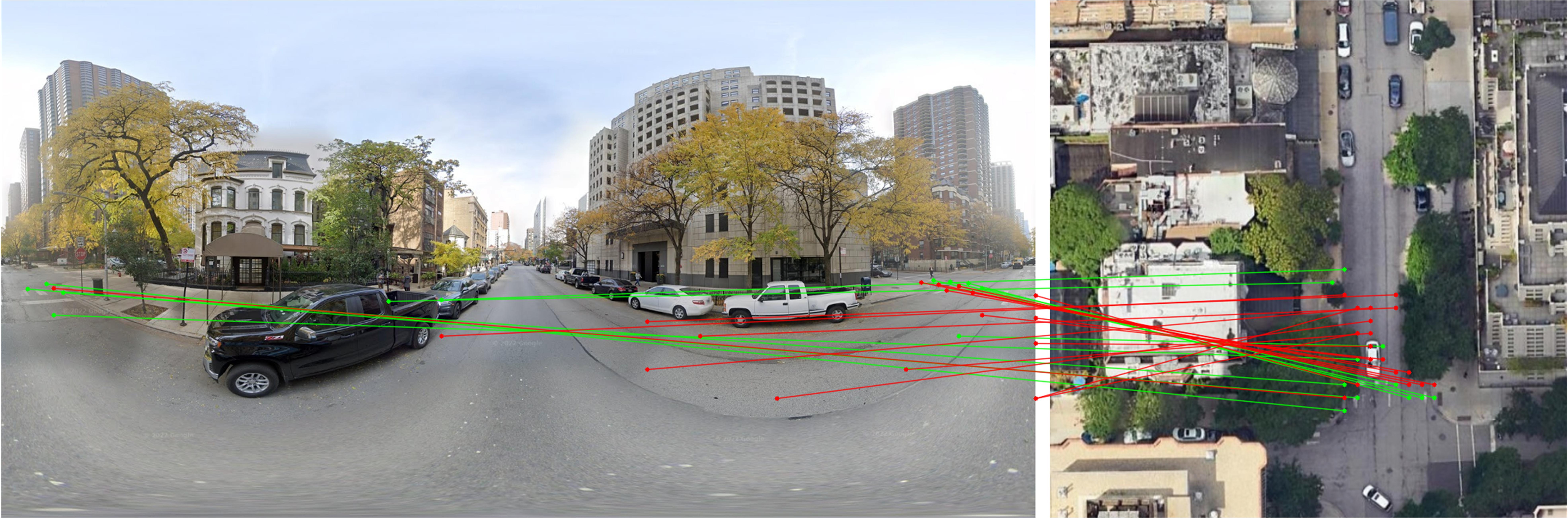}};
        \node[below right=2mm] at (a.north west) {(b)}; 
      }}
    \hfil
    \subfloat[]{%
    \tikz{\node (a) {\includegraphics{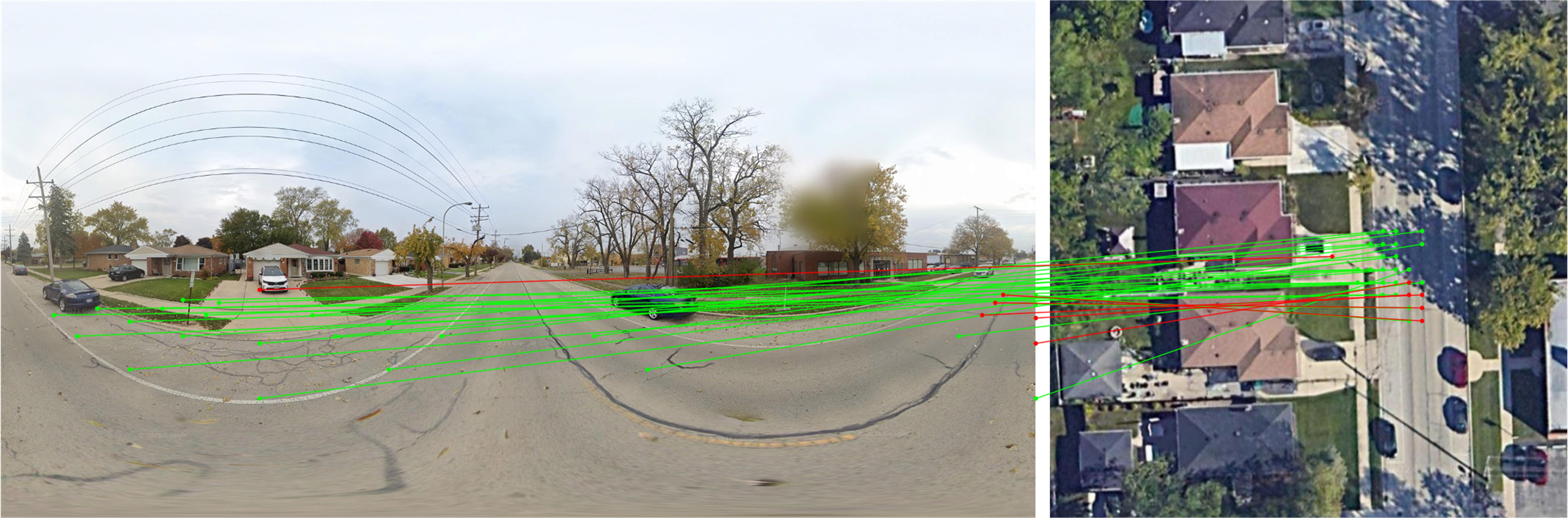}};
        \node[below right=2mm] at (a.north west) {(c)}; 
      }}
    \hfil
    \subfloat[]{%
    \tikz{\node (a) {\includegraphics{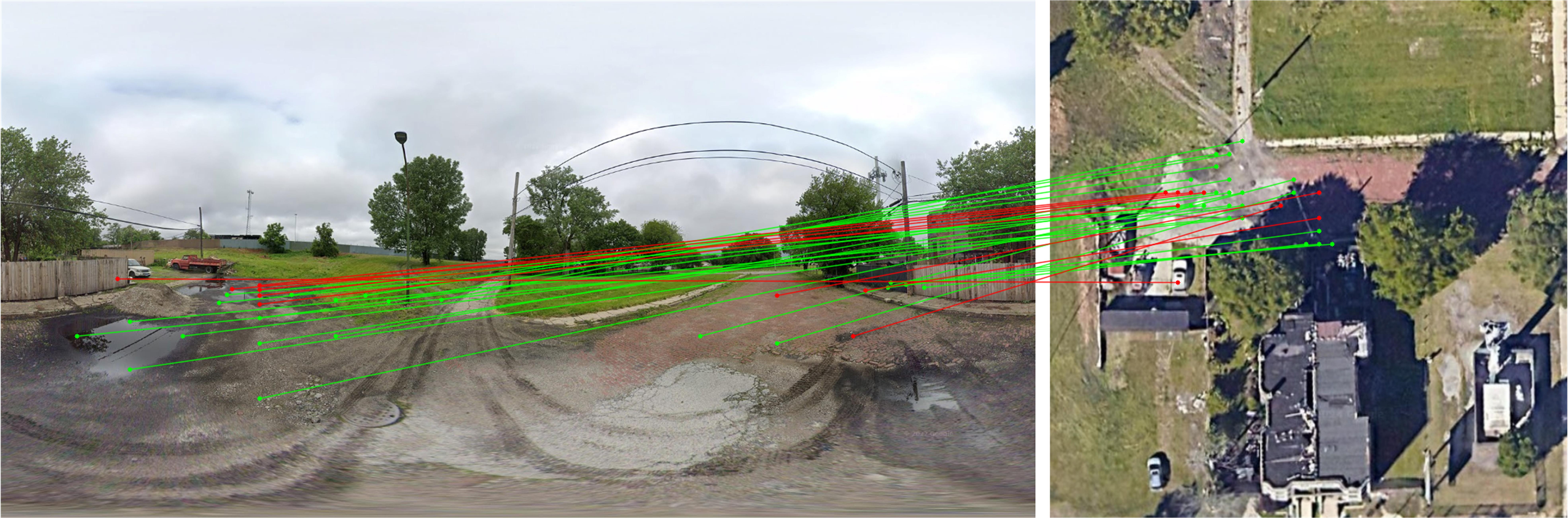}};
        \node[below right=2mm] at (a.north west) {(d)}; 
      }}
    \hfil
    \subfloat[]{%
    \tikz{\node (a) {\includegraphics{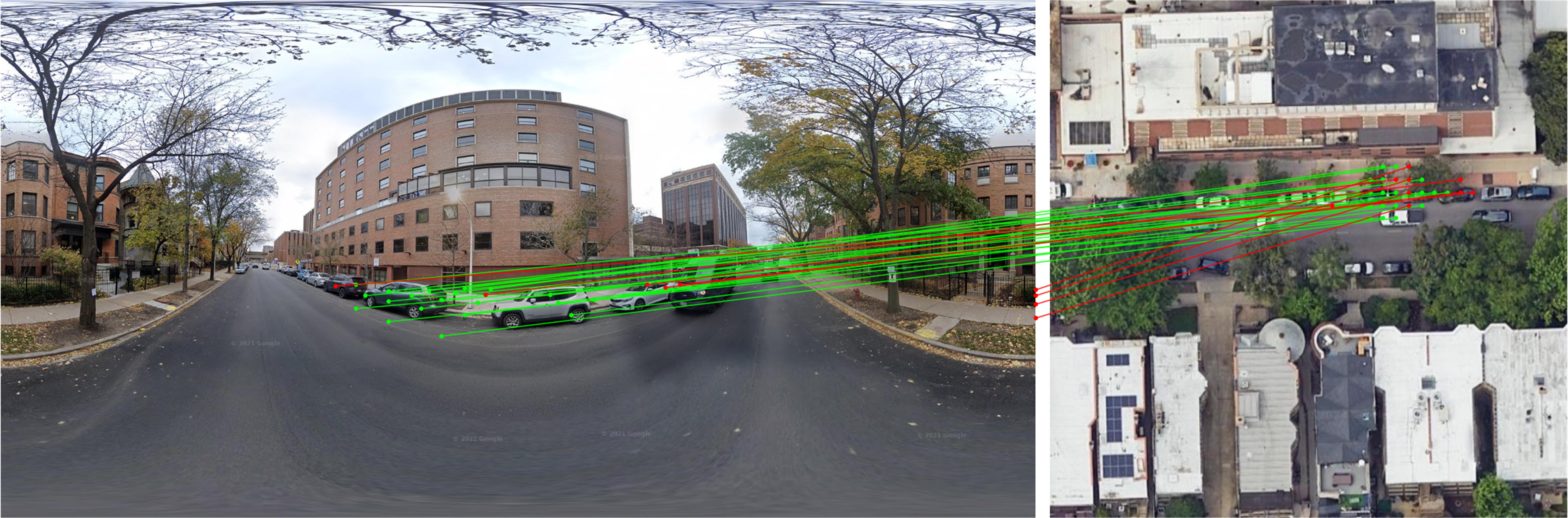}};
        \node[below right=2mm] at (a.north west) {(e)}; 
          }}
    \hfil
    \subfloat[]{%
    \tikz{\node (a) {\includegraphics{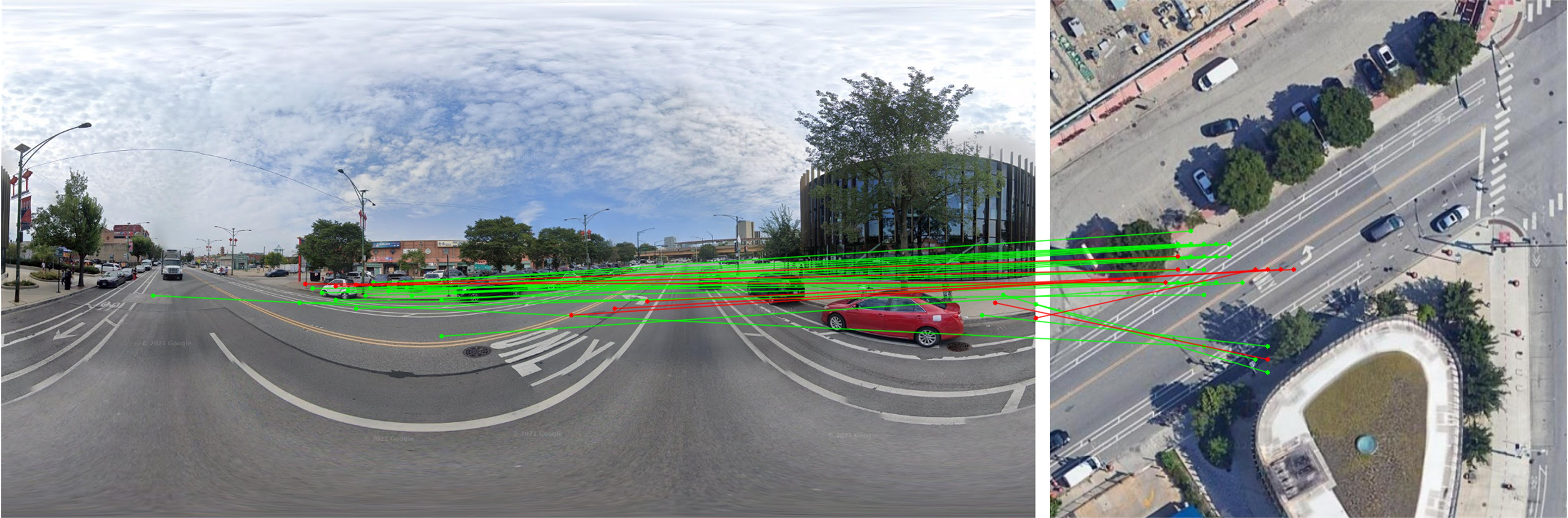}};
        \node[below right=2mm] at (a.north west) {(f)}; 
      }}
    \hfil
    \subfloat[]{%
    \tikz{\node (a) {\includegraphics{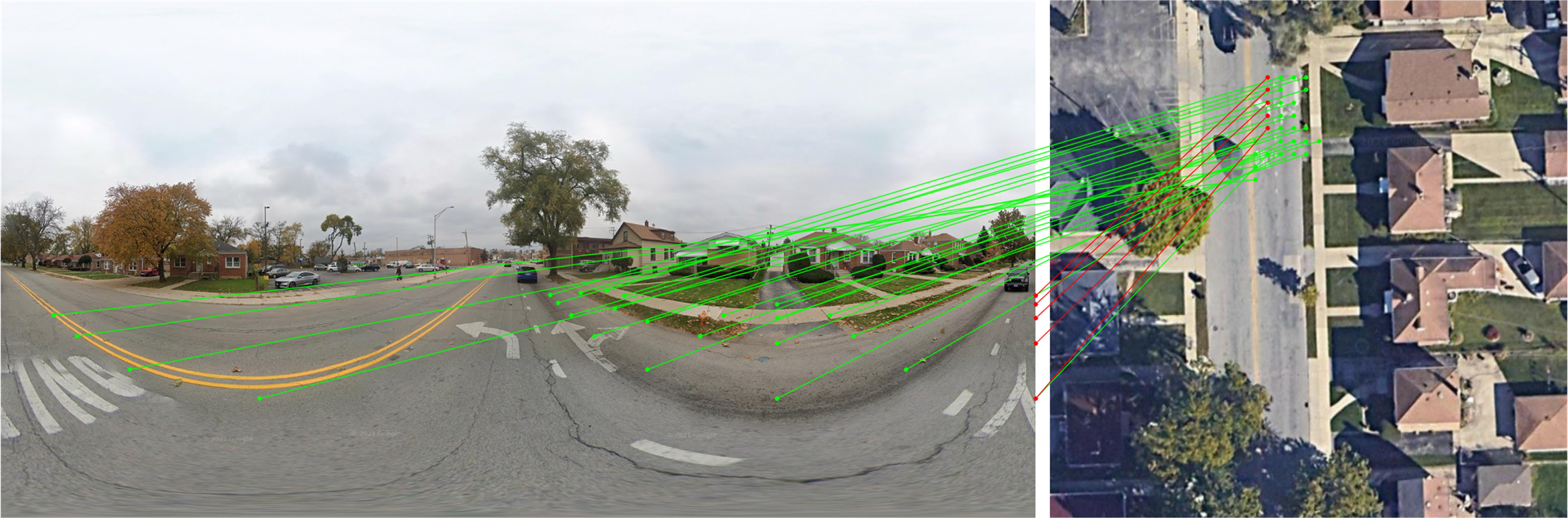}};
        \node[below right=2mm] at (a.north west) {(g)}; 
      }}
    \hfil
    \subfloat[]{%
    \tikz{\node (a) {\includegraphics{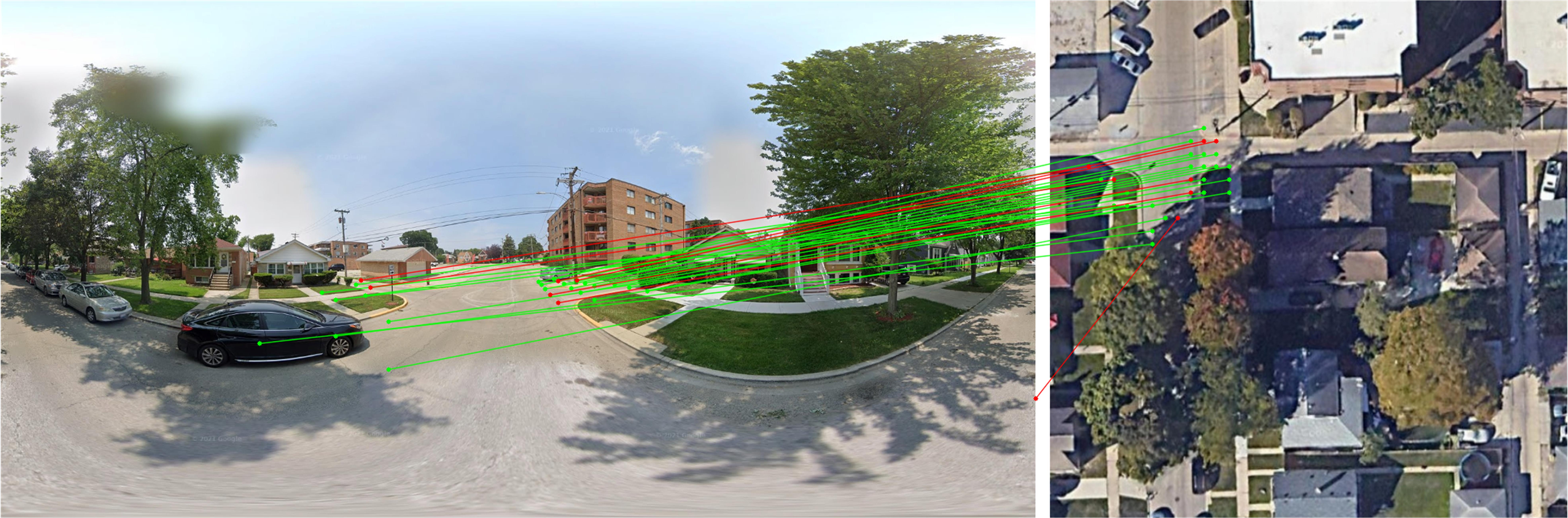}};
        \node[below right=2mm] at (a.north west) {(h)}; 
      }}
    \hfil
    \subfloat[]{%
    \tikz{\node (a) {\includegraphics{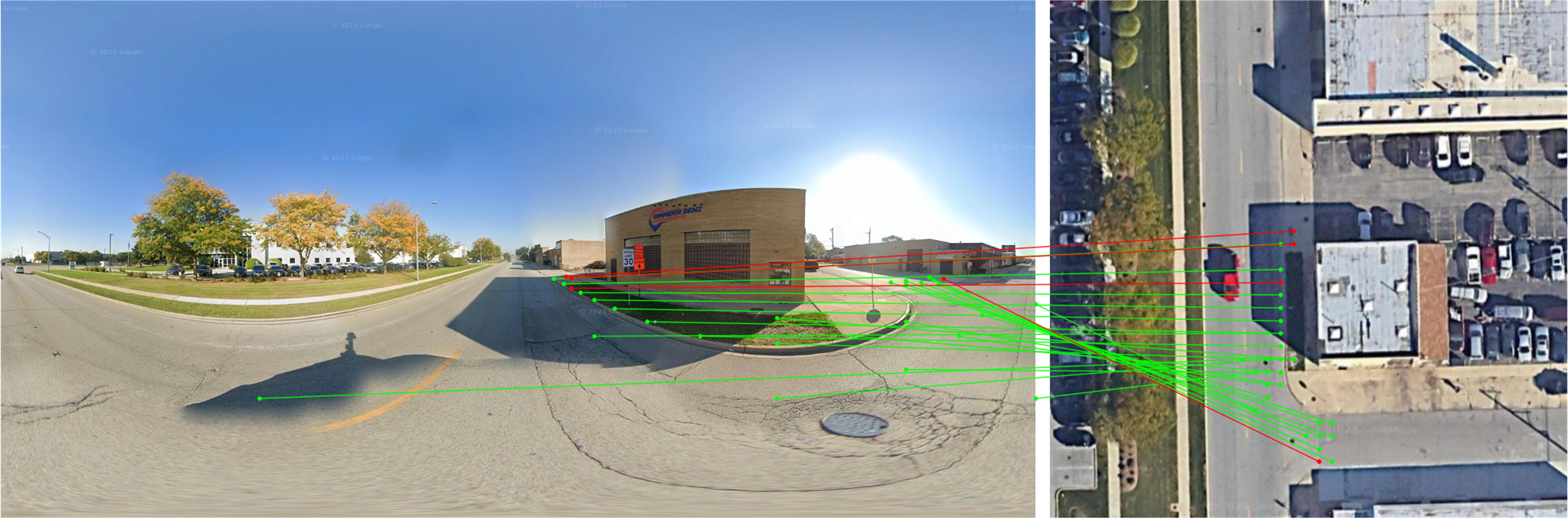}};
        \node[below right=2mm] at (a.north west) {(i)}; 
      }}
    \hfil
    \subfloat[]{%
    \tikz{\node (a) {\includegraphics{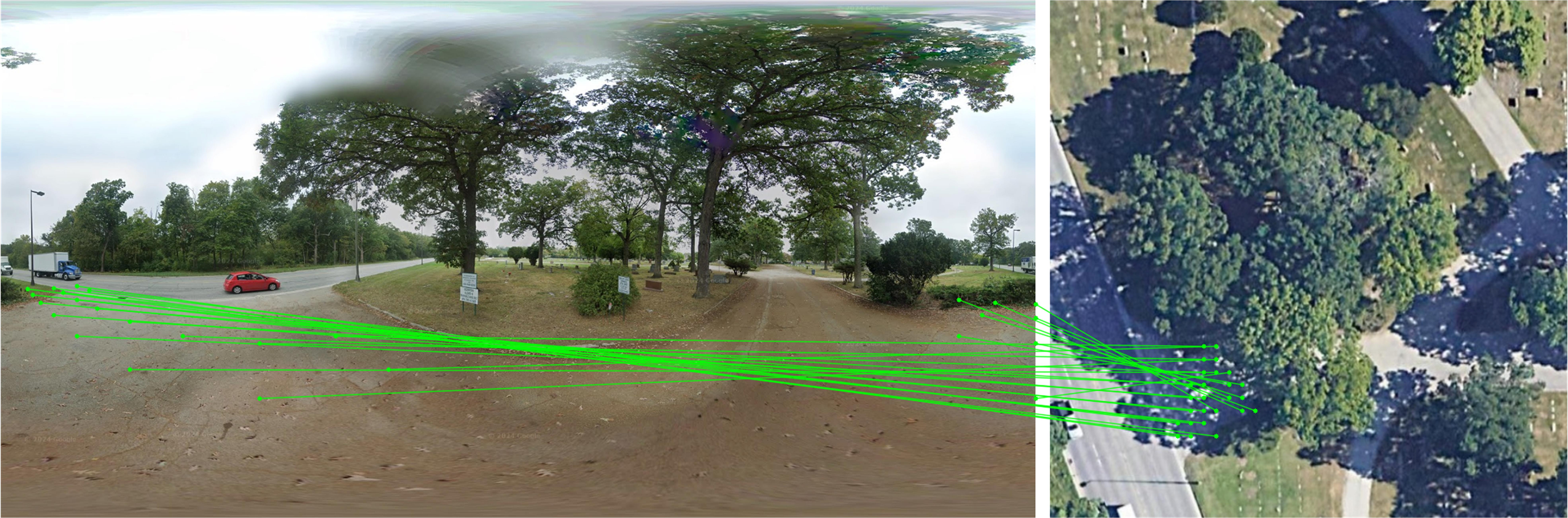}};
        \node[below right=2mm] at (a.north west) {(j)}; 
      }}
    \hfil
    \subfloat[]{%
    \tikz{\node (a) {\includegraphics{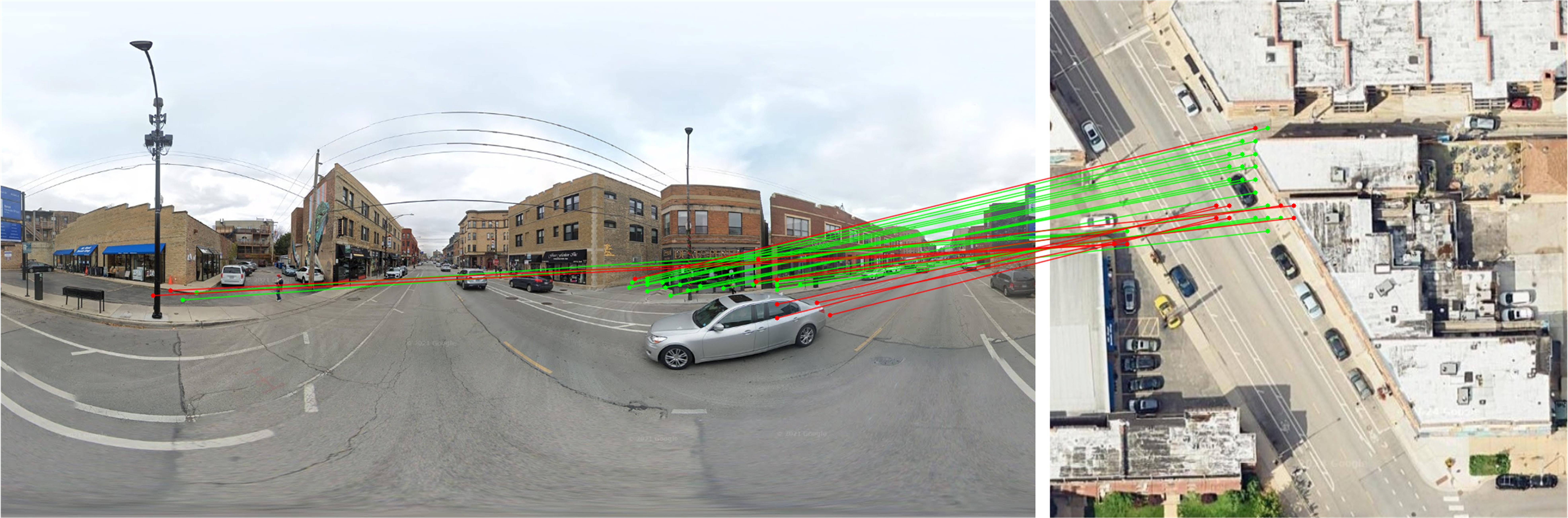}};
        \node[below right=2mm] at (a.north west) {(k)}; 
      }}
    \hfil
    \subfloat[]{%
    \tikz{\node (a) {\includegraphics{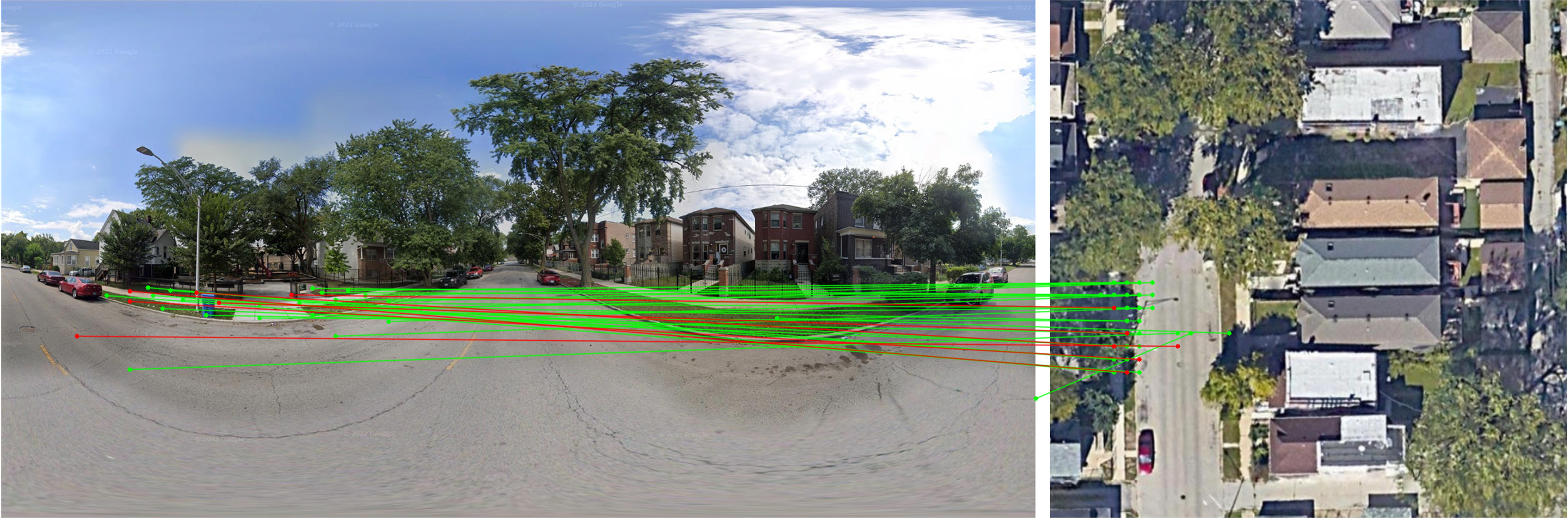}};
        \node[below right=2mm] at (a.north west) {(l)}; 
      }}
    \hfil
  \caption{
Additional examples of our cross-view image matching results on the CVFM benchmark. 
Each pair shows dense correspondences between ground panoramas and satellite images, where green and red lines represent correct and incorrect matches under a 15-pixel threshold. 
}

    \label{fig:extra_match}
\end{figure*}

\textbf{Additional cross-view image matching.}  
Fig.~\ref{fig:extra_match} provides more qualitative examples of our dense cross-view matching results. 
The visualizations demonstrate that our method maintains high geometric consistency and strong semantic alignment between panoramas and satellite views, even in challenging conditions such as occlusions, vegetation, and large illumination differences.  

Overall, these results reinforce the interpretability and robustness of our framework in both localization and cross-view image matching.

